\documentclass[pdflatex,sn-mathphys-num]{sn-jnl}
\usepackage{newunicodechar}

\usepackage{graphicx}
\usepackage{xcolor}
\usepackage{graphicx}%
\usepackage{multirow}%
\usepackage{amsmath,amssymb,amsfonts}%
\usepackage{amsthm}%
\usepackage{mathrsfs}%
\usepackage[title]{appendix}%
\usepackage{xcolor}%
\usepackage{textcomp}%
\usepackage{manyfoot}%
\usepackage{booktabs}%
\usepackage{subfig}
\usepackage{algorithm}%
\usepackage{algorithmicx}%
\usepackage{algpseudocode}%
\usepackage{listings}%
\usepackage{hyperref}
\usepackage{comment}
\usepackage{adjustbox}
\usepackage{float}
\usepackage{tabularx}
\usepackage{longtable}
\usepackage{lscape}
\usepackage{multirow}
\usepackage{boldline} 
\usepackage{booktabs} 

\theoremstyle{thmstyleone}%
\newcolumntype{Y}{>{\raggedright\arraybackslash}X} 

\theoremstyle{thmstyletwo}%

\theoremstyle{thmstylethree}%

\begin{document}

\title[Get away with less: Need of source side data curation to build parallel corpus for low resource Machine Translation 
]{Get away with less: Need of source side data curation to build parallel corpus for low resource Machine Translation
}


\author*{\fnm{Saumitra} \sur{Yadav}}\email{saumitra.yadav@research.iiit.ac.in}

\author{\fnm{Manish} \sur{Shrivastava}}\email{m.shrivastava@iiit.ac.in}


\affil{\orgdiv{Language Technologies Research Centre (LTRC)}, \orgname{International Institute Information Technology}, \orgaddress{ \city{Hyderabad}, \postcode{500032}, \state{Telangana}, \country{India}}}




\abstract{Data curation is a critical yet under-researched step in the machine translation (MT) training paradigm. To train translation systems, data acquisition  relies primarily on human translations and digital parallel sources or, to a limited degree, synthetic generation. But, for low-resource languages, human translation to generate sufficient data is prohibitively expensive. Therefore, it is crucial to develop a framework that screens source sentences to form efficient parallel text, ensuring optimal MT system performance in low-resource environments. We approach this by evaluating English-Hindi bi-text to determine effective sentence selection strategies for optimal MT system training. Our extensively tested framework, (\textbf{L}exical \textbf{A}nd \textbf{L}inguistically \textbf{I}nformed \textbf{T}ext \textbf{A}nalysis) LALITA, targets source sentence selection using lexical and linguistic features to curate parallel corpora. We find that by training mostly on complex sentences from both existing and synthetic datasets, our method significantly improves translation quality. We test this by simulating low-resource data availability  with curated datasets of 50K to 800K English sentences and report improved performances on all data sizes. LALITA demonstrates remarkable efficiency, reducing data needs by more than half across multiple languages (Hindi, Odia, Nepali, Norwegian Nynorsk, and German). This approach not only reduces MT systems training cost by reducing training data requirement, but also showcases LALITA's utility in data augmentation. }




\maketitle

\section{Introduction}\label{sec:intro}
The global landscape increasingly demands effective Machine Translation (MT) systems to bridge linguistic divides and facilitate information exchange. Traditionally, the development of robust MT systems has relied heavily on the availability (through human efforts) of vast quantities of parallel data. However, this reliance presents a significant challenge: while major languages benefit from abundant digital resources \cite{tiedemann-2012-parallel, kunchukuttan-etal-2018-iit, parida-etal-2020-odiencorp, siripragada-etal-2020-multilingual, haddow2020pmindia,nakazawa-etal-2021-overview,ramesh-etal-2022-samanantar, mujadia-sharma-2022-ltrc, gala2023indictrans2, pal-etal-2023-findings}, numerous other languages, particularly those classified as low-resource and coming from the global south, suffer from a severe scarcity of digitally available data \cite{ranathunga2023neural}. A thorough audit of existing multilingual datasets confirms this, revealing that many corpora for low-resource languages are unusable due to a high rate of noise and misalignment \cite{kreutzer-etal-2022-quality,ranathunga-etal-2024-quality}. This disparity creates a fundamental bottleneck, as acquiring high-quality parallel data through human translation for these languages is prohibitively expensive and resource-intensive \cite{jha2012tdil}, especially for the global south. This situation highlights a critical problem: the very languages that could benefit most from MT are precisely those where data acquisition is most constrained by economic and practical limitations.

To address this scarcity, foundational efforts have focused on creating and managing large-scale corpora from the ground up. For many low-resource languages, like Mizo (one of the languages spoken in Mizoram state of India), the primary challenge is the sheer lack of any parallel data. Initial efforts, such as the recent work constructing a large-scale Mizo-English parallel corpus \cite{10.1145/3610404}, focus on creating these foundational resources from scratch. Similarly, the Indian government's  TDIL program and the Indian Language Corpora Initiative (ILCI) \citet{jha2012tdil} have focused on building a foundational, annotated corpus for key languages. Similarly, \citet{mujadia-sharma-2022-ltrc} meticulously constructed Hindi-Telugu parallel Corpus using a multi-pronged approach that included human review and back-translation to ensure quality. These foundational efforts, while monumental, are often a first step, as they do not explore whether a smaller, more strategically selected subset of a corpus could achieve comparable or even superior performance. Our work builds on this progress by addressing the next logical challenge: how to most effectively guide such  corpus creation by strategically selecting source sentences for cost-efficient model training.

Given the high cost and limited capacity for human translation, it becomes imperative to develop strategies that ensure every translated sentence contributes optimally to the performance of an MT system. The conventional approach of simply gathering more data, irrespective of its content, proves inefficient. This is further highlighted by the availability of large, web-mined corpora. For instance, the \citet{suarez_asynchronous_2019} created the massive OSCAR corpus, providing billions of sentences in 166 languages. Similarly, the creation of the Samanantar corpus \cite{ramesh-etal-2022-samanantar} prioritized scale by using advanced mining and alignment techniques \cite{feng-etal-2022-language}. However, this approach, while valuable, may include a large number of simple or redundant sentences that do not meaningfully contribute to a model's learning. A detailed look at web-mined corpora confirms that their quality is not uniform throughout, with only the highest-ranked portions proving effective for training an NMT model \cite{ranathunga-etal-2024-quality}. This observation challenges the conventional wisdom of simply increasing data volume or aiming for broad diversity, advocating instead for strategic data selection based on inherent linguistic properties.

Instead of reactive filtering from noisy parallel data, research must focus on a more discerning selection process: identifying and prioritizing source sentences that yield the greatest learning signal for the MT model. This proactive approach can be seen in the development of tools like OpusFilter \cite{aulamo-etal-2020-opusfilter}, a configurable toolbox for filtering noisy parallel corpora. While effective for general cleaning, such toolboxes often rely on a combination of basic, low-level features that do not fully capture a sentence's intrinsic linguistic value. Our research addresses this challenge of selecting the most impactful source sentences when human translation resources are fixed, aiming to curate training data that maximizes MT system performance under resource constraints. This strategic shift from merely accumulating data to intelligently curating it holds the potential to democratize MT development, making it feasible for communities with limited resources to build and deploy effective translation systems.

This paper introduces LALITA (\textbf{L}exical \textbf{A}nd \textbf{L}inguistically \textbf{I}nformed \textbf{T}ext \textbf{A}nalysis), a rigorously tested framework designed for source sentence selection in parallel corpus creation. A key innovation within LALITA is the derivation of the \textbf{LALITA score}, a quantifiable metric for sentence complexity\footnote{When we refer to a sentence's complexity, we are primarily discussing its structural and syntactic attributes, such as dependency relations and part-of-speech distributions, rather than its semantic difficulty or its topic.}. 
A higher LALITA score signifies a more structurally complex sentence. The core finding of this work is that by strategically focusing on and curating complex sentences, whether from existing datasets or synthetic sources, LALITA significantly enhances translation quality while drastically reducing the required data volume. The contributions of this research are multi-faceted:
\begin{itemize}
\item Provides an extensively validated pipeline for source sentence analysis.
\item Offers a linguistically informed method for augmenting synthetic datasets.
\item Empirically demonstrates that training on structurally complex sentences yields comparable or superior MT performance (especially in low-resource contexts) while reducing training dataset requirements.
\item Presents a straightforward yet effective method (the LALITA score) for assessing structural complexity.
\item  In situations where the expense of developing and training MT data is high, LALITA provides an effective approach by determining the most useful data subset.
\end{itemize}
Furthermore, the results show that LALITA achieves \textbf{comparable} performance for English-Hindi MT with over 60\% less data (800K sentence pairs compared to 1.8 million) and demonstrates its utility in high-resource scenarios like English-German , where it reduces data requirements by approximately 60\% (8 million compared to 20 million) while simultaneously \textbf{improving} translation performance. This strategic approach also yields significant data reductions for other low-resource language pairs, including English to Odia (14\% of the original data), English to Nepali (a reduction by 50\%), and English to Norwegian Nynorsk (64\% of the original data). This suggests that complex source sentences carry a disproportionately higher learning signal for MT models. By identifying and prioritising these sentences using the LALITA score, the model extracts more valuable information from fewer examples, leading to both efficiency (less data) and efficacy (better performance). 

\section{English-Hindi Data and preprocessing}
\subsection{Samanantar English-Hindi Bi-text}
The dataset used for the main experiments in this research is the Samanantar English-Hindi bi-text \cite{ramesh-etal-2022-samanantar}. This comprehensive corpus acts as a vital tool for simulating different data preprocessing approaches and pipelines specifically designed for organizing source sentences. In this study, the Hindi portion of the bitext, along with Monolingual Hindi employed for synthetic English generation, is regarded as human-translated text, serving as a dependable target for evaluation. The choice of Samanantar is motivated by its large volume, which facilitates an in-depth analysis of diverse data management strategies in the field of Machine Translation. 

\subsection{Data Filtration: Ensuring Quality and Reliability}
\begin{figure}[ht]
  \centering
  \includegraphics[width=1\columnwidth]{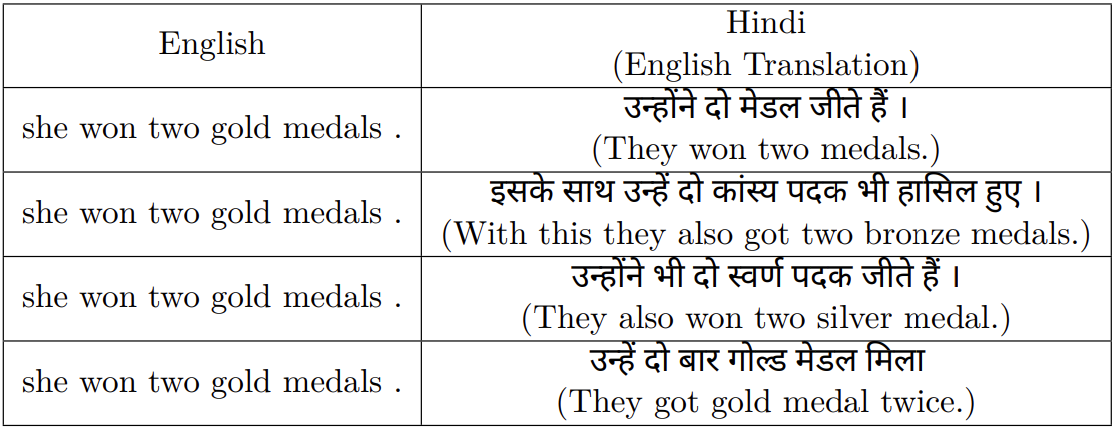}
  \caption{Example from Samanantar: Multiple Wrong Translations}
  \label{fig:issue}
\end{figure}

To maintain the integrity and reliability of the training data, a rigorous filtering process is applied to the raw Samanantar dataset. This step is necessary due to observed noise within the dataset, including cases where a single English source sentence associates with multiple, often incorrect, Hindi translations, as shown in Figure \ref{fig:issue}.
These ambiguities may lead to inaccuracies and obstruct the MT system's learning process. Following bitext deduplication, the following steps are focused on eliminating sentence pairs:
\begin{itemize}
    \item Any Hindi sentence featuring more than 35\% tokens in the Roman script is discarded. Importantly, this condition is exclusive to the English-Hindi pair and it is not used for other experiments involving English $\rightarrow$ Odia, Nepali, and Norwegian Nynorsk because of a lack of sufficient parallel text. Additionally, it, obviously,  does not apply to English-German datasets.
\item Sentence pairs with a length ratio over 4 between English and Hindi are removed. Such significant length differences typically indicate misalignment or subpar translations.
\item If a sentence in one language has more than one corresponding translation in another language, all related pairs are eliminated. This step is essential for the English-Hindi dataset because it is impossible to conclusively determine the "true" translation among several choices. However, it is not applied to other experiments involving English $\rightarrow$ Odia, Nepali, and Norwegian Nynorsk due to a lack of sufficient parallel text.
\item Keeping sentence pairs, if the source instance is a single sentence. 
\end{itemize}
\begin{table}
\centering
\caption{Performance of Samanantar Data set (CHRF++score) Pre (8 Million) and Post-filtering (1.8 Million)}
\label{tab:filtering}
\begin{tabular}{|m{5cm}|m{5cm}|} 
\hline
Pre-filtering & Post-filtering  \\ 
\hline
52.43         & 53.13           \\
\hline
\end{tabular}
\end{table}
The Samanantar dataset, after thorough filtration, resulted in 1.85 million clean sentence pairs. These pairs are designated as the Filtered English-Hindi Parallel Sentences (FPS) and serve as the entire training data for the English-Hindi language pair in this study. A notable validation of this filtration method is the observed enhancement in performance when training on FPS compared to the original, unfiltered dataset, as shown in Table \ref{tab:filtering}, with the model architecture and hyperparameters detailed in Section \ref{subsec:modelArch}. This outcome is particularly significant as it implies that the eliminated data was extraneous noise or redundant information rather than a valuable training signal. This underscores that effective data curation entails not only the judicious selection to reduce volume but also the elimination of noisy data, thereby minimizing computational demands without sacrificing quality.
\subsection{Model Architecture and Training Setup}
\label{subsec:modelArch}
For English, Norwegian Nynorsk preprocessing is done using Moses scripts\footnote{\url{https://github.com/moses-smt/mosesdecoder/}}. Hindi, Odia, and Nepali texts are pre-processed with the IndicNLP library \cite{kunchukuttan2020indicnlp}, specialized for Indian languages. 
Each experiment employs a Transformer architecture \cite{vaswani2017attention} for training, with hyperparameters specified in the Appendix \ref{subsec:hyperparameters}. Text data for each language undergoes tokenization via Byte-Pair Encoding (BPE) \cite{sennrich-etal-2016-neural}, trained on Filtered Parallel Sentences (FPS) with 16,000 merge operations for all experiments. This uniform tokenization strategy is implemented across all systems, and their performance is evaluated on the FLORES \cite{goyal-etal-2022-flores} test set using CHRF++ \cite{popovic-2015-chrf} after detokenization with Sacrebleu \cite{post-2018-call}.
\section{ LALITA Score: Feature Engineering and Complexity Quantification}
\subsection{Analysis of English Source Sentences}

\begin{figure}[!ht]
\subfloat[]{\includegraphics[width=0.48\textwidth]{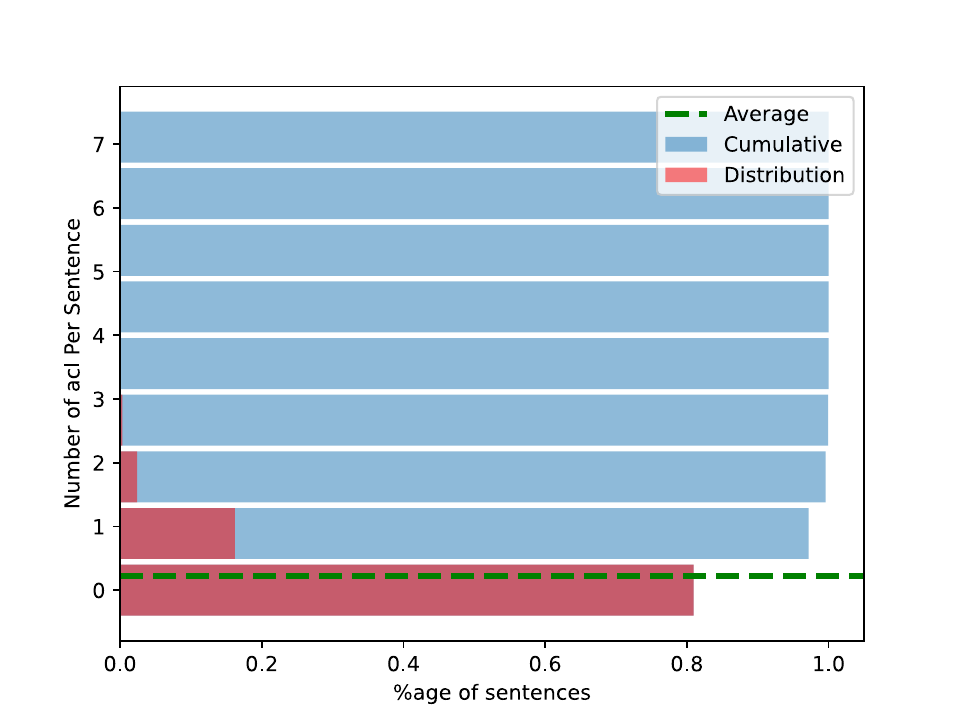}}\hfill
\subfloat[]{\includegraphics[width=0.48\textwidth]{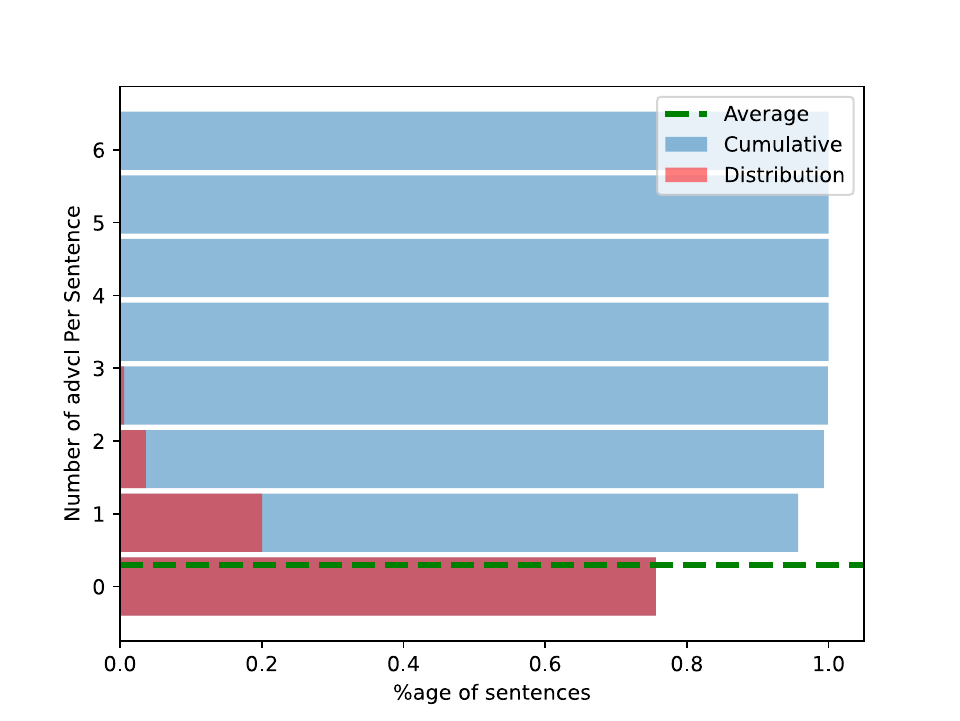}}\hfill

\subfloat[]{\includegraphics[width=0.48\textwidth]{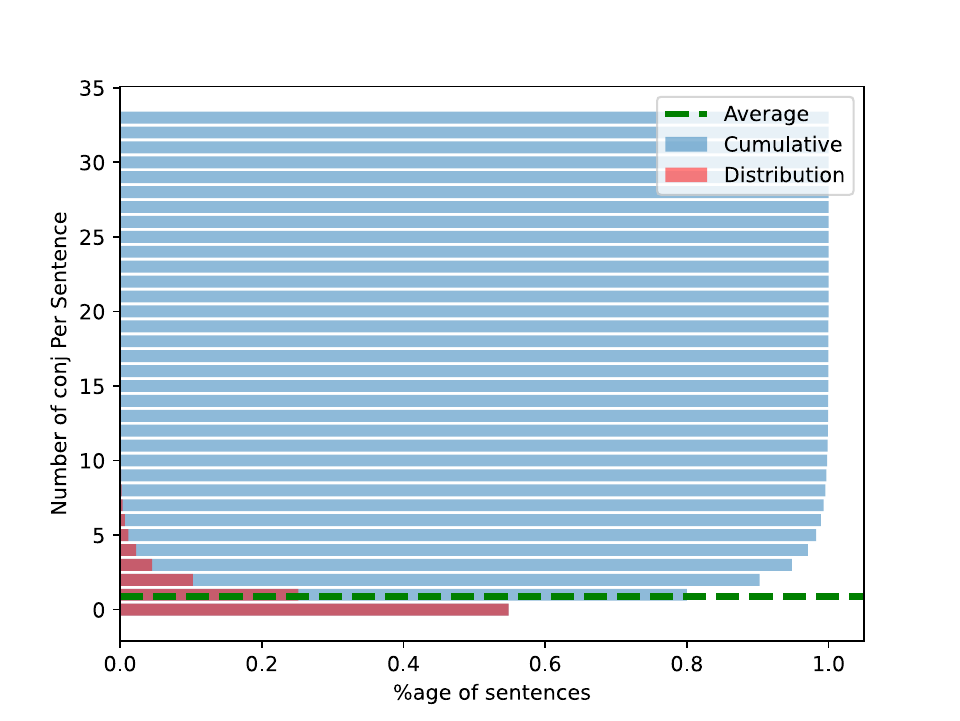}}\hfill
\subfloat[]{\includegraphics[width=0.48\textwidth]{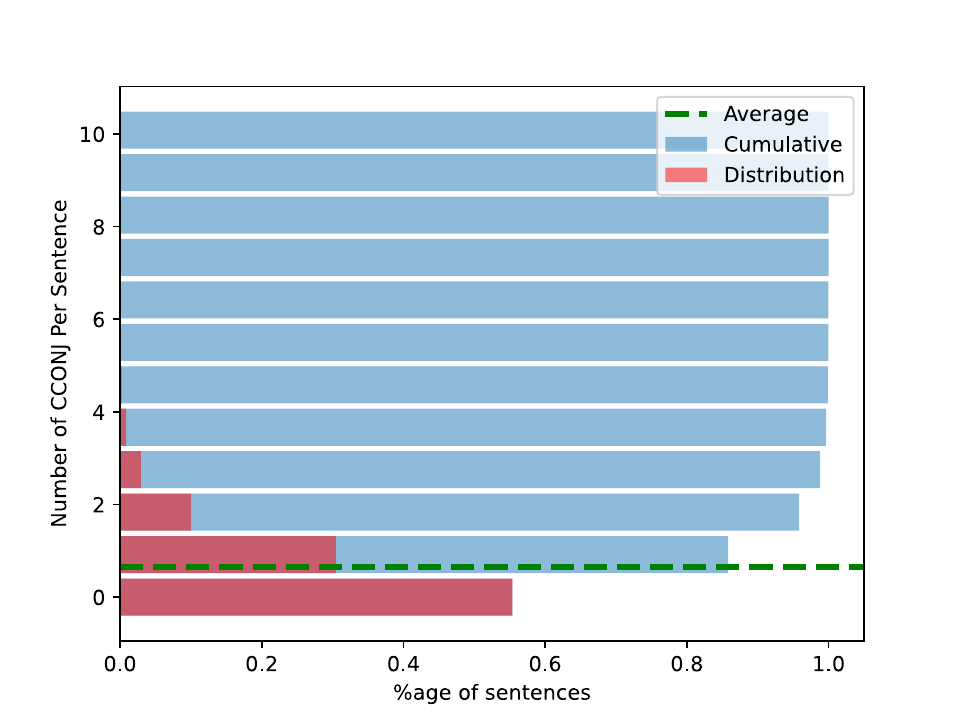}}\hfill

\subfloat[]{\includegraphics[width=0.48\textwidth]{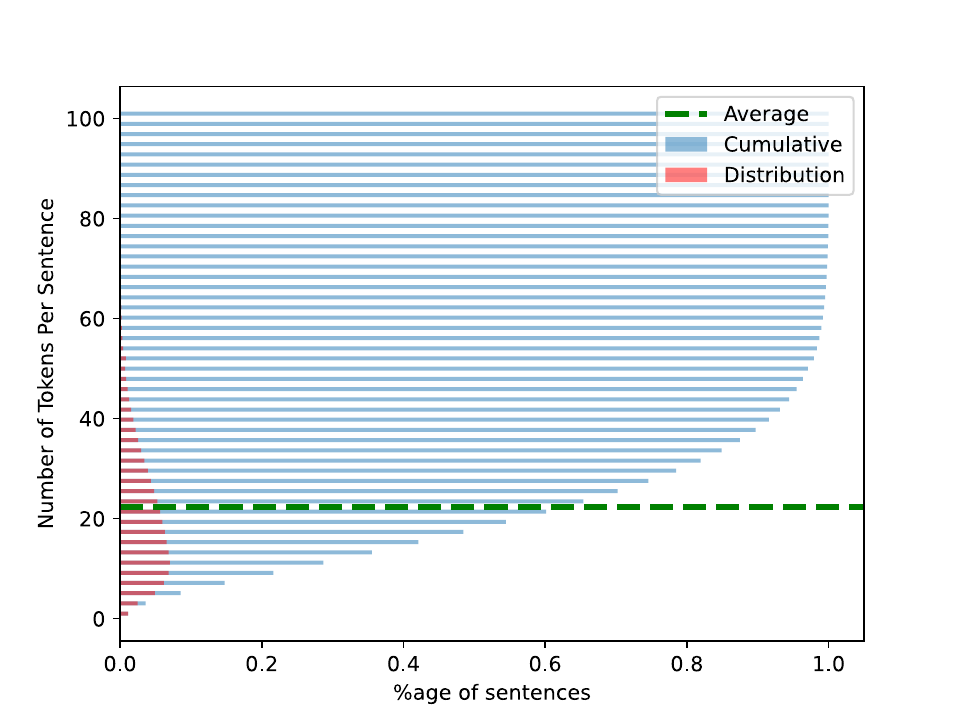}}\hfill
\subfloat[]{\includegraphics[width=0.48\textwidth]{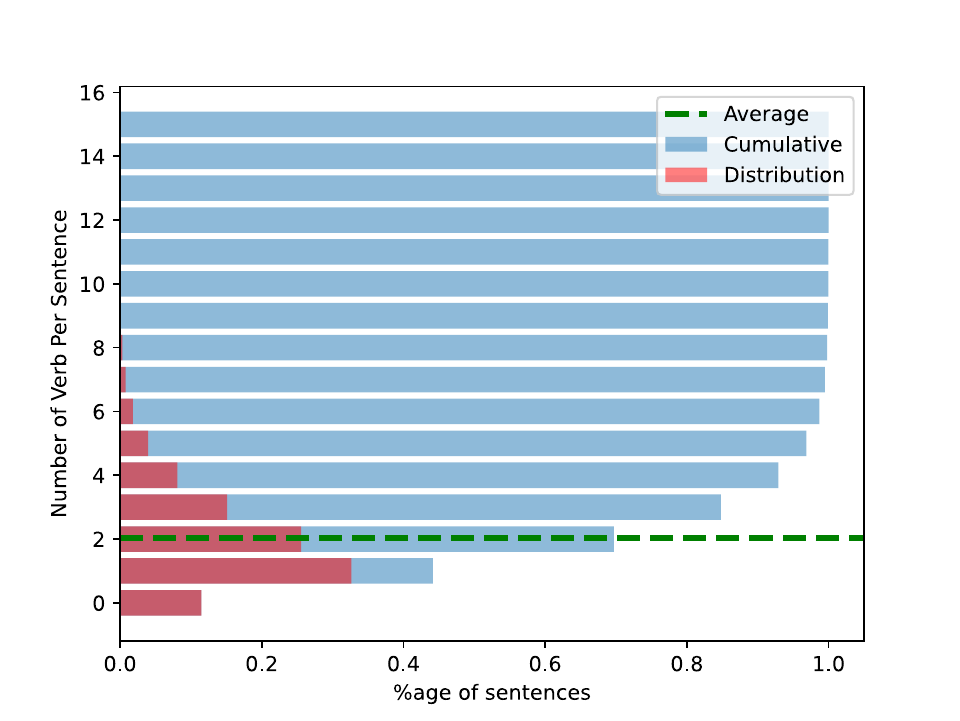}}\hfill

\caption{Feature-wise distribution of sentences in Training Data}
\label{fig:training_Sentence_Distribution_Dep}
\end{figure}

In this research, the initial phase entails a comprehensive exploratory analysis of English source sentences from the Filtered English-Hindi Parallel Sentences (FPS) dataset. The goal is to comprehend the intrinsic distribution of these sentences regarding their syntactic diversity, acting as a proxy for complexity. Trankit \cite{nguyen-etal-2021-trankit}, a transformer-based toolkit for multilingual natural language processing, is utilized to parse the English sentences in FPS and extract various linguistic features, as shown in Figure \ref{fig:training_Sentence_Distribution_Dep}.
The analysis revealed several key characteristics of the dataset's composition:
\begin{itemize}
    \item \textbf{Sentence Length (Number of Tokens)}: A substantial portion of sentences, roughly 60\%, contain 23 tokens or fewer. The average sentence length stands at 22, with a median of 20 and a standard deviation of 12.79. This suggests that shorter sentences are common in the corpus.
\item Part-of-Speech (POS) Tags: The distribution of verbs and conjunctions (CCONJ) points towards simpler structures. About 69.69\% of sentences contain at most two verbs, with 11.49\% having none, 32.65\% having one, and 25.55\% having two. A similar distribution is observed for the CCONJ POS tag.
\item \textbf{Dependency Tree Features}: Advanced markers of syntactic complexity, like clausal modifiers (acl), adverbial modifiers (advcl), and conjunctions (conj), are often missing. Notably, 80.96\% of sentences lack acl edges, advcl edges are missing in 75.65\% of sentences, and 54.84\% of sentences do not include a conj dependency link, while another 25.16\% have just one.
 
\end{itemize}

These findings collectively indicate that the dataset predominantly comprises sentences with simpler structures and shorter lengths. This observation is critical because it highlights a representational bias in typical parallel corpora towards simpler sentences. Consequently, an MT model trained on such a dataset would likely perform well on simple sentences but will struggle with more complex linguistic structures encountered in real-world applications. This inherent limitation in data distribution underscores the necessity of actively curating complex sentences to ensure robust MT system performance across the full spectrum of linguistic complexity. To address this challenge, it becomes essential not only to recognize such biases but also to systematically characterize the linguistic and lexical composition of sentences. By quantifying features that reflect sentence complexity, lexical diversity, and structural variation, one can better identify gaps in the dataset and strategically select useful instances. Such an approach ensures that corpus creation is not left to chance but is guided by informed, data-driven principles that maximize both cost-efficiency and the eventual robustness of the MT system.
\subsection{Feature Vector Creation}
To obtain a complete grasp of the linguistic and lexical characteristics of each sentence, a multi-dimensional feature vector is constructed. This method is selected to prevent information loss and allow for an in-depth exploration of optimal data curation possibilities. The feature vector for each sentence incorporates statistical, lexical, and linguistic attributes, offering a comprehensive descriptor. 
The specific features included are:
\begin{itemize}
    \item \textbf{Statistical Features}: The perplexity of the sentence is determined using both \href{https://github.com/facebookresearch/fairseq/blob/main/examples/language_model/README.md}{neural} \cite{baevski2018adaptive} and 5-gram statistical language models. These models are trained on the English text of the parallel dataset, utilizing Kneser-Ney smoothing for the statistical model \cite{479394}. For LM training, we lowercase, normalize, and tokenize English sentences.
\item \textbf{Lexical Features}: The length of the sentence, measured by the number of tokens.
\item \textbf{Linguistic Features}: These are quantified by counting the occurrences of various linguistic elements, recognizing that certain features can make a sentence distinctive simply by appearing multiple times.
\begin{itemize}
    \item \textbf{Named Entities}: Counts for four categories: Location, Person, Organization, and Miscellaneous.
\item \textbf{Part of Speech (POS) Tags}: Counts for 17 different POS tags.
\item \textbf{Dependency Relations}: Counts for 68 distinct dependency relations.
\item \textbf{Universal Morphological Features} \cite{de-marneffe-etal-2021-universal}: Counts for 156 features, including a ``NoUMF" tag for words without specific Universal Morphological Features. To determine the frequency with which features \textit{X} adopt the values $x_1$, $x_2$, $x_3$, we record the occurrences of X\_$x_1$, X\_$x_2$, X\_$x_3$ within a sentence. For instance, \textit{animacy} can assume four values: Animate(Anim), human(Hum), inanimate(Inan), and nonhuman(Nhum). Therefore, we count the instances where words in a sentence exhibit animacy, such as animacy\_Anim, animacy\_Hum, animacy\_Inan, and animacy\_Nhum.
\end{itemize}

\end{itemize}
\begin{figure*}[ht]
  \centering
  \includegraphics[width=1\linewidth]{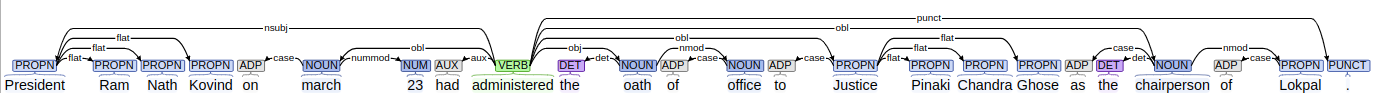}
  \caption{A parsed Sentence}
  \label{fig:sentenceExample}
\end{figure*}%

These diverse features are concatenated to form a comprehensive sentence descriptor. For instance, consider the sentence ``President Ram Nath Kovind on March 23 had administered the oath of office to Justice Pinaki Chandra Ghose as the chairperson of Lokpal." (Figure \ref{fig:sentenceExample}). Its feature vector would reflect specific counts such as one verb, two persons, one organization, one auxiliary word, and two words in the past tense. This multi-faceted definition of complexity, encompassing a wide array of linguistic properties, is crucial for developing a robust and nuanced metric. This detailed feature engineering is done to ensure that data curation has a deep linguistic understanding, moving beyond superficial metrics like sentence length.
The structure of a partial sentence feature vector is illustrated in Table \ref{tab:sentenceVector}.
\begin{table*}[]
\centering
\caption{Partial Sentence Feature Vector. For each group of features, such as the Named Entity tag, we sorted the values alphabetically according to the feature name. Thus, for Named Entity, the fourth cell of the sentence vector will be Location, the fifth will be Miscellaneous, the sixth will be Organisation, and so on. For POS, the feature value of ADJ will be recorded first in the eighth cell, followed by ADP, etc.}\label{tab:sentenceVector}
\resizebox{\textwidth}{!}{%
\begin{tabular}{|c|cc|c|cccc|ccccc|cccc|cccc|}
\hline
Groups &
  \multicolumn{2}{c|}{Statistical Features} &
  \begin{tabular}[c]{@{}c@{}}Leixcal\\ Feature\end{tabular} &
  \multicolumn{4}{c|}{Named Entity} &
  \multicolumn{5}{c|}{Part of Speech} &
  \multicolumn{4}{c|}{Dependency Relation} &
  \multicolumn{4}{c|}{Universal Morphological Features} \\ \hline
Feature &
  \multicolumn{1}{c|}{\begin{tabular}[c]{@{}c@{}}NLM \\ Perplexity\end{tabular}} &
  \begin{tabular}[c]{@{}c@{}}SLM\\ Perplexity\end{tabular} &
  \begin{tabular}[c]{@{}c@{}}Sentence\\ Length\end{tabular} &
  \multicolumn{1}{c|}{Location} &
  \multicolumn{1}{c|}{MISC} &
  \multicolumn{1}{c|}{ORG} &
  PER &
  \multicolumn{1}{c|}{ADJ} &
  \multicolumn{1}{c|}{ADP} &
  \multicolumn{1}{c|}{...} &
  \multicolumn{1}{c|}{VERB} &
  X &
  \multicolumn{1}{c|}{acl} &
  \multicolumn{1}{c|}{acl:relcl} &
  \multicolumn{1}{c|}{...} &
  xcomp &
  \multicolumn{1}{c|}{Abbr\_Yes} &
  \multicolumn{1}{c|}{...} &
  \multicolumn{1}{c|}{Voice\_Rcp} &
  others \\ \hline
Value &
  \multicolumn{1}{c|}{32} &
  19.2503 &
  24 &
  \multicolumn{1}{c|}{0} &
  \multicolumn{1}{c|}{0} &
  \multicolumn{1}{c|}{1} &
  2 &
  \multicolumn{1}{c|}{0} &
  \multicolumn{1}{c|}{5} &
  \multicolumn{1}{c|}{...} &
  \multicolumn{1}{c|}{1} &
  0 &
  \multicolumn{1}{c|}{0} &
  \multicolumn{1}{c|}{0} &
  \multicolumn{1}{c|}{...} &
  0 &
  \multicolumn{1}{c|}{0} &
  \multicolumn{1}{c|}{0} &
  \multicolumn{1}{c|}{0} &
  0 \\ \hline
\end{tabular}%
}
\end{table*}

\subsection{Dimension Reduction with PCA: Deriving the LALITA Score}



\begin{figure}
    \centering
  \includegraphics[width=1\linewidth]{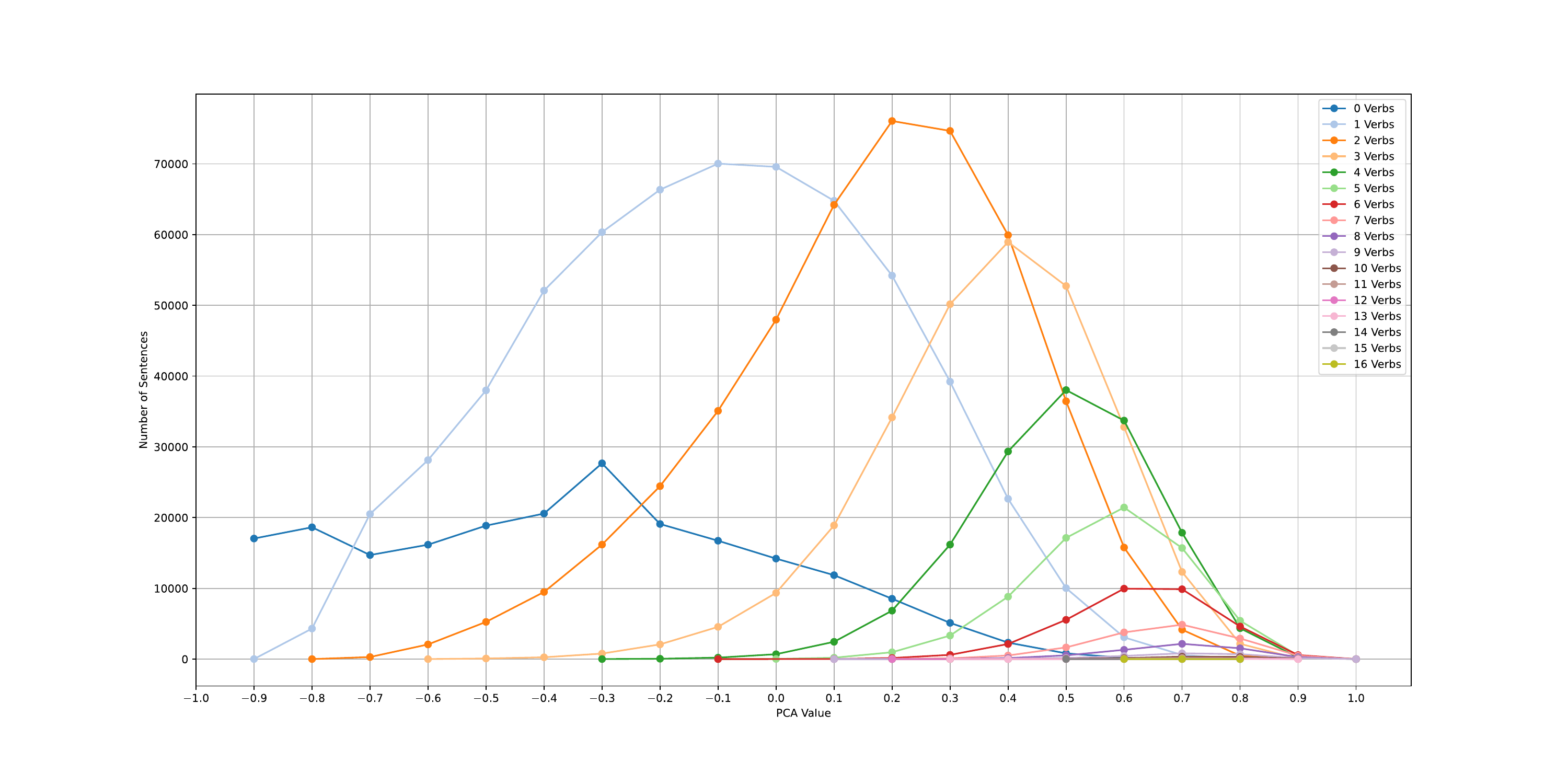}
  
\caption{Distribution of PCA1 Value v/s Number of sentence for Number of Verbs per sentence.}
\label{fig:pca1Verb}
\end{figure}

The initial count-based feature vectors are generally high-dimensional and sparse, posing challenges for direct processing and generalization. To make the dataset more manageable while retaining the most informative aspects, dimensionality reduction is performed using Principal Component Analysis (PCA). The transformation process consists of three main steps:
\begin{itemize}
\item \textbf{Standardization:} Features are rescaled to achieve comparable ranges, mitigating the impact of prominent numerical values.
\item \textbf{Normalization:} Each feature vector is adjusted to ensure it has a uniform magnitude, promoting balanced contributions across features.
\item \textbf{PCA Transformation:} The vectors are projected onto a lower-dimensional space, capturing the principal directions of variance that are most pertinent for analysis.
\end{itemize}
A significant discovery emerges from this process: the first component of \(X_{PCA}\), where $X$ represents a feature vector, identified in our work as \textbf{PCA1}, directly correlates with the structural complexity of a sentence. This finding is particularly valuable because PCA components are often abstract and challenging to interpret. Interpreting PCA1 as a measure of complexity bridges the gap between complex statistical methods and practical linguistic insights, providing a powerful and actionable metric. Consequently, PCA1 of the feature vector is labeled as the \textbf{LALITA score}. A higher LALITA score indicates greater sentence complexity, reflecting a transition from low negative to high positive PCA1 values with increasing complexity, as shown in Figure \ref{fig:pca1Verb}, which displays the frequency of sentences for each PCA against the frequency of verbs per sentence (Figures for features such as conjunction, acl, advcl, and token count are provided in Appendix \ref{app:pca1Feature}).
Table \ref{tab:coeffPCA1} shows the features with highest absolute coefficient values for the first principal component (PCA1), underscoring their role in influencing the LALITA score. Appendix \ref{sec:all_coeff_pca1} displays all non-zero coefficient absolute magnitudes.
\begin{table}[]
\caption{Features with the highest Absolute Magnitude of coefficient of 1st principal component.}
\label{tab:coeffPCA1}
\begin{tabular}{cc|cc|cc}
\hline
Feature & \begin{tabular}[c]{@{}c@{}}Absolute \\ Magnitude \\ of coefficient\end{tabular} & Feature & \begin{tabular}[c]{@{}c@{}}Absolute \\ Magnitude \\ of coefficient\end{tabular} & Feature & \begin{tabular}[c]{@{}c@{}}Absolute \\ Magnitude \\ of coefficient\end{tabular} \\ \hline
sentenceLength & 0.244 & VerbForm\_Fin & 0.167 & cc & 0.132 \\
NoUMF & 0.219 & Definite\_Def & 0.158 & CCONJ & 0.132 \\
VERB & 0.190 & Mood\_Ind & 0.149 & punct & 0.130 \\
ADP & 0.185 & obl & 0.145 & PUNCT & 0.129 \\
case & 0.185 & Tense\_Past & 0.143 & nsubj & 0.126 \\
NOUN & 0.185 & nmod & 0.140 & Number\_Plur & 0.126 \\
Number\_Sing & 0.184 & amod & 0.138 & obj & 0.123 \\
det & 0.181 & ADJ & 0.138 & AUX & 0.120 \\
DET & 0.180 & mark & 0.137 & SCONJ & 0.118 \\
PronType\_Art & 0.177 & Degree\_Pos & 0.135 & VerbForm\_Part & 0.116 \\ \hline
\end{tabular}
\end{table}
Table \ref{tab:examplePCA1} illustrates the relationship between PCA1 and complexity by presenting examples of sentences that exhibit diverse LALITA scores (PCA1 values). The table reveals a clear association of the LALITA score with the length of the sentence, the number of verbs, and the overall structural complexity.
\begin{table}
\caption{Sentences with PCA1 Values, sentence length and number of verbs, to show the rise in complexity as PCA1 increases.}
\label{tab:examplePCA1}
\begin{tabularx}{\textwidth}{|c|Y|c|c|}
\hline
PCA1 Value & Preprocessed Sentences & \begin{tabular}[c]{@{}c@{}}Sentence \\ length\end{tabular} & \begin{tabular}[c]{@{}c@{}}Number \\ of Verbs\end{tabular} \\ \hline
-0.81 & state legislation enactments & 3 & 0 \\ \hline
-0.51 & the minister has denied any wrongdoing & 6 & 1 \\ \hline
0 & neither speaker kanwar pal gujjar nor chief minister manohar lal khattar paid heed to abhays demand . & 17 & 1 \\ \hline
0.51 & and we also measured depth to the water level in these wells , and then we had done this on 6 times in october 2016 , november 2016 , then december and then january , february , and march . & 40 & 2 \\ \hline
1 & while felicitating the team of scientists , shri pokhriyal said that the ministry of hrd is extremely proud of its all institutions, researchers , academicians , faculty members and students who are working tirelessly in the time of complete lockdown , to bring out solutions to the problems arising out of outbreak of covid 19 and which are being faced not only within the country but by the entire humanity . & 72 & 6 \\ \hline
\end{tabularx}
\end{table}

As observed in Table \ref{tab:examplePCA1}, sentences with low LALITA scores (e.g., -0.81) are short and simple (``state legislation enactments" with 3 tokens and 0 verbs), while those with increasing scores (e.g., 0.51 and 1) exhibit significantly greater length and verb count, along with more intricate grammatical structures. This confirms the LALITA score as a reliable marker of sentence complexity.

\section{Experimental Design and Data Curation Strategies}
\subsection{Clustering Sentences by LALITA Score}
\begin{figure}
    \centering
  \includegraphics[width=0.6\linewidth]{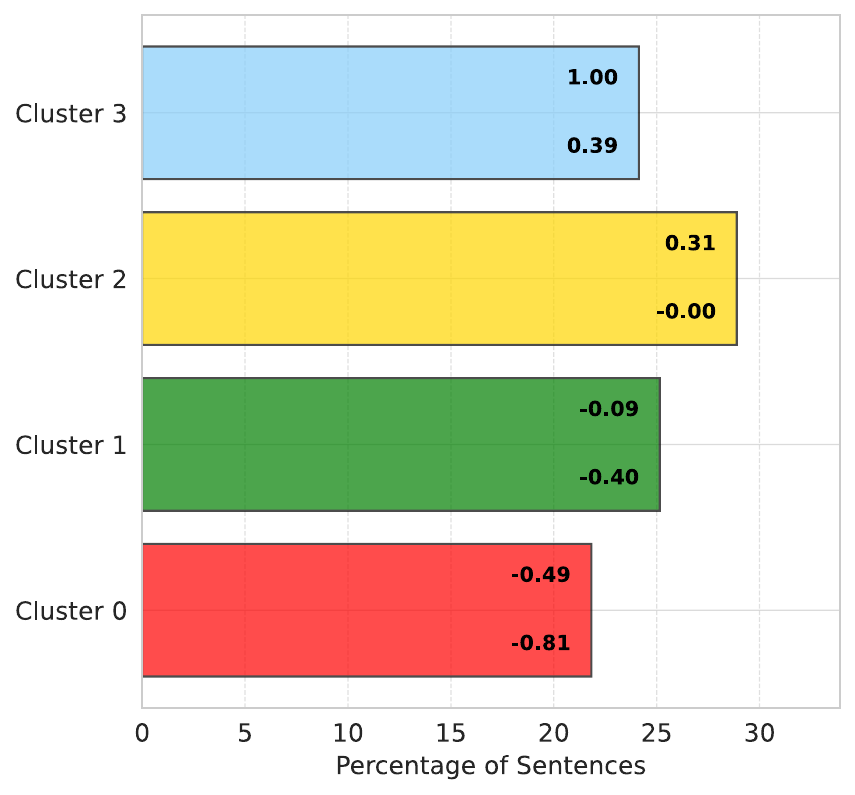}
  \caption{Training data cluster wise distribution along with boundary \(X_{PCA1}\) Values}
\label{fig:trainingDataDistribution}
\end{figure}

\begin{figure}[!h]
\includegraphics[width=0.9\textwidth]{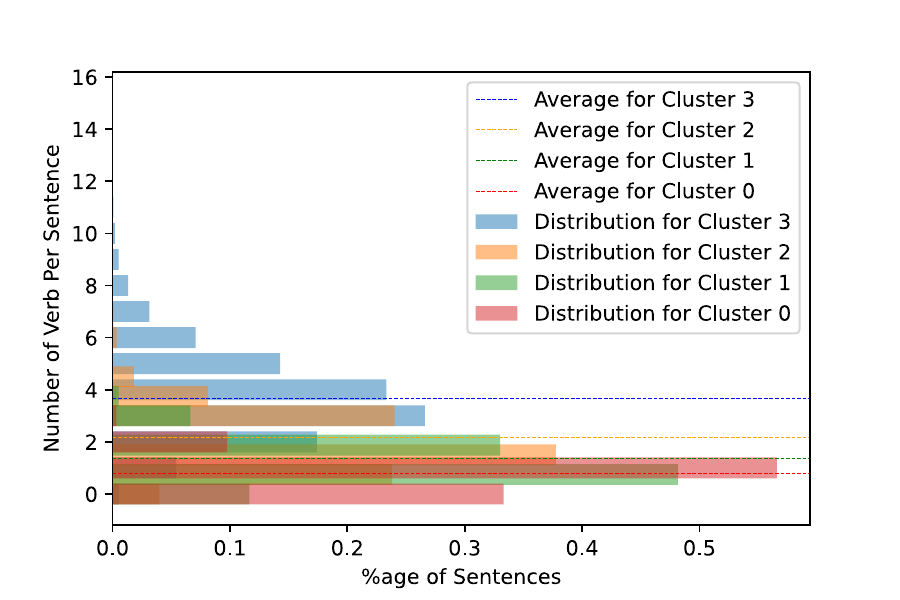}

\caption{Percentage wise Sentence distribution based on the verb count per sentence for each Cluster}
\label{fig:Sentence_Distribution_Dep_tokens}
\end{figure}

\begin{figure}[!h]
\includegraphics[width=0.9\textwidth]{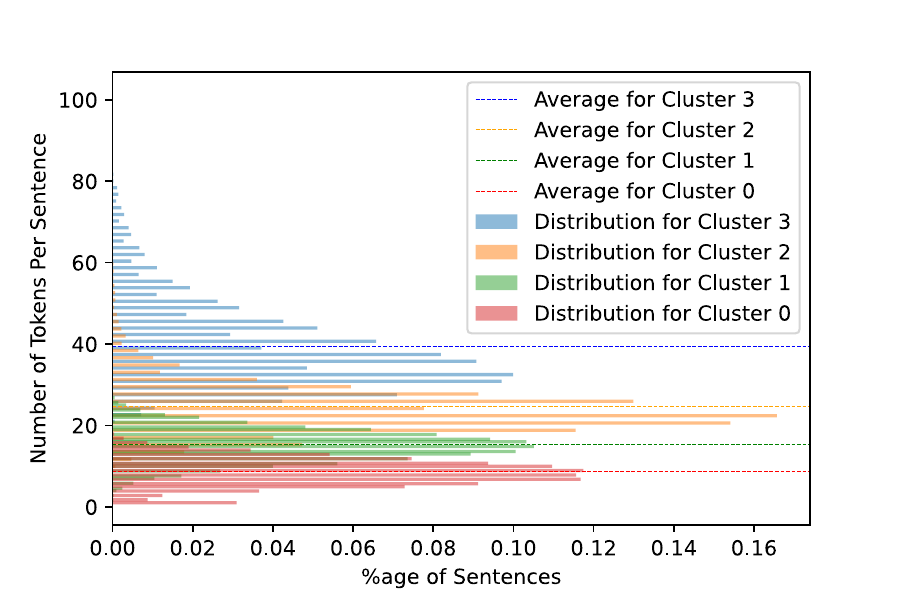}

\caption{Percentage wise Sentence distribution based on the token count per sentence for each Cluster}
\label{fig:Sentence_Distribution_Dep_verbs}
\end{figure}

\begin{figure}
    \centering
  \includegraphics[width=0.9\linewidth]{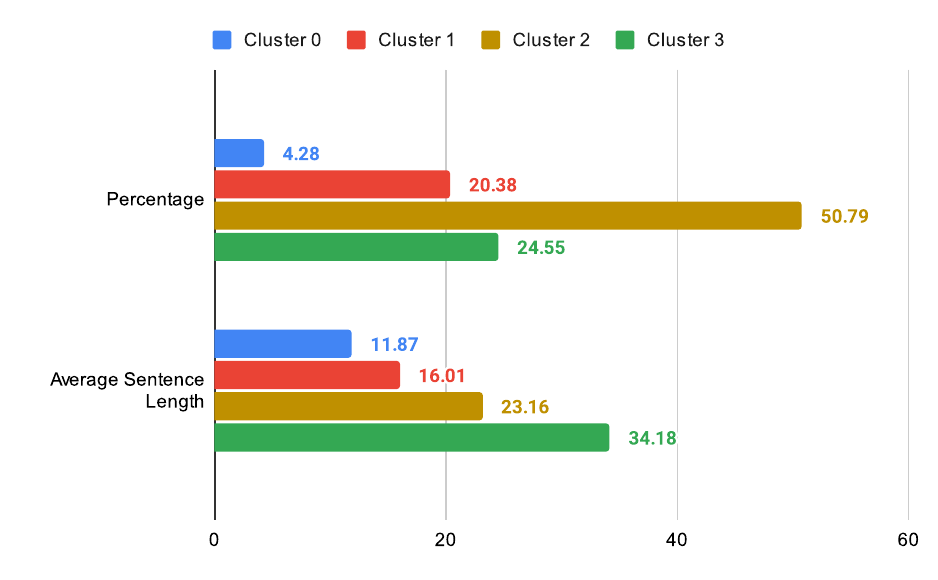}
\caption{Cluster wise distribution and average English sentence length for FLORES Testset}
\label{fig:flores}
\end{figure}

To effectively utilize the LALITA score for data curation, the sentences are grouped into distinct clusters based on their LALITA score (PCA1 values). The Fisher-Jenks algorithm \cite{Fisher1958OnGF} is selected for this clustering task. This statistical method is specifically designed to classify univariate numerical data into contiguous intervals, optimising for maximum homogeneity within each interval while minimizing variance between intervals. The application of Fisher-Jenks successfully divides the \(X_{PCA1}\)\footnote{We will be using PCA1 and \(X_{PCA1}\) interchangeably} values of the sentence feature vectors into four distinct clusters, as shown in Figure \ref{fig:trainingDataDistribution}. The quality of this clustering is assessed using the silhouette score, yielding a value of 0.54, which indicates good separation and cohesion of the clusters.
To confirm the robustness of this method, other clustering options, including k-means and Gaussian Mixture Models (GMM), are applied to \(X_{PCA2}\) through \(X_{PCA10}\) (additional principal components). Nonetheless, these alternative approaches consistently yield substantially lower silhouette scores (below 0.1), reinforcing the superiority of clustering based on the \(X_{PCA1}\).
Following the clustering of the Filtered English-Hindi Parallel Sentences (FPS), an analysis is conducted on the distribution of source sentences across four clusters, as illustrated in Figure 10. Distinct boundaries are observed between the clusters, with the majority of sentences residing in Cluster 2. A progressive increase in structural complexity among the clusters with LALITA score:
\begin{itemize}
    \item Cluster 0: Contains the simplest sentences.
\item Cluster 1: Features sentences that are more complex than those in Cluster 0.
\item Cluster 2: Contains the largest share of sentences, and the sentences exhibit greater complexity compared to those in Cluster 1.
\item Cluster 3: Comprises the most structurally complex sentences.
\end{itemize}
To further corroborate the structural intricacy within each cluster, the feature distributions associated with complexity, such as the number of verbs and sentence length, are analyzed for each cluster (Figures \ref{fig:Sentence_Distribution_Dep_verbs}, \ref{fig:Sentence_Distribution_Dep_tokens} respectively). This examination confirms that Cluster 3, for example, comprises a higher proportion of sentences featuring more verbs, clausal modifiers (acl), adverbial modifiers (advcl), and typically longer sentences. On the other hand, Cluster 0 is distinguished by the shortest sentences and a significant percentage of sentences with low values for these complexity indicators. These findings decisively reaffirm that the range of LALITA scores (PCA1 values) and the resulting clusters directly align with the sentence's structural complexity. This granular control over complexity, accomplished by clustering the LALITA score into distinct bins, offers a robust mechanism for precisely adjusting the complexity profile of the training data, advancing from simple filtering to actively crafting datasets with desired linguistic attributes.
\paragraph{What about cluster wise distribution in FLORES?}
\label{para:flores}
We analyse the cluster-wise distribution of English sentences in the FLORES test set. Figure \ref{fig:flores} shows that the sentences in FLORES include a significant number of complex sentences. However, this prompts the question of the nature of the training data, which exhibits a more ``uniform distribution". This offers insight into the disparities between the training and test distributions. It doesn't imply that a distribution similar to Figure \ref{fig:flores} leads to optimum performance, as the system with a comparable distribution in our configuration (\textit{10\_10\_40\_40}) does not perform as effectively as other configurations (Figure \ref{fig:allClusters}). It simply signifies that more consideration is required for the complexity profile of the training data for an effective translation system.
\subsection{Preliminary Experiments: LALITA Score as a Selection Criterion}
Preliminary experiments are conducted to assess the effectiveness of the LALITA score (PCA1) as a criterion for sentence selection. In these experiments, 300K sentence pairs are incrementally added using three strategies:
\begin{itemize}
    \item Increasing LALITA score (IncPCA)
    \item Decreasing LALITA score (DecPCA)
    \item Random sampling (RS) - average of two runs.
\end{itemize}
\begin{figure}
    \centering
  \includegraphics[width=1\linewidth]{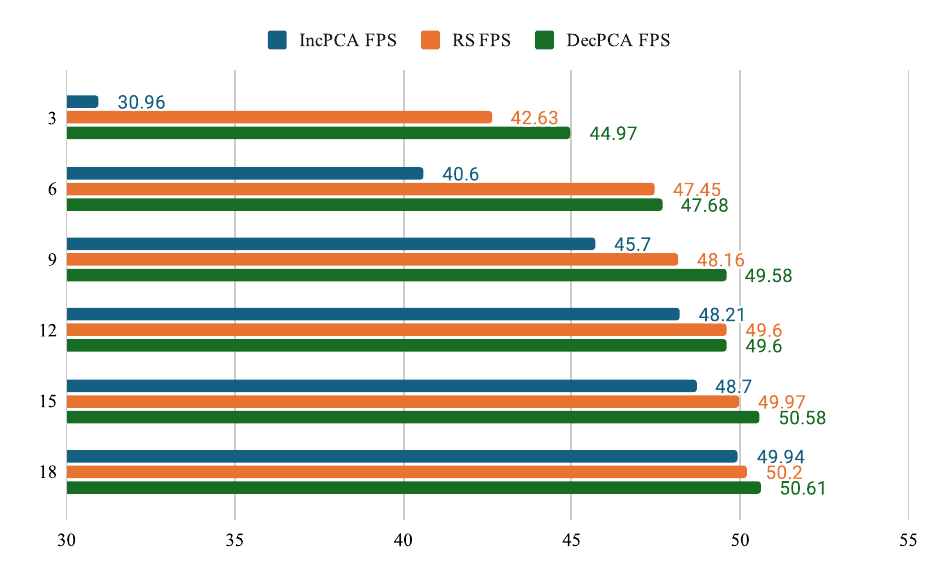}
  
\caption{CHRF++ scores (X-axis) on the FLORES test after adding 300K sentences based in step wise manner based on PCA1 value (Y-axis)}
\label{fig:stepWise}
\end{figure}
Figure \ref{fig:stepWise} illustrates a critical analysis of performance, demonstrating that initially choosing simple sentences (IncPCA) results in the MT system consistently falling behind as the training data set expands. Random sampling (RS) typically outperforms IncPCA. Importantly, starting with complex sentences (DecPCA) achieves strong early performance but reaches a plateau after approximately 900K sentences, suggesting that adding simpler sentences afterward does not substantially boost MT learning. This finding implies that simpler sentences contribute less over time to MT model learning, challenging the assumption that more data invariably leads to better results, regardless of complexity.

\subsection{Experimental Configurations for Data Selection}
Building on these preliminary findings, a comprehensive experimental design is developed using the four LALITA score-based clusters \ref{fig:trainingDataDistribution}. In this setup, a configuration set, represented as $<$a,b,c,d$>$, defines the percentage of sentences originating from various clusters. This implies that each configuration set results in multiple permuted configurations (a\_b\_c\_d, d\_c\_a\_b,...). For instance, the \textit{configuration set} $<$60,20,20,0$>$ generates up to 12 potential \textit{configurations} such as (\textit{60\_20\_20\_0}, \textit{60\_20\_0\_20}, \textit{60\_0\_20\_20}, etc.). Therefore, a \textit{configuration} \textit{a\_b\_c\_d} indicates that the chosen training data comprises \textit{a}\% of \(x\) from Cluster 0, \textit{b}\% from Cluster 1, \textit{c}\% from Cluster 2, and \textit{d}\% from Cluster 3, where \{a, b, c, d $\in 0, 10, 15, 20, 25, 33.34, 40, 50, 60, 70, 75, 100$\} and $a+b+c+d=100$. Consequently, training configurations vary from a uniform distribution of sentences (\textit{25\_25\_25\_25}) to those drawn completely from a single cluster (\textit{0\_0\_0\_100}). Table \ref{tab:config} presents the comprehensive list of configuration sets under consideration.
When selecting sentence pairs from any cluster, whether from FPS or synthetic sources, the bitext for those clusters is organized in descending order according to the LALITA score. As we sample \textit{m} sentence pairs, the top \textit{m} bitext from the respective group is chosen.
The experiments are conducted over various total dataset sizes (TDS), encompassing 50K, 100K, 200K, 400K, and 800K sentence pairs. Consequently, a total of 65 unique configuration sets are considered as shown in Table \ref{tab:config}, resulting in the evaluation of 325 distinct MT systems. For comparison, two baseline systems are established: \textit{RS} (Randomly Sampled), where 3 systems are trained on independently random samples from FPS, and \textit{baselineP} (Proportional), which samples sentences from each cluster in the same proportion as found in the original FPS dataset (21.83\% from Cluster 0, 25.15\% from Cluster 1, 28.89\% from Cluster 2, and 24.13\% from Cluster 3).

\begin{table}[h!]
\centering
\caption{Configuration of Percentage of sentences sampled from each cluster, with 3 systems trained on independently random sampled sentences from FPS and one system trained with baselineP.}\label{tab:config}
\begin{tabular}{|c|c|}
\hline
\begin{tabular}[c]{@{}c@{}}Configuration set\end{tabular} & \begin{tabular}[c]{@{}c@{}}Total Number of \\ configurations\end{tabular} \\ \hline
\textless{}25,25,25,25\textgreater{}         & 1  \\ \hline
\textless{}70,10,10,10\textgreater{}         & 4  \\ \hline
\textless{}40,40,10,10\textgreater{}         & 6  \\ \hline
\textless{}33.34,33.34,33.34,0\textgreater{} & 4  \\ \hline
\textless{}60,20,20,0\textgreater{}          & 12 \\ \hline
\textless{}70,15,15,0\textgreater{}          & 12 \\ \hline
\textless{}50,50,0,0\textgreater{}           & 6  \\ \hline
\textless{}75,25,0,0\textgreater{}           & 12 \\ \hline
\textless{}100,0,0,0\textgreater{}           & 4  \\ \hline
Randomly Sampled                             & 3  \\ \hline
baselineP                                    & 1  \\ \hline
Total                                        & 65 \\ \hline
\end{tabular}
\end{table}


  

\subsection{Utility in Data Augmentation: Filling Complexity Gaps}
\label{subsec:dataAug}
An essential component of the experimental design is the data augmentation strategy. In various dataset sizes and configurations, such as configuration \textit{0\_20\_20\_60} for an 800K dataset size, Cluster 3 is unable to supply 640K sentences. In these situations, it becomes necessary to augment the data.
The decision focuses on expanding within the deficient cluster instead of sourcing the remaining sentences from other clusters with lower complexity. This choice is crucial as it preserves the intended structural diversity and complexity profile of the particular configuration, ensuring that the experiment aligns with its objective of assessing the effect of targeted complexity.
Synthetic parallel sentences are generated to support this augmentation. A Hindi-to-English MT system trains on the Samanantar dataset, and Hindi monolingual text \cite{goldhahn-etal-2012-building} is then back-translated \footnote{Beam Size is 5 for all synthetic data generation.} \cite{sennrich-etal-2016-improving} to create synthetic English-Hindi parallel sentences. Backtranslation is specifically selected over forward translation to ensure that the fluency of the target language (Hindi) remains unaffected by the augmentation process \cite{burlot-yvon-2018-using}. We acknowledge that the quality of these synthetic data hinges on the base system utilized for back-translation. Nonetheless, we use this as a proxy to demonstrate how system performance would change if the necessary data were actually available in the respective clusters.
Once synthetic English is generated, the steps to insure quality are as follows:
\begin{itemize}
\item \textbf{Log-likelihood threshold}: We ensure that the synthetic English sentences exhibit an average log-likelihood of $-1.0$ or higher, according to measurements taken by the \href{https://fairseq.readthedocs.io/en/latest/getting_started.html}\textit{fairseq-interactive} tool. This procedure is vital for eliminating excessively noisy sentences.
\item \textbf{Single-sentence selection}: We choose a sentence pair only when the produced English text consists of a single complete sentence.
\end{itemize}
The English side of this newly generated synthetic corpus is then clustered employing the same LALITA score method, leading to the formation of four synthetic clusters ($C^l_0$ to $C^l_3$). If a configuration calls for `m' sentences from an FPS cluster $C^l_i$ but only `n' are present, the deficit of `m-n' sentences is compensated by sampling synthetic parallel sentences with the highest LALITA scores from the corresponding synthetic cluster $C^l_i$. This intelligent augmentation strategy ensures that the augmented data precisely aligns with the desired linguistic profile, transforming data augmentation from a brute-force method into a refined tool for tackling specific representational gaps in training data.

We ensure clarity and conciseness by displaying and discussing solely the configurations for each experiment with respect to each dataset size, as illustrated in Table \ref{tab:selected_config}. 
\begin{table}[]
\caption{Configurations being analysed}
\label{tab:selected_config}
\begin{tabular}{c|c|c|c}
\hline
\textbf{Baseline} & \textbf{\begin{tabular}[c]{@{}c@{}}Worst Performing\\ Systems\end{tabular}} & \textbf{\begin{tabular}[c]{@{}c@{}}Using Single\\ Cluster\end{tabular}} & \textbf{\begin{tabular}[c]{@{}c@{}}Best Performing\\ Systems\end{tabular}} \\ \hline
baselineP & 100\_0\_0\_0 & 100\_0\_0\_0 & 0\_20\_20\_60 \\
\multirow{2}{*}{\begin{tabular}[c]{@{}c@{}}RS (average of\\ 3 runs)\end{tabular}} & 75\_25\_0\_0 & 0\_100\_0\_0 & 15\_0\_15\_70 \\
 & 50\_50\_0\_0 & 0\_0\_100\_0 & 0\_25\_0\_75 \\
 & 70\_15\_15\_0 & 0\_0\_0\_100 & 0\_0\_25\_75 \\
 & 60\_20\_20\_0 &  & 0\_0\_0\_100 \\ \hline
\end{tabular}
\end{table}
The performance for the remaining configurations is presented in the Appendix \ref{sec:configAll}.

\section{Results and Discussion}
\subsection{Performance on English-Hindi Systems}
\begin{figure}
    \centering
  \includegraphics[width=1\linewidth]{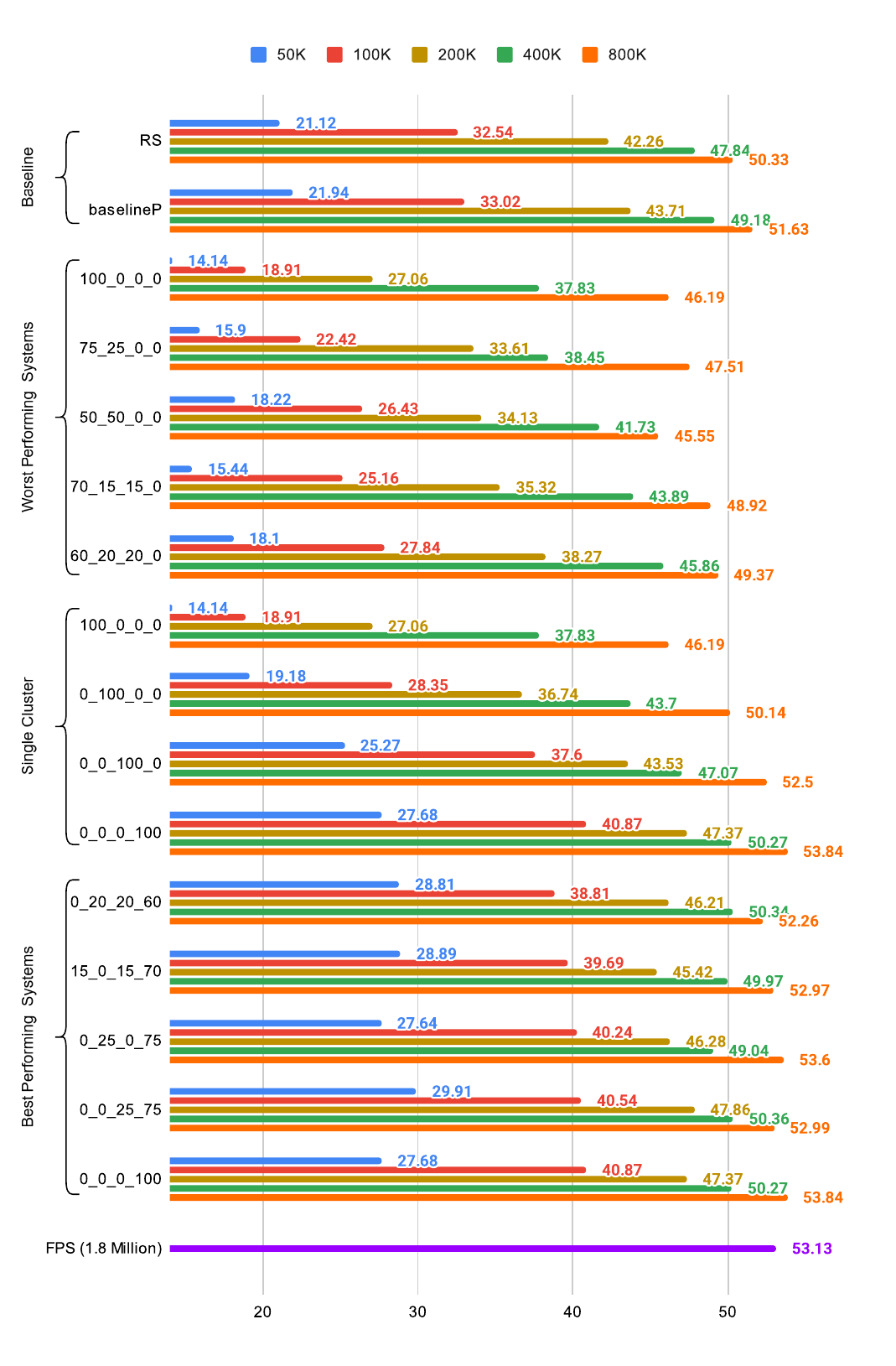}
  
\caption{CHRF++ scores in the FLORES test for English-Hindi systems.}
\label{fig:final_1}
\end{figure}

The experimental results for English-Hindi Machine Translation systems clearly show that optimal performance occurs when the training data predominantly features structurally complex sentences, particularly those from Cluster 3. Figure \ref{fig:final_1}, which displays the CHRF++ scores on the FLORES test set for different English-Hindi systems, strongly supports this observation. Systems with a high percentage of Cluster 3 sentences consistently and significantly outperform other systems, including baseline systems (RS and baselineP). Notably, a system trained with only 800K complex sentences using the 0\_0\_0\_100 configuration (meaning 100\% from Cluster 3, supplemented with synthetic data if needed) achieves results comparable to a system trained on the full 1.8 million sentence FPS dataset. This represents a substantial 44\% reduction in data requirements while maintaining or slightly enhancing translation quality. These findings highlight a practical \textit{efficiency frontier} in MT training- similar or improved performance is possible with much less data, provided that sentence selection is carried out strategically.

Additional investigation of systems trained exclusively on one cluster (e.g., 100\_0\_0\_0 for Cluster 0 versus 0\_0\_0\_100 for Cluster 3) indicates that systems focused solely on sentences from Cluster 3 consistently surpass those trained with sentences from other separate clusters, regardless of dataset size. On the other hand, the systems that exhibit low performance are those having few or no sentences from Cluster 3. Taken together, these results not only demonstrate the importance of complex sentences but also reinforce the central theme of this work- effective corpus creation hinges on strategic sentence selection to achieve cost-efficient training while ensuring robust downstream performance.

\textbf{Total Tokens:}
Table \ref{tab:totalTokens_enhi_short} displays the total tokens for each training dataset depicted in Figure \ref{fig:final_1}. A crucial observation is that our top-performing curated systems utilize roughly 44\% fewer sentence pairs and about 64\% fewer total tokens compared to the full parallel sentence (FPS) corpus. This indicates that our approach successfully curates a more token-dense dataset, delivering a focused training signal within a smaller corpus. The total tokens for all experiments appear in the Appendix \ref{appsec:totalTokens}.
\begin{table}[h!]
\caption{Total Tokens (in Millions) for each training data set size and configuration. The last row contains English (Eng) and Hindi (Hin) tokens for Full Parallel Sentences. Best Systems are marked in Bold}\label{tab:totalTokens_enhi_short}
\begin{tabularx}{\textwidth}{c*{5}{|YY}}
\hline
\textbf{Data Set Size} & \multicolumn{2}{>{\centering\arraybackslash}X|}{\textbf{50K}} & \multicolumn{2}{>{\centering\arraybackslash}X|}{\textbf{100K}} & \multicolumn{2}{>{\centering\arraybackslash}X|}{\textbf{200K}} & \multicolumn{2}{>{\centering\arraybackslash}X|}{\textbf{400K}} & \multicolumn{2}{>{\centering\arraybackslash}X}{\textbf{800K}} \\ \hline
\textbf{Language Pair} & \textbf{Eng} & \textbf{Hin} & \textbf{Eng} & \textbf{Hin} & \textbf{Eng} & \textbf{Hin} & \textbf{Eng} & \textbf{Hin} & \textbf{Eng} & \textbf{Hin} \\ \hline
\textbf{RS} & 1.36 & 1.46 & 2.73 & 2.94 & 5.46 & 5.88 & 10.91 & 11.74 & 21.84 & 23.5 \\
\textbf{baselineP} & 1.37 & 1.47 & 2.73 & 2.94 & 5.46 & 5.87 & 10.93 & 11.74 & 21.83 & 23.5 \\
\textbf{100\_0\_0\_0} & 0.74 & 0.87 & 1.43 & 1.68 & 2.74 & 3.25 & 5.54 & 6.42 & 10 & 11.76 \\
\textbf{0\_100\_0\_0} & 1.19 & 1.32 & 2.32 & 2.57 & 4.42 & 4.93 & 8.02 & 9.05 & 14.49 & 16.63 \\
\textbf{0\_0\_100\_0} & 1.77 & 1.86 & 3.49 & 3.68 & 6.76 & 7.15 & 12.68 & 13.52 & 23.27 & 25.42 \\
\textbf{0\_0\_0\_100} & 2.88 & 2.9 & 5.38 & 5.45 & 9.89 & 10.11 & 17.82 & 18.38 & \textbf{35.91} & \textbf{38.58} \\
\textbf{50\_50\_0\_0} & 0.98 & 1.11 & 1.93 & 2.19 & 3.75 & 4.26 & 7.16 & 8.17 & 13.56 & 15.47 \\
\textbf{75\_25\_0\_0} & 0.87 & 0.99 & 1.7 & 1.96 & 3.31 & 3.8 & 6.24 & 7.27 & 12.28 & 14.08 \\
\textbf{0\_25\_0\_75} & 2.52 & 2.56 & 4.76 & 4.86 & 8.88 & 9.15 & 16.32 & 16.96 & \textbf{31.78} & \textbf{33.88} \\
\textbf{0\_0\_25\_75} & 2.67 & 2.69 & 5.05 & 5.14 & 9.46 & 9.7 & 17.48 & 18.06 & \textbf{34.12} & \textbf{36.11} \\
\textbf{60\_20\_20\_0} & 1.05 & 1.17 & 2.08 & 2.33 & 4.08 & 4.56 & 7.9 & 8.87 & 15.64 & 17.36 \\
\textbf{0\_20\_20\_60} & 2.41 & 2.45 & 4.6 & 4.71 & 8.7 & 8.98 & 16.25 & 16.89 & \textbf{30.55} & \textbf{32.14} \\
\textbf{70\_15\_15\_0} & 0.98 & 1.1 & 1.93 & 2.18 & 3.78 & 4.25 & 7.23 & 8.22 & 14.35 & 16.09 \\
\textbf{15\_0\_15\_70} & 2.47 & 2.5 & 4.67 & 4.77 & 8.76 & 9.02 & 16.2 & 16.82 & \textbf{31.36} & \textbf{33.25} \\ \hline
\textbf{FPS} & 50.47 & 54.31 & & & & & & & & \\ \hline
\end{tabularx}
\end{table}

This finding may lead to the hypothesis that simply accumulating a large number of tokens by opting for lengthy sentences or using many tokens provides an adequate heuristic for effective data curation. However, existing research indicates that sentence length alone offers a weak signal for evaluating translation quality, structural complexity, or informativeness, and NMT systems generally perform poorly with sentences that are either excessively short or long \cite{wan-etal-2022-challenges}. Our subsequent ablation study, detailed next, supports this argument. A more effective strategy involves a comprehensive assessment that thoroughly evaluates linguistic and lexical elements rather than relying solely on sentence length.

\subsection{Ablation study}
\begin{figure}
\centering
\includegraphics[width=1\linewidth]{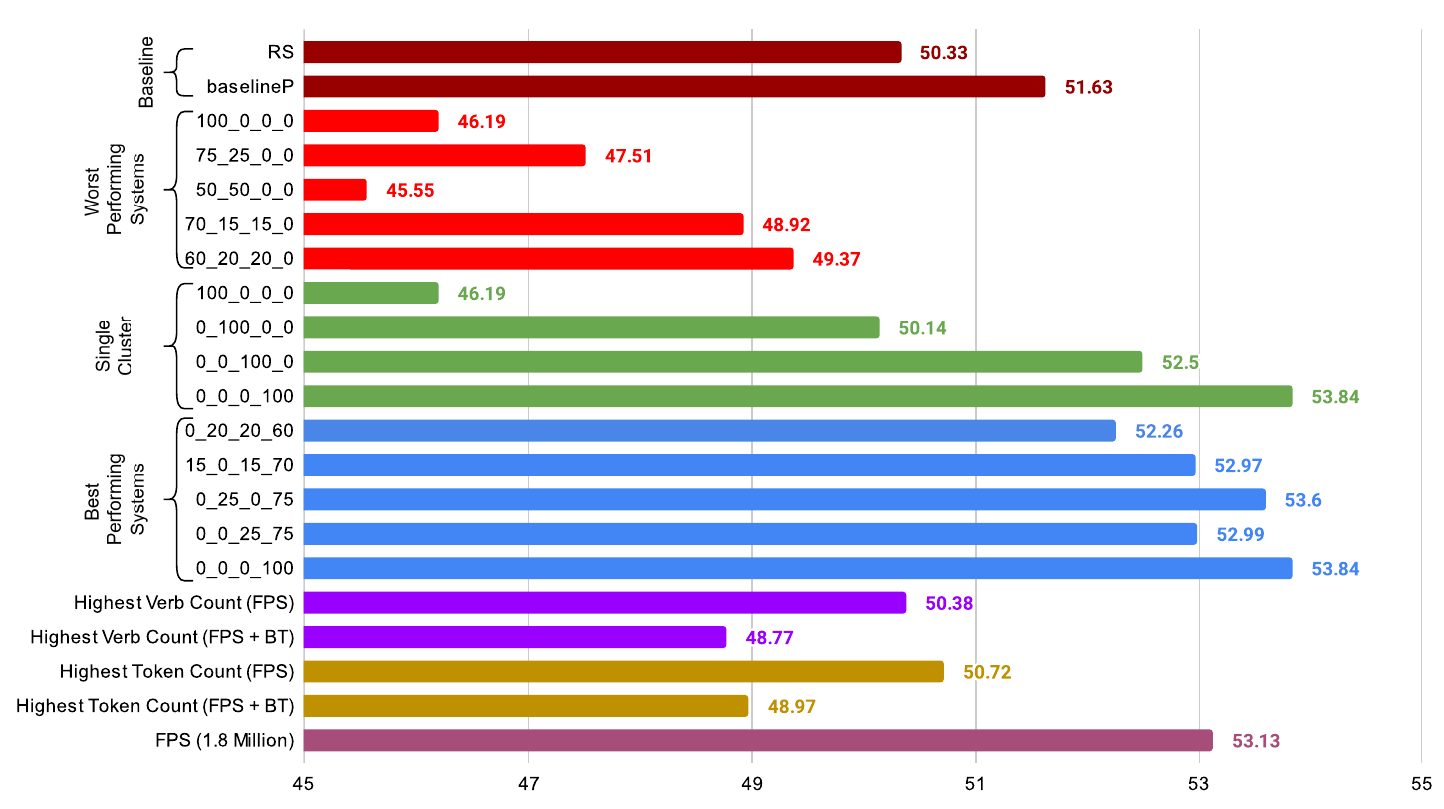}

\caption{CHRF++ scores in the FLORES test for English-Hindi systems with FPS and FPS+BT using sentence length and number of verbs (for 800K sentence pairs).}
\label{fig:ablation}
\end{figure}

A performance comparison reveals that LALITA surpasses selection strategies that depend on individual features, such as the highest verb count or the longest sentences.
Figure \ref{fig:ablation} shows that selecting based on a single feature, such as sentence length or verb count, does not reach the effectiveness of LALITA at processing 800K sentences. To investigate further, we conduct two forms of data curation (for 800K sentences) for each single-feature method: the first uses only filtered parallel sentences (FPS), and the second combines FPS with synthetic data. Our findings indicate a noticeable performance decline when synthetic data is selected using a single-feature heuristic, showing that this method often fails to identify high-quality augmented sentences. This underscores the advantage of a comprehensive strategy, one that integrates various linguistic and lexical features through the LALITA score. Since the LALITA score is calculated from the complete feature set of the FPS, it captures the typical value range for source sentences, offering a more reliable metric for choosing synthetic sentences. This establishes LALITA as not just a tool for providing a solid heuristic for data augmentation but also as a promoter of more sustainable and resource-efficient  progress in natural language processing.
\subsection{Examples of Translation Performance}
\begin{figure}
    \centering
  \includegraphics[width=1\linewidth]{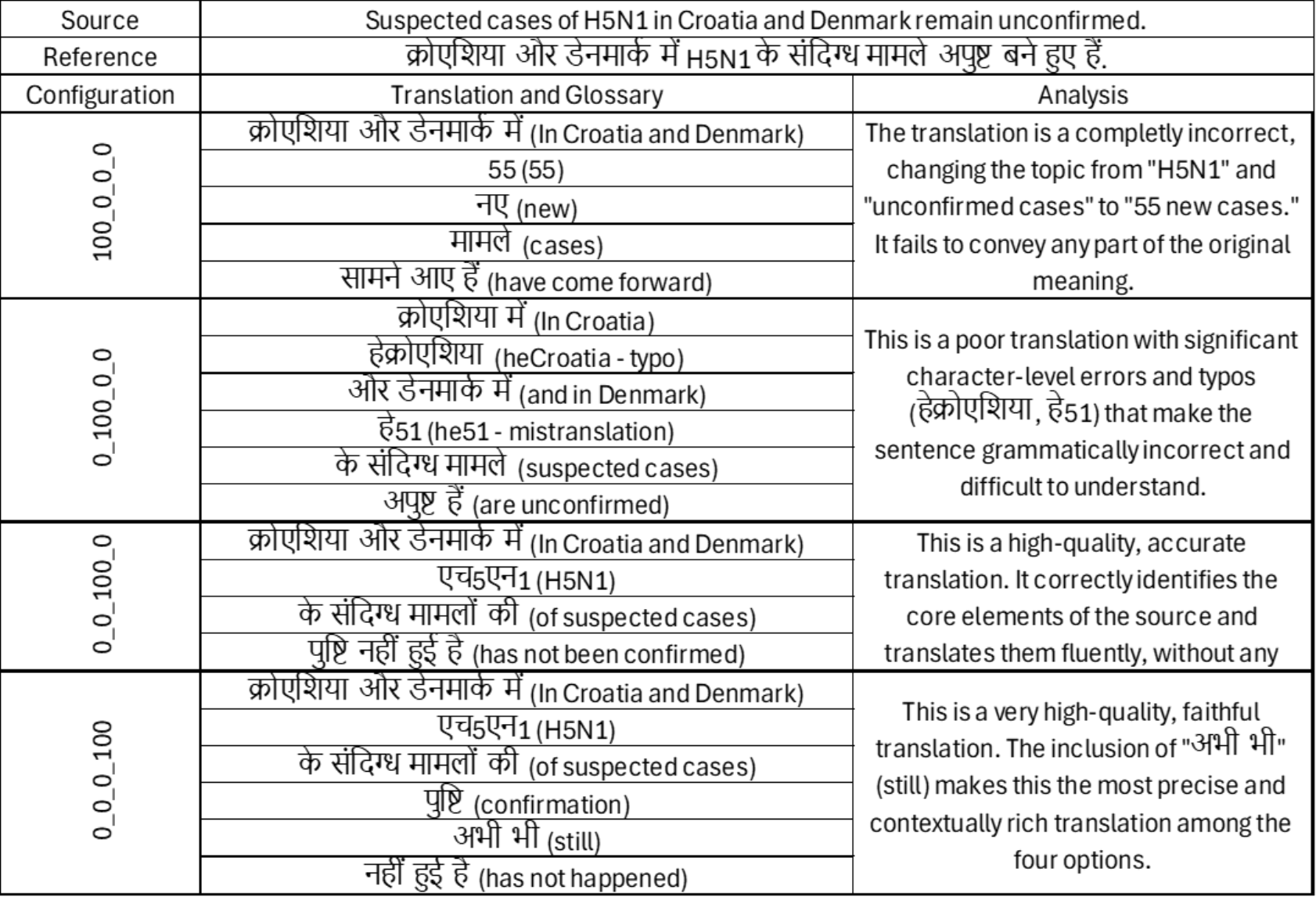}
  
\caption{Translation Example of a Test input belonging to cluster 0.}
\label{fig:s0}
\end{figure}

\begin{figure}
    \centering
  \includegraphics[width=1\linewidth]{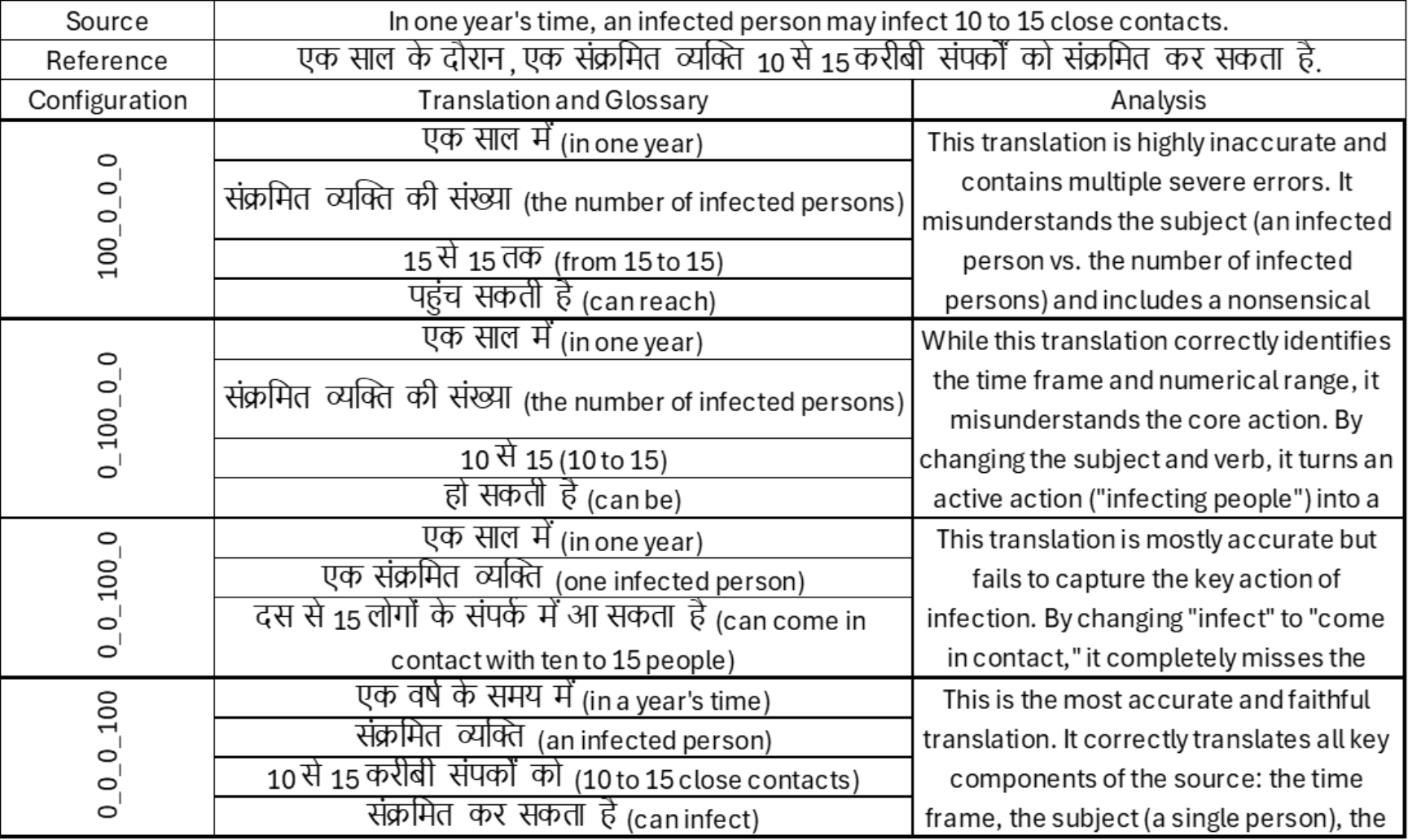}
  
\caption{Translation Example of a Test input belonging to cluster 1.}
\label{fig:s1}
\end{figure}

\begin{figure}
    \centering
  \includegraphics[width=1\linewidth]{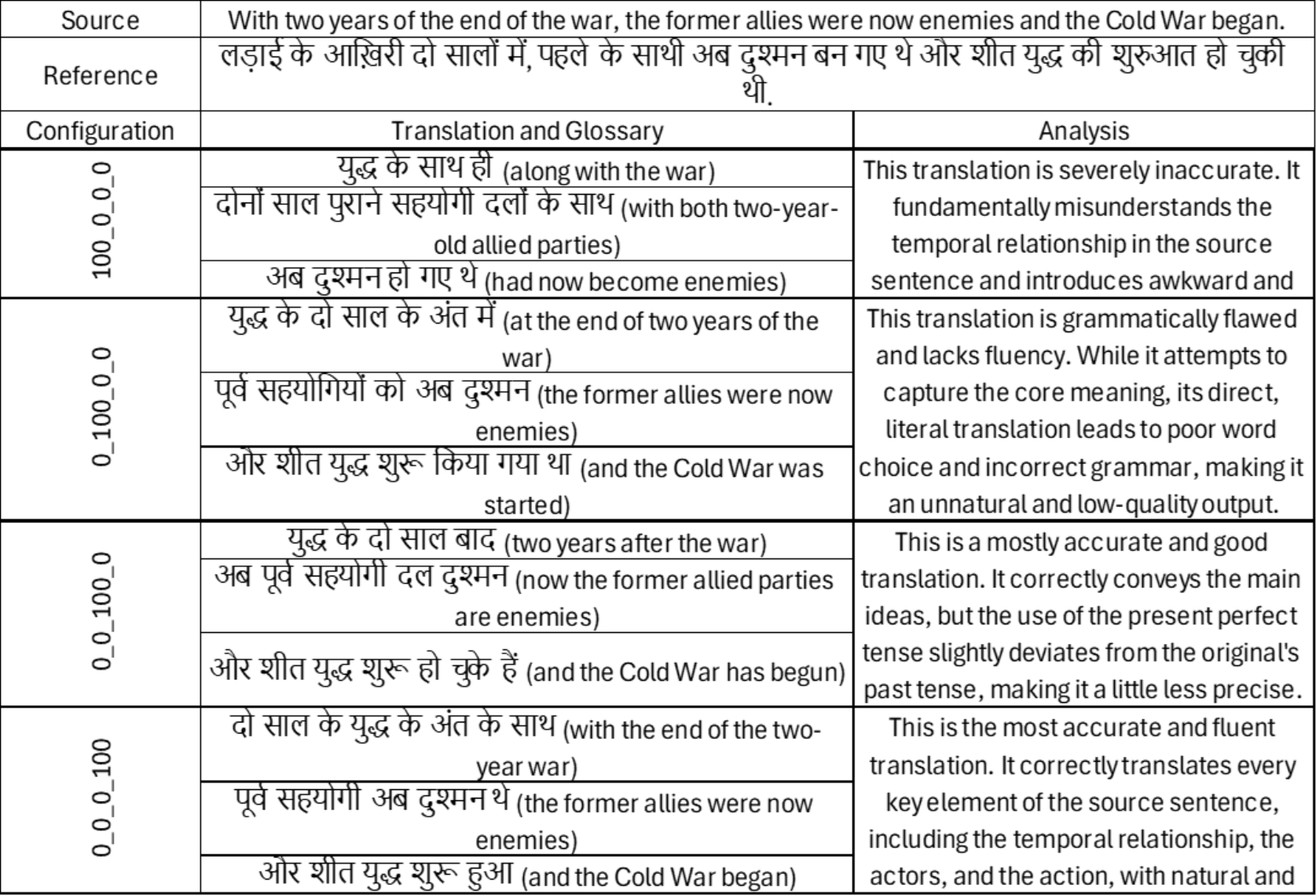}
  
\caption{Translation Example of a Test input belonging to cluster 2.}
\label{fig:s2}
\end{figure}

\begin{figure}
    \centering
  \includegraphics[width=1\linewidth]{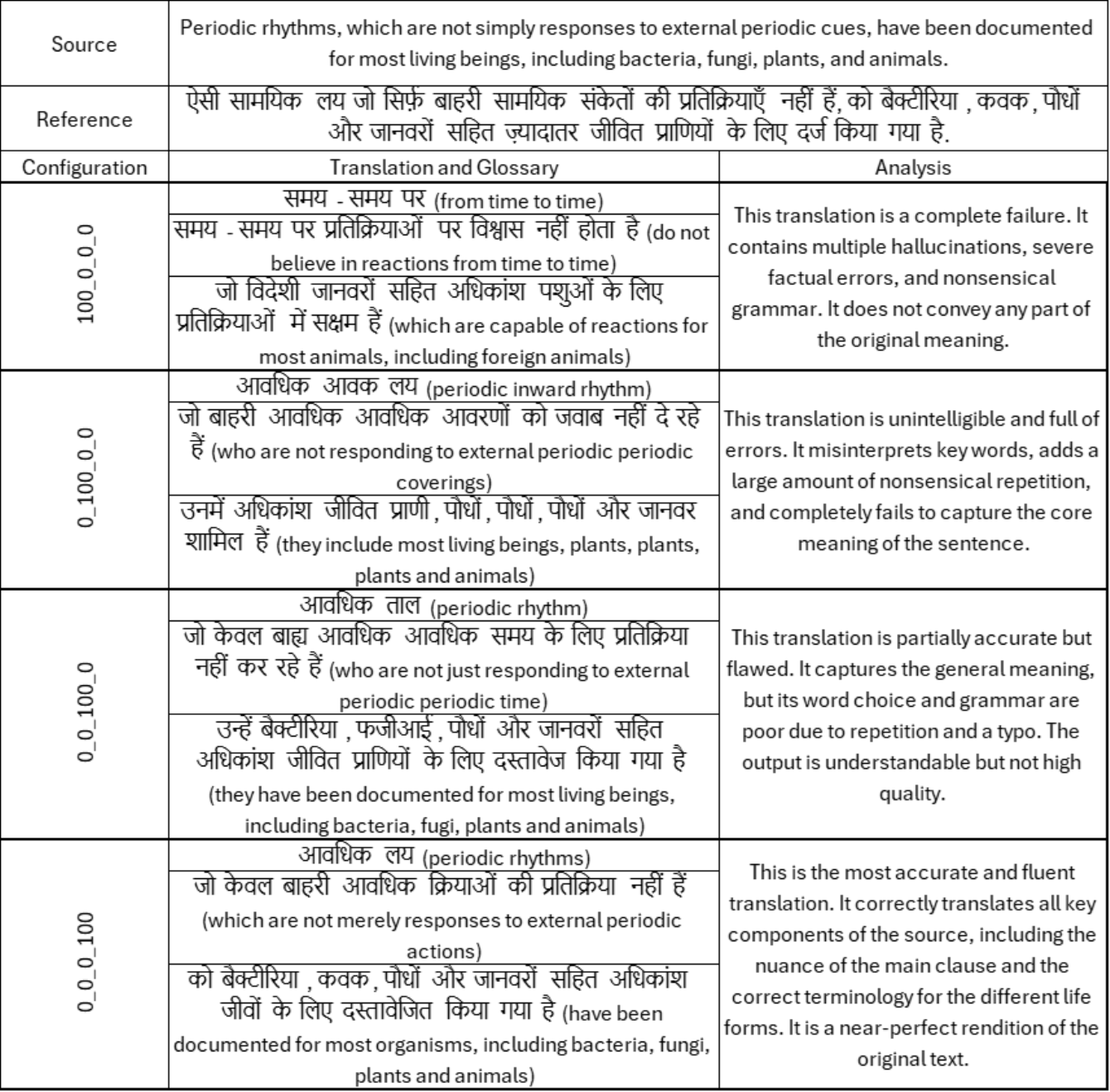}
  
\caption{Translation Example of a Test input belonging to cluster 3.}
\label{fig:s3}
\end{figure}

To further demonstrate the variations in translation quality, Figures \ref{fig:s0}, \ref{fig:s1}, \ref{fig:s2}, and \ref{fig:s3} present translation outputs and corresponding glossaries from four systems, each trained using a distinct \textbf{200K-sentence dataset} derived from one of four clusters. These figures illustrate test inputs from clusters 0, 1, 2, and 3 of FLORES, respectively.
The outputs from models trained on the simplest sentences (Cluster 0) and moderately simple sentences (Cluster 1) consistently show significant inaccuracies and generally fail to preserve the original meaning. Translations from these systems contain numerous severe errors, such as factual mistakes, nonsensical phrasing, typos, and grammatical flaws rendering them unintelligible. In stark contrast, models trained on more complex sentences from Cluster 2 and the most complex sentences from Cluster 3 produce progressively higher-quality translations. The translations from the Cluster 3 model are notably the most accurate and fluent, often capturing subtle nuances that other models overlook, demonstrating its ability to provide a near-perfect rendition of the original text. These qualitative examples compellingly demonstrate that using more complex sentences in a low-resource setting yields better results.


\subsection{Transferability to Other Low-Resource Languages (Odia, Nepali, Norwegian Nynorsk)}
The LALITA framework is primarily tested on English-Hindi but is also created for widespread applicability to other language pairs. To demonstrate the transferability of this framework and our English-Hindi configurations (illustrated in Figure \ref{fig:final_1}), we carry out further experiments on English-Odia, English-Nepali, and English-Norwegian Nynorsk.
The parallel corpora for these language pairs are sourced from the following:
\begin{itemize}
    \item English-Odia and English-Nepali: \citet{gala2023indictrans}
    \item English-Norwegian Nynorsk: \citet{gowda-etal-2021-many}
\end{itemize}
However, unlike the English-Hindi data, we only performed sentence-pair-wise deduplication on these datasets, as more rigorous cleaning resulted in a low amount of usable data.
Table \ref{tab:otherStats} provides the statistics for the parallel texts after deduplication.

\begin{figure}
    \centering
  \includegraphics[width=1\linewidth]{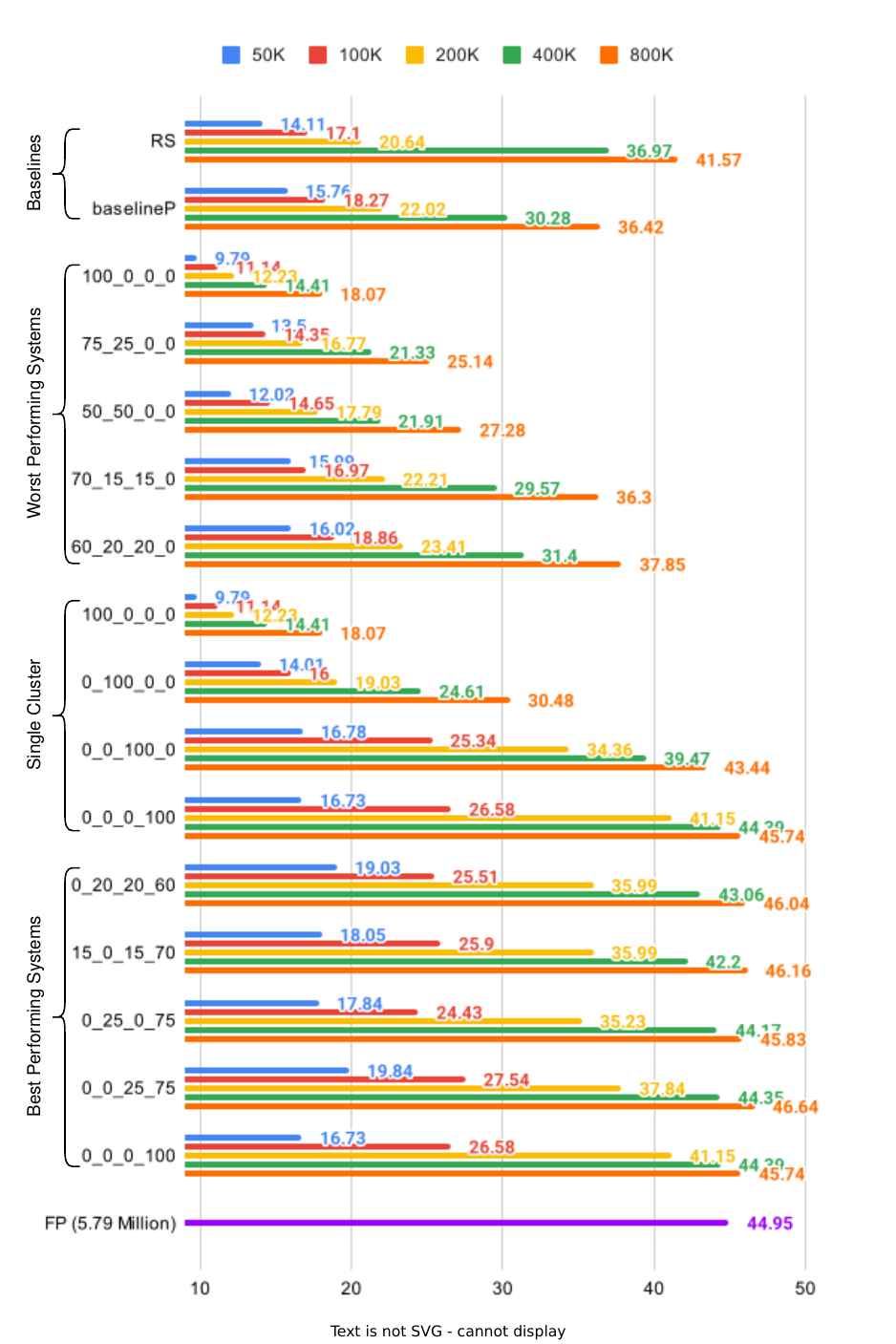}
  
\caption{CHRF++ scores in the FLORES test for English-Odia systems.}
\label{fig:ord}
\end{figure}

\begin{figure}
    \centering
  \includegraphics[width=1\linewidth]{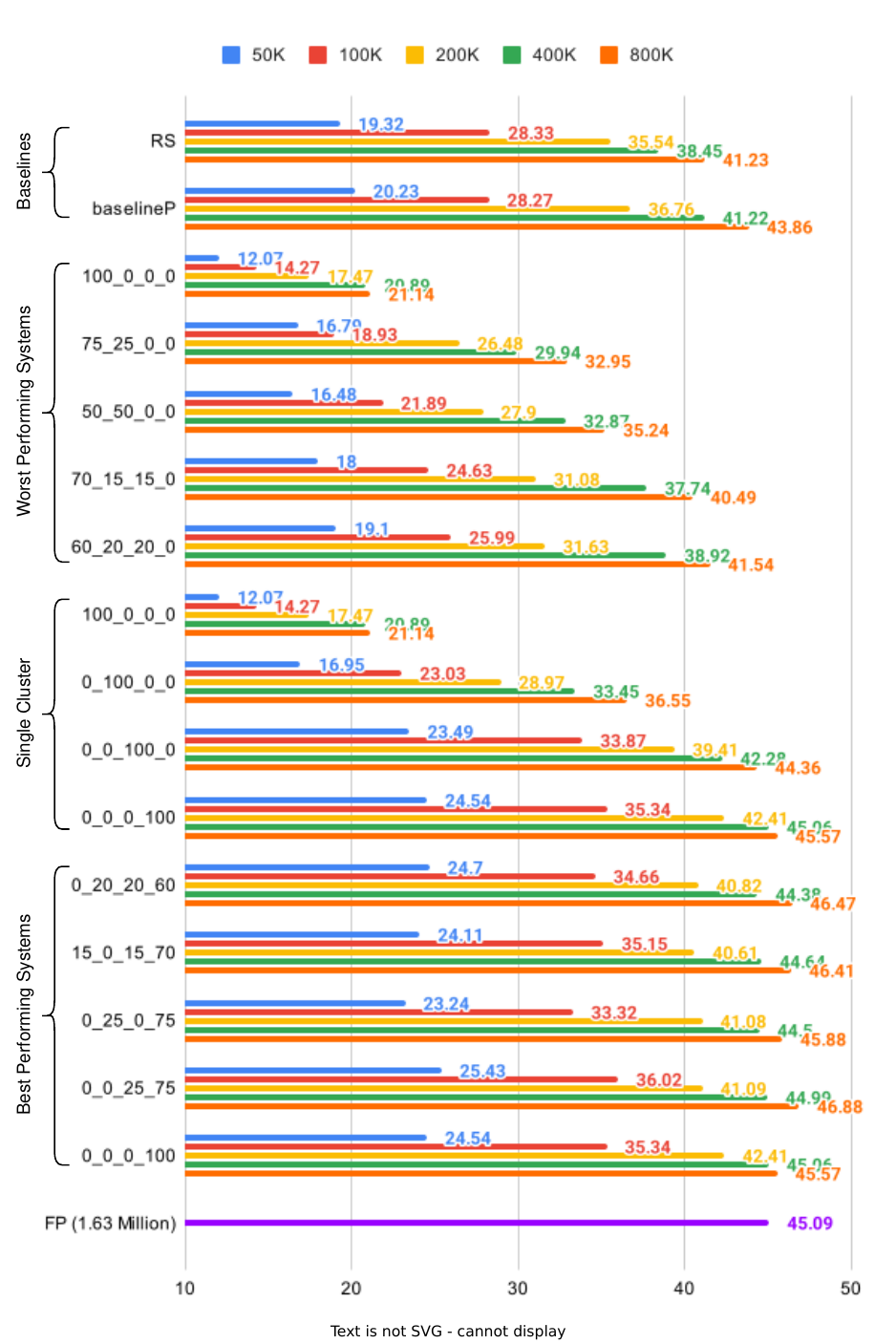}
  
\caption{CHRF++ scores in the FLORES test for English-Nepali systems.}
\label{fig:npi}
\end{figure}

\begin{figure}
    \centering
  \includegraphics[width=1\linewidth]{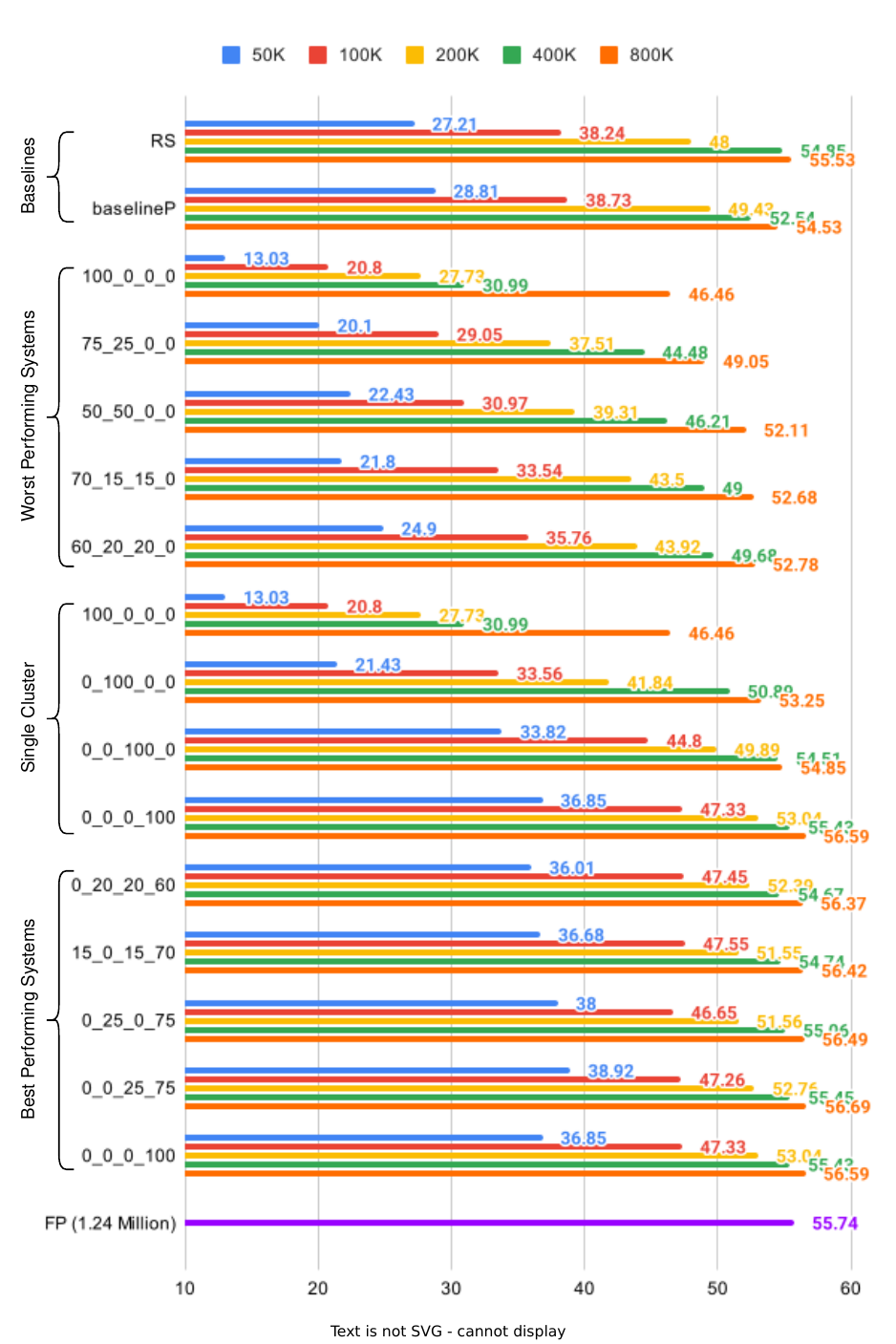}
  
\caption{CHRF++ scores in the FLORES test for English-Norwegian Nynorsk systems.}
\label{fig:nno}
\end{figure}

\begin{table}
\caption{Full Parallel (FP) for each language pair after deduplication and clustering (shown in percentage)}
\label{tab:otherStats}
\begin{tabularx}{\textwidth}{lXXXXX}
\hline
Language & Sentence pairs after preprocessing & Cluster 0 & Cluster 1 & Cluster 2 & Cluster 3 \\ \hline
Odia & 5790440 & 57.86 & 32.14 & 7.46 & 2.55 \\
Nepali & 1632700 & 45.08 & 24.51 & 19.8 & 10.61 \\
Norwegian Nynorsk & 1242791 & 37.72 & 24.35 & 25.44 & 12.49 \\ \hline
\end{tabularx}
\end{table}
Here we see that the cluster-wise distribution appears more skewed than in the English-Hindi case. We utilize monolingual data for Odia, Nepali, and Norwegian Nynorsk collected from the following sources:
\begin{itemize}
    \item \citet{doddapaneni-etal-2023-towards}
    \item \citet{goldhahn-etal-2012-building}
    \item \citet{ortiz-suarez-etal-2020-monolingual}
\end{itemize}
In a manner akin to the English-Hindi experiment, we filter and cluster the synthetic bitext (refer to Section \ref{subsec:dataAug}). The performance of the respective systems is given in Figures \ref{fig:ord}, \ref{fig:npi}, and \ref{fig:nno}.


These results, alongside the skewed sentence distribution across clusters shown in Table \ref{tab:otherStats}, indicate that using all available data might not be the best approach, especially if we look at sentence distribution across clusters for language pairs such as English-Odia. Instead, a strategic heuristic in curating and compiling parallel text can greatly decrease the overhead required in terms of training data to develop a quality translation system.
The consistent performance across these various language pairs indicates that the linguistic features and the notion of ``complexity" assessed by the LALITA score extend beyond English-Hindi, possessing a certain universality. This generalizability forms a key component of the framework’s usefulness, implying that LALITA can act as an essential tool for numerous language combinations, particularly those with limited data availability. In such low-resource scenarios, LALITA helps by enabling strategic selection of structurally complex and linguistically rich sentences, which provide the most informative training signals. This focused selection maximizes the utility of scarce parallel data, improving translation quality even with smaller corpora. Thus, LALITA expands the research's influence, establishing itself as a potentially significant tool that can be used for guiding efficient corpus creation and enhancing MT performance across a broad range of language pairs.
\subsection{Utility in High-Resource Language Data Reduction (English-German)}
\begin{figure}
    \centering
  \includegraphics[width=0.8\columnwidth]{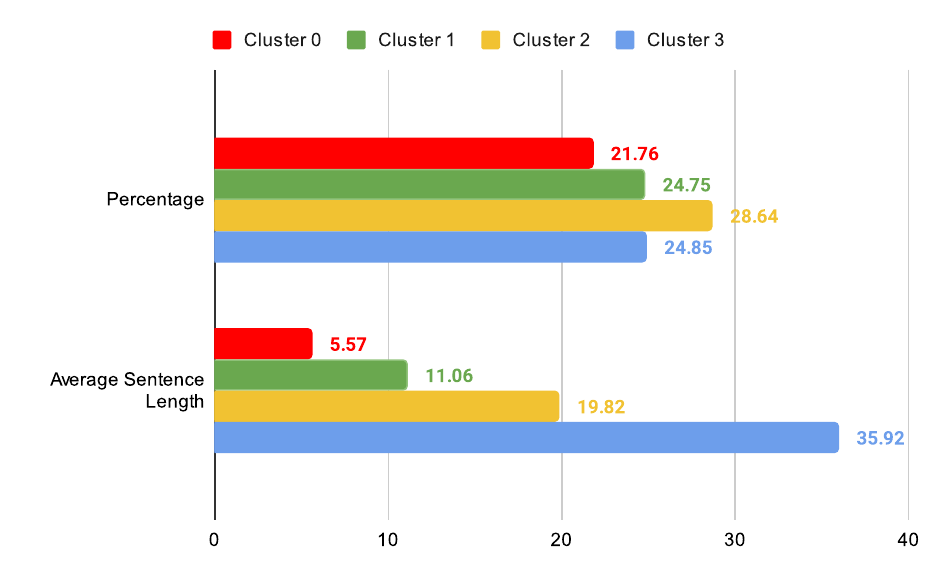}
\caption{Cluster Wise Distribution of Sentences in FPSG and average English sentence length in each cluster}
\label{fig:endeTrnDistribution}
\end{figure}

To establish the LALITA framework's effectiveness and transferability in high-resource scenarios, preliminary experiments focus on English-to-German translation, a language pair with abundant resources. A significant dataset of 20 million sentence pairs from the WMT24 Dataset (referred to as FPSG) undergoes testing for this revalidation. Through feature extraction and clustering with the LALITA score approach, the cluster-wise distribution appears similar (Figure \ref{fig:endeTrnDistribution}) to that observed for English-Hindi, as seen in Figure \ref{fig:trainingDataDistribution}. This consistency, particularly with Cluster 3 encompassing notably longer sentences, verifies that the LALITA score effectively captures complexity across various language pairs and data scales.

\begin{figure}
    \centering
  \includegraphics[width=0.8\columnwidth]{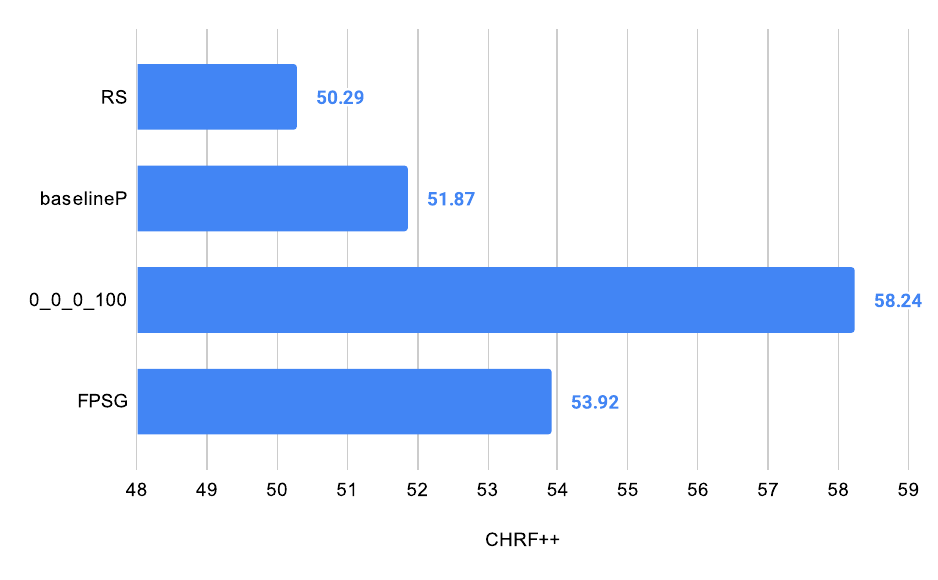}
\caption{CHRF++ scores in the FLORES test for English-German systems. FPSG is trained on 20 million Sentences, while rest are on 8 million.}
\label{fig:ende}
\end{figure}
In high-resource scenarios, our again focus on which type of synthetic sentences should be chosen for augmentation and which type should be utilized from FPSG. Building on insights from the English-Hindi experiments and due to resource constraints, four systems are trained for English-German, incorporating both FPSG and synthetic bitext:, \begin{itemize}
    \item 0\_0\_0\_100: This system employs 5.16 million sentences from Cluster 3 of FPSG, enhanced with 3.11 million synthetic sentence pairs from the corresponding synthetic Cluster 3 ($C^l_3$), generated using a German-to-English system trained on FPSG.
    \item RS (Randomly Sampled): This baseline involves randomly sampling 8.28 million sentences from FPSG for comparison.
    \item baselineP (Proportional Baseline): This system samples 8.28 million sentences from each cluster in the same proportion as present in the original FPSG dataset.
    \item FPS (Full Dataset): An MT system trained on the complete 20 million sentence FPSG dataset serves as the upper performance limit.
\end{itemize}
Figure \ref{fig:ende} shows the performance of the English-German systems. The MT system configured with more complex sentences (0\_0\_0\_100) consistently surpasses both the randomly sampled (RS) and proportional baseline (baselineP) systems. Notably, the LALITA-score-curated system (0\_0\_0\_100) also exceeds the performance of the system trained on the full 20 million sentence FPSG dataset, despite its larger size. This result demonstrates that the advantages of LALITA are not restricted to low-resource situations but apply equally in high-resource contexts. Consequently, LALITA can be leveraged to achieve exceptional performance with less data, even when substantial data is available, delivering universal efficiency benefits. This expands the practical and commercial potential of the method, indicating a more efficient data management strategy for MT development, potentially reducing computational costs and decreasing the environmental impact of large-scale MT models.

\subsection{A Bird's Eye View on Improvements}
Tables \ref{tab:comp_with_base} and \ref{tab:comp_with_fp} deliver a concise overview of our experimental results, presenting a ``bird's eye view" of the notable performance improvements realized via data curation when compared to both the baselines and full parallel datasets (FP).
\paragraph{Table \ref{tab:comp_with_base}: Comparison with Baseline Systems}
This table demonstrates the marked enhancement of systems trained on curated data compared to the top ``baseline" systems. The ``Accuracy with Humans" metric, as elucidated by \citet{kocmi-etal-2024-navigating}, evaluates the alignment of the system's metric-based improvements with human assessments. The findings suggest substantial performance gains across all language pairs when curated data is utilized. The most significant increase occurs in English-German, with a rise of 6.37 CHRF++ scores, underscoring our method's effectiveness even in high-resource contexts. For low-resource languages, English-Odia and English-Nepali, the curated systems exhibit considerable gains of 5.07 and 3.02, respectively, over their baselines. These enhancements underscore that an advanced data curation process can achieve notable enhancements in translation quality while reducing training demands.
\begin{table}
\caption{Comparison between the Best baseline System and the Best System on Curated Data, along with the Accuracy with Human}
\label{tab:comp_with_base}
\begin{tabularx}{\textwidth}{XXXXX}
\hline
English to & Best baseline & Best System on Curated Data & Difference & Accuracy with humans (\%) \\ \hline
Hindi & 51.63 & 53.84 & 2.21 & 85.7 \\
Odia & 41.57 & 46.64 & 5.07 & 92.7 \\
Nepali & 43.86 & 46.88 & 3.02 & 89.9 \\
Norwegian Nynorsk & 55.53 & 56.69 & 1.16 & 73 \\
German & 51.87 & 58.24 & 6.37 & 92.9 \\ \hline
\end{tabularx}
\end{table}
\paragraph{Table \ref{tab:comp_with_fp}: Comparison with Full Parallel Sentences System} This table provides additional evidence of the efficacy of our approach by contrasting the top curated systems with those trained on the complete parallel (FP) datasets. The column titled ``Data Size Ratio" stands out, highlighting the relatively modest amount of curated data needed to attain equal or superior results.
\begin{table}
\caption{Comparison between the system trained on Full Parallel Sentences and the Best System on Curated Data, reduction in dataset size of best curated data, along with the Accuracy with Human.}
\label{tab:comp_with_fp}
\begin{tabularx}{\textwidth}{XXXXXX}
\hline
English to & FP & Best System on Curated Data & Difference & Data Size Ratio & Accuracy with humans (\%) \\ \hline
Hindi & 53.13 & 53.84 & 0.71 & 0.44 & 64 \\
Odia & 44.95 & 46.64 & 1.69 & 0.14 & 80.7 \\
Nepali & 45.09 & 46.88 & 1.79 & 0.49 & 81.9 \\
Norwegian Nynorsk & 55.74 & 56.69 & 0.95 & 0.64 & 69 \\
German & 53.92 & 58.24 & 4.32 & 0.4 & 92.3 \\ \hline
\end{tabularx}
\end{table}
Both tables' results together demonstrate that employing a more intelligent data curation process achieves comparable or improved translation quality, even as it decreases the amount of data needed for training.


\section{Related Work}
This section provides a focused review of key research in data-centric approaches, but it's not an exhaustive list of all related work.
\subsection{Data acquisition}

The development of foundational corpora has been a primary focus in low-resource machine translation (MT). Government initiatives like the TDIL program and the Indian Language Corpora Initiative (ILCI) \cite{jha2012tdil} and large-scale data collection projects \cite{tiedemann-2012-parallel, kunchukuttan-etal-2018-iit, parida-etal-2020-odiencorp, siripragada-etal-2020-multilingual, haddow2020pmindia, nakazawa-etal-2021-overview, ramesh-etal-2022-samanantar, mujadia-sharma-2022-ltrc, gala2023indictrans2, pal-etal-2023-findings, AbadjiOrtizSuarezRomaryetal,Costa-jussà2024NLLB} have made significant strides in this area. 
While these efforts are crucial for providing the necessary data, their primary objective is to build massive corpora, often prioritizing quantity over the fine-grained linguistic quality of each sentence. LALITA is distinct from these approaches as it focuses on curating data based on its inherent linguistic value. The use of powerful tools like LaBSE \cite{feng-etal-2022-language,heffernan-etal-2022-bitext} provides a means for efficient data acquisition, but it is a collection method, not a curation method. It can gather vast amounts of data that still require further filtering to be of optimal use for model training.

Similarly, other foundational projects focus on creating resources for specific languages or linguistic phenomena, such as \citet{khenglawt-etal-2022-language}. The construction of a parallel corpus is a critical first step for low-resource languages, for example, the first large-scale Mizo-English corpus of over 529K sentence pairs was created using an integrated approach of web crawling and manual translation \cite{10.1145/3610404}. This immense undertaking highlights the challenge of building a foundational resource and ensures that if selected for dataset creation, a sentence can improve MT performance. Our work complements such projects by providing a method to maximize the value of this painstakingly collected data.

\subsection{Data Augmentation}

Data augmentation, particularly through back-translation \cite{sennrich2015improving}, is a well-established strategy to overcome data scarcity by generating synthetic parallel data. While a powerful technique, studies have shown that a basic, uncurated approach can be suboptimal \cite{edunov-etal-2018-understanding}. Work like \citet{soto-etal-2020-selecting} and \citet{10.5555/3495724.3496564} have explored more sophisticated methods to enhance model performance, such as combining data from different sources or generating diverse sentences. Another work, \citet{burchell-birch-and-kenneth-heafield-2022-exploring}, investigated the effect of different data generation methods on model performance, comparing techniques like nucleus sampling and a syntax-group fine-tuning approach. They found that nucleus sampling yields the better performance, but this we believe was because they were sampling from most frequent syntax-group. These methods represent significant progress by focusing on improving the quality and variety of synthetic data. However, our work offers a complementary perspective. Unlike these studies, which are centered on how to best \textit{generate} diverse back-translated data from scratch, our LALITA framework provides a systematic, linguistically-informed approach to \textit{select} the most valuable sentences from both pre-existing and synthetic corpora. This provides a method to further enhance existing augmentation pipelines.

\subsection{Data Filtering}

The sheer scale of modern corpora necessitates robust data filtering techniques to remove noise. Shared tasks like \citet{sloto-etal-2023-findings} and previous iterations in the \href{https://machinetranslate.org/wmt}{WMT} have given a platform to filter and curate parallel text. While some efforts rely on manual and heuristic-based filtering, such as in \citet{baturova-etal-2025-low}, more automated approaches have emerged. Work like \citet{aulamo-etal-2020-opusfilter} provides a wide range of basic filters based on heuristics and metadata. However, the effectiveness of such tools relies on a user's ability to configure a winning combination of filters, rather than providing a single, principled metric for assessing a sentence's intrinsic value. Our LALITA framework fills this gap by offering a scientifically grounded, multi-faceted score for linguistic complexity that moves beyond these ad-hoc combinations.

\subsection{Web Crawled Dataset Quality}

The problem of data quality in web-mined corpora is a well-documented issue. Work such as \citet{ranathunga-etal-2024-quality} and \citet{kreutzer-etal-2022-quality} have empirically demonstrated that the quality of these corpora can vary drastically, and that a random sample may not be sufficient for building an effective MT system. However, these works either rely on manual and labor-intensive audits or use metrics that lack the nuance to capture a sentence's inherent learning signal. LALITA's methodology is proactive and provides a deeper, more nuanced measure of a sentence's linguistic and lexical quality, enabling a more targeted curation process.
\subsection{Domain Adaptation}

A related area of research is domain adaptation, which focuses on selecting data relevant to a specific target domain. Work such as \citet{van-der-wees-etal-2017-dynamic,axelrod-etal-2015-data} demonstrate the value of data selection by using metrics like cross-entropy difference to identify relevant data. While highly effective for their specific use cases, these methods are inherently tied to a predefined target domain or seed corpus. This contrasts with the LALITA framework, which is a universal and unsupervised method. By deriving a metric of linguistic complexity from the training data itself, LALITA is broadly applicable to any translation task, including those with no clear target domain or in truly low-resource settings with no parallel development data available.

The LALITA framework introduces a unique and proactive data curation methodology that focuses exclusively on the linguistic and lexical attributes of the source sentences. This source-centric approach is a critical differentiator, making LALITA particularly applicable in scenarios where target-side quality assessment is difficult or not available. By providing a principled method for curating linguistically complex data, LALITA offers a direct solution to the data quality challenges identified by \citet{khayrallah-koehn-2018-impact}. Our work ensures that NMT models are trained on high-quality data that provide a rich learning signal, thereby improving robustness and performance in a way that simply adding more, uncurated data cannot achieve. Unlike reactive methods that merely clean data or expand its volume, LALITA proactively curates data based on its inherent linguistic value. This approach to data management could lead to new research directions, shifting the focus from the quantity of data to its quality and characteristics.

\section{Limitations}
Despite the significant advancements offered by the LALITA framework, certain limitations exist in the current work, which also delineate clear avenues for future research.

\subsection{Computational Resource Constraints}

The exploratory nature of this research, particularly the exhaustive search for optimal configurations across various dataset sizes and cluster distributions, proved to be computationally intensive and resource-consuming. While the results are compelling, the extensive computational requirements pose a practical barrier to fully exploring the optimal parameter space for LALITA across every possible language pair and resource scenario. This highlights a scalability challenge for comprehensive exploration, suggesting that finding the absolute best configuration might be computationally prohibitive without further optimisation of the search process itself.

\subsection{Dependency on Source Language Resources}

A fundamental requirement for the application of LALITA is the availability of an adequate and accurate parser for the source language. This dependency presents a challenge for truly low-resource language pairs where both languages lack such robust linguistic tools. In scenarios where foundational NLP tools for the source language are scarce, the direct execution of this work becomes difficult. This points to a ``cold start problem" for deep linguistic analysis, as LALITA relies on features extracted by parsers that themselves require significant linguistic resources. To address this, future work may need to explore pivoting approaches or cross-lingual transfer methods that can circumvent the need for direct, high-quality source language parsing in extremely low-resource settings.

\subsection{Absence of Explicit Discourse Information}

The current iteration of the LALITA framework primarily operates at the sentence level, meaning it does not incorporate explicit discourse information. While sentence-level complexity is crucial, higher-level linguistic phenomena such as cohesion, coherence, and anaphora also significantly contribute to the overall complexity and translation difficulty of a text. This limitation suggests that LALITA currently captures only a part of the full spectrum of linguistic intricacy. This opens a clear avenue for future research to extend LALITA's capabilities to higher levels of linguistic analysis, potentially yielding even more nuanced and effective curation for document-level machine translation.

\subsection{Exclusion of Sentence Embeddings as Features}
The current work does not incorporate sentence embeddings as additional features in the LALITA score. This decision is intentional, as a core aspect of the framework is its emphasis on explicit linguistic and lexical properties, which are interpretable and less resource-intensive to extract. That said, we plan to include sentence embeddings in future work, as their integration could create a more comprehensive sentence descriptor. Combining structural and semantic information in this way would further refine data curation strategies and improve model performance.

\section{Conclusion and Future Work}

\subsection{Conclusion: The Value of Complexity-Driven Data Curation}

This research demonstrates a transformative approach to Machine Translation data curation. It unequivocally shows that training MT systems predominantly on structurally complex sentences leads to superior performance. The LALITA score, derived from a linguistically and lexically informed analysis, provides a robust mechanism for identifying and prioritizing these high-value sentences. This strategic data selection significantly reduces the required dataset size while maintaining or even improving translation quality, thereby addressing the high costs associated with data creation and model training. 

Furthermore, the utility of LALITA as a viable strategy for data augmentation has been established. By intelligently augmenting data within specific complexity clusters, the framework ensures diverse and impactful representation, moving beyond mere volume expansion. While the best-performing systems consistently included a high proportion of longer and more complex sentences, the analysis also underscored that relying solely on single metrics like sentence length is insufficient; a holistic consideration of all integrated linguistic and lexical features, as captured by the LALITA score, is crucial. In contexts where the cost of creating and training MT data is substantial, LALITA offers an efficient solution by identifying the most valuable subset of data, thereby translating into tangible economic benefits for developers and a reduced environmental footprint for large-scale AI.

\subsection{Future Work: Expanding LALITA's Horizons}

Building upon the insights gained and addressing the identified limitations, several promising directions for future research are envisioned:
\begin{itemize}
    
\item \textbf{Enhanced Computational Efficiency}: Developing more computationally efficient methods for searching optimal LALITA configurations is a priority. This could involve exploring advanced optimisation algorithms or meta-learning approaches to accelerate the discovery of ideal data curation strategies. 
\item \textbf{Using both Source and Target linguistic and lexical features}: In current study we have done only source side analysis with a practical view that one of the languages might not have enough resources to extract linguistic features. Having said that, it is an interesting thread of work, which can give us more insights in how we can curate dataset for MT. 
\item \textbf{Data Augmentation}: Combine LALITA's complexity-based curation with semantic uncertainty-based sampling \cite{jia-etal-2025-semantic} for Data augmentation. This would create a truly multi-faceted data selection pipeline that considers both source-side linguistic complexity and cross-lingual semantic uncertainty.

\item \textbf{Heuristics for Parallel Text Creation}: Investigating how the LALITA score can be used to develop heuristics that \textbf{actually} guide the creation of new parallel text, rather than just the selection from existing corpora. This could involve directing human translators or automated generation processes to produce sentences with desired complexity profiles, thereby optimising the return on investment for new data acquisition, especially for low-resource languages where human translation is expensive. 

\item \textbf{Curriculum Learning}: Exploring the application of LALITA scores within a curriculum learning paradigm \cite{mohiuddin-etal-2022-data}. This would involve systematically training MT models by progressively introducing sentences of increasing complexity (as measured by the LALITA score), rather than one feature. This approach could potentially lead to more stable and efficient learning, allowing models to build foundational understanding on simpler structures before tackling more intricate linguistic phenomena.

\item \textbf{Multilingual Applications and Beyond English Source Languages}: Expanding the LALITA framework to directly analyze and score complexity in source languages other than English. This would involve developing language-specific feature extraction and PCA models for a wider range of languages, enabling LALITA's benefits to be fully realized in truly multilingual contexts \cite{johnson-etal-2017-googles,aharoni-etal-2019-massively,de-gibert-etal-2023-four,roy-etal-2024-enhancing} and for language pairs where English is not the source. This is particularly relevant for addressing low-resource language challenges where robust English-centric tools might not be sufficient. 

\item \textbf{Incorporation of Discourse Information}: Extending the LALITA framework to incorporate explicit discourse information is a vital next step. This would allow the framework to capture higher-level linguistic complexities beyond the sentence level, potentially leading to further improvements in document-level MT performance.  

\item \textbf{Ablation of Sentence Embeddings for LALITA Score and Data Curation}: 

A comprehensive data curation pipeline can be created by combining advanced embedding models like LASER3 \cite{heffernan-etal-2022-bitext} and LaBSE \cite{feng-etal-2022-language} with the LALITA score. While embedding models are effective at mining and filtering for semantic similarity, the LALITA score quantifies a sentence's structural and syntactic complexity, enabling a more nuanced selection process. This combined approach allows for the creation of a smaller, high-quality dataset rich in complex linguistic constructions, leading to comparable or superior machine translation performance with significantly less data. Furthermore, incorporating sentence embeddings as additional features in the LALITA framework could create a more comprehensive sentence descriptor, capturing both structural and semantic information to further refine data curation strategies and improve model performance.


\item \textbf{Optimising LLM Pretraining with the LALITA Score}: Building on this research, a promising future direction is to integrate the LALITA framework with the second-stage training methodology for Large Language Models (LLMs) \cite{qorib-etal-2025-just}. Given that LLM pretraining is resource-intensive and ``quality trumps over quantity" in data selection, using the LALITA score to proactively curate a high-quality, complex dataset could reduce the vast data volumes and computational costs currently required. By exposing LLMs to a richer variety of structurally complex linguistic constructions, the LALITA score could enhance their ability to learn nuanced grammar and semantics, improving generalization and preventing homogenization to common patterns. Furthermore, it offers a quantifiable metric for strategic data curation and augmentation, allowing for the strategic inclusion of complex synthetic data to address specific linguistic deficiencies in the training corpus. 
\item \textbf{Further Study of LALITA Score and Sentence Complexity}: While this work demonstrates the utility of the LALITA score in data curation for Machine Translation, a significant future thread involves a deeper, more theoretical study of the score itself. This includes investigating its intrinsic properties and its precise mathematical and linguistic bindings with various facets of sentence complexity. For instance, how does sentence structure vary when the LALITA score changes by approximately 0.1? Such research could lead to a more profound understanding of linguistic complexity metrics and their potential applications beyond data curation.
\end{itemize}

This roadmap for universal applicability demonstrates a commitment to ongoing research and a vision for LALITA to become a widely adopted and foundational tool in the Machine Translation community.

\bmhead{Acknowledgements}
We would like to thank Pruthwik Mishra, as a conversation with him sparked the initial idea for this direction. We are also grateful to Aparajitha Allamraju, Prashant Kodali, Sneha Nanavati, Nikhilesh Bhatnagar, Vandan Mujadia, Nirmal Surange, Mitesh Khapra, Anoop Khunchukuttan, Sudip Kumar Naskar, Radhika Mamidi, Sandipan Dandapat,  and Dipti Misra Sharma for their valuable input and suggestions throughout this work.




\section*{Declarations}
\begin{itemize}
\item Funding: No funding was received for conducting this study.
\item Conflict of interest/Competing interests: The authors declare that they have no conflict of interest
\item Ethics approval and consent to participate: Not applicable
\item Consent for publication: Yes
\item Data availability: Data already publicly available. 
\item Materials availability: Not applicable
\item Code availability: The code uses libraries publicly available, and steps are reproducible. 
\item Author contribution: Yadav and Shrivastava conceptualized the framework, with Yadav building the framework and model development. Both authors contributed to writing and reviewing the manuscript.
\end{itemize}

\noindent





\newpage
\begin{appendices}
\section{Features with non-zero Absolute Magnitude of coefficient of 1st principal component.}
\label{sec:all_coeff_pca1}

\begin{table}[h!]
\caption{Features with the Non-zero Absolute Magnitude of the coefficient of the 1st principal component.}
\label{tab:all_coeff_pca1}
\begin{tabular}{cc|cc|cc}
\hline
Feature & \begin{tabular}[c]{@{}c@{}}Absolute \\ Magnitude \\ of coefficient\end{tabular} & Feature & \begin{tabular}[c]{@{}c@{}}Absolute \\ Magnitude \\ of coefficient\end{tabular} & Feature & \begin{tabular}[c]{@{}c@{}}Absolute \\ Magnitude \\ of coefficient\end{tabular} \\ \hline
sentenceLength & 0.244 & VerbForm\_Ger & 0.100 & Case\_Acc & 0.036 \\
NoUMF & 0.219 & PRON & 0.100 & obl:npmod & 0.036 \\
VERB & 0.190 & VerbForm\_Inf & 0.099 & compound:prt & 0.036 \\
ADP & 0.185 & PROPN & 0.098 & obl:tmod & 0.034 \\
case & 0.185 & Definite\_Ind & 0.096 & Degree\_Cmp & 0.032 \\
NOUN & 0.185 & PART & 0.094 & PronType\_Int & 0.030 \\
Number\_Sing & 0.184 & aux & 0.093 & NumType\_Ord & 0.029 \\
det & 0.181 & acl & 0.087 & SYM & 0.027 \\
DET & 0.180 & acl:relcl & 0.085 & Person\_1 & 0.027 \\
PronType\_Art & 0.177 & PronType\_Prs & 0.085 & expl & 0.025 \\
VerbForm\_Fin & 0.167 & PronType\_Rel & 0.083 & Degree\_Sup & 0.024 \\
Definite\_Def & 0.158 & ADV & 0.081 & parataxis & 0.023 \\
Mood\_Ind & 0.149 & advmod & 0.081 & X & 0.019 \\
obl & 0.145 & nmod:poss & 0.076 & det:predet & 0.019 \\
Tense\_Past & 0.143 & ORG & 0.075 & Gender\_Fem & 0.018 \\
nmod & 0.140 & aux:pass & 0.074 & csubj & 0.017 \\
amod & 0.138 & Voice\_Pass & 0.074 & cc:preconj & 0.017 \\
ADJ & 0.138 & Tense\_Pres & 0.074 & Person\_2 & 0.014 \\
mark & 0.137 & xcomp & 0.068 & nmod:npmod & 0.013 \\
Degree\_Pos & 0.135 & nsubj:pass & 0.068 & iobj & 0.011 \\
cc & 0.132 & Poss\_Yes & 0.067 & Reflex\_Yes & 0.010 \\
CCONJ & 0.132 & MISC & 0.059 & NumType\_Mult & 0.006 \\
punct & 0.130 & Case\_Nom & 0.057 & Typo\_Yes & 0.006 \\
PUNCT & 0.129 & Gender\_Neut & 0.054 & discourse & 0.006 \\
nsubj & 0.126 & NUM & 0.053 & INTJ & 0.005 \\
Number\_Plur & 0.126 & NumType\_Card & 0.053 & vocative & 0.004 \\
obj & 0.123 & appos & 0.052 & list & 0.004 \\
AUX & 0.120 & nummod & 0.052 & Mood\_Imp & 0.003 \\
SCONJ & 0.118 & LOC & 0.049 & Abbr\_Yes & 0.003 \\
VerbForm\_Part & 0.116 & PER & 0.048 & goeswith & 0.003 \\
advcl & 0.115 & Gender\_Masc & 0.048 & nmod:tmod & 0.003 \\
compound & 0.110 & flat & 0.047 & Foreign\_Yes & 0.002 \\
conj & 0.109 & fixed & 0.044 & orphan & 0.002 \\
Person\_3 & 0.103 & cop & 0.043 & csubj:pass & 0.001 \\
ccomp & 0.101 & PronType\_Dem & 0.037 & reparandum & 0.001 \\ \hline
\end{tabular}%
\end{table}
\newpage
\section{PCA1 vs for features}
\label{app:pca1Feature}
\begin{figure}[h!]
    \centering
  \includegraphics[width=1\linewidth]{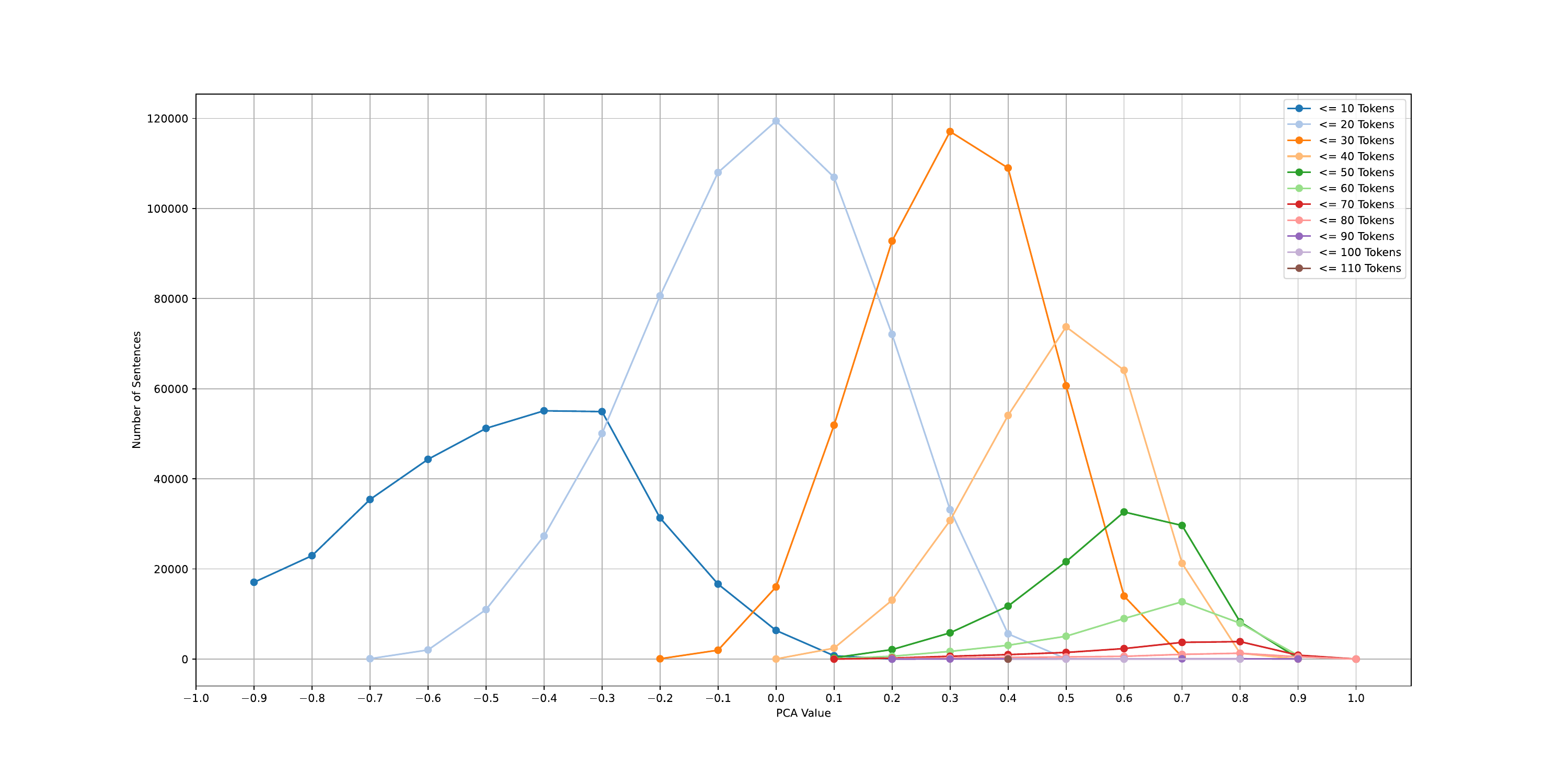}
  
\caption{Distribution of PCA1 Value v/s Number of sentence for Sentence Length.}
\label{fig:pca1tok}
\end{figure}

\begin{figure}[h!]
    \centering
  \includegraphics[width=1\linewidth]{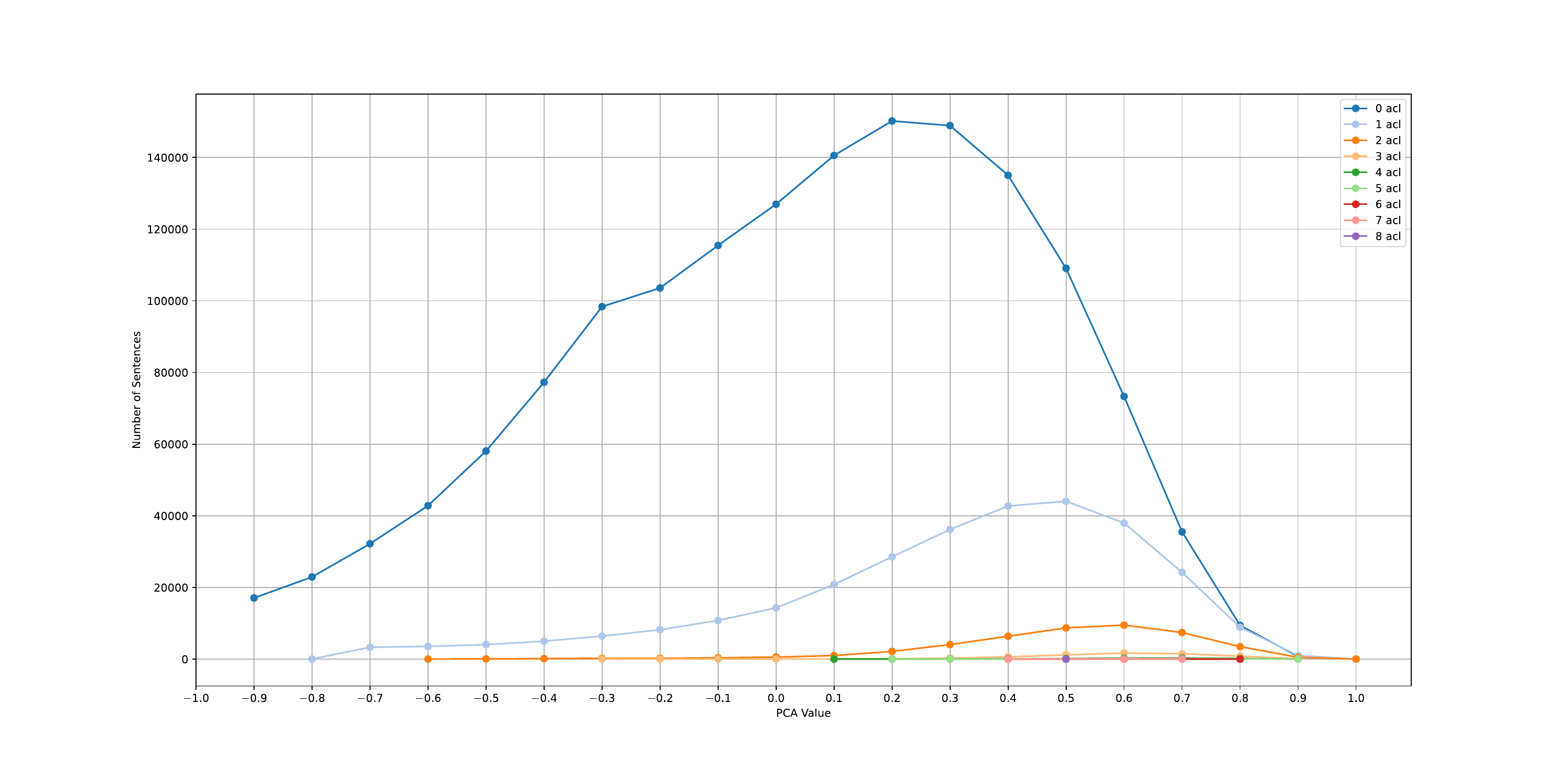}
  
\caption{Distribution of PCA1 Value v/s Number of sentence for Clausal Modifier (acl).}
\label{fig:pca1acl}
\end{figure}

\begin{figure}[h!]
    \centering
  \includegraphics[width=1\linewidth]{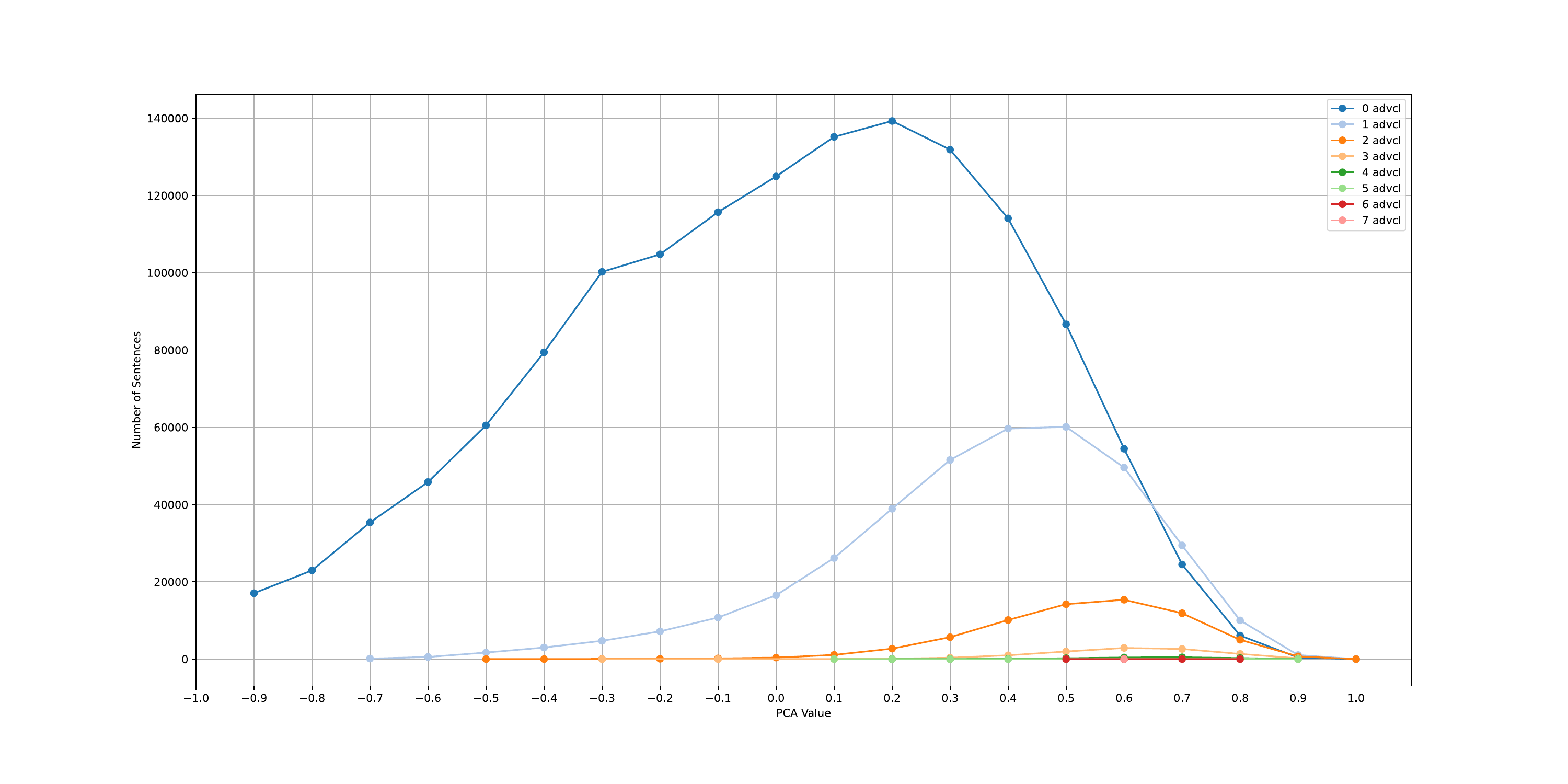}
  
\caption{Distribution of PCA1 Value v/s Number of sentence for adverbial Modifier (advcl).}
\label{fig:pca1advcl}
\end{figure}

\begin{figure}[h!]
    \centering
  \includegraphics[width=1\linewidth]{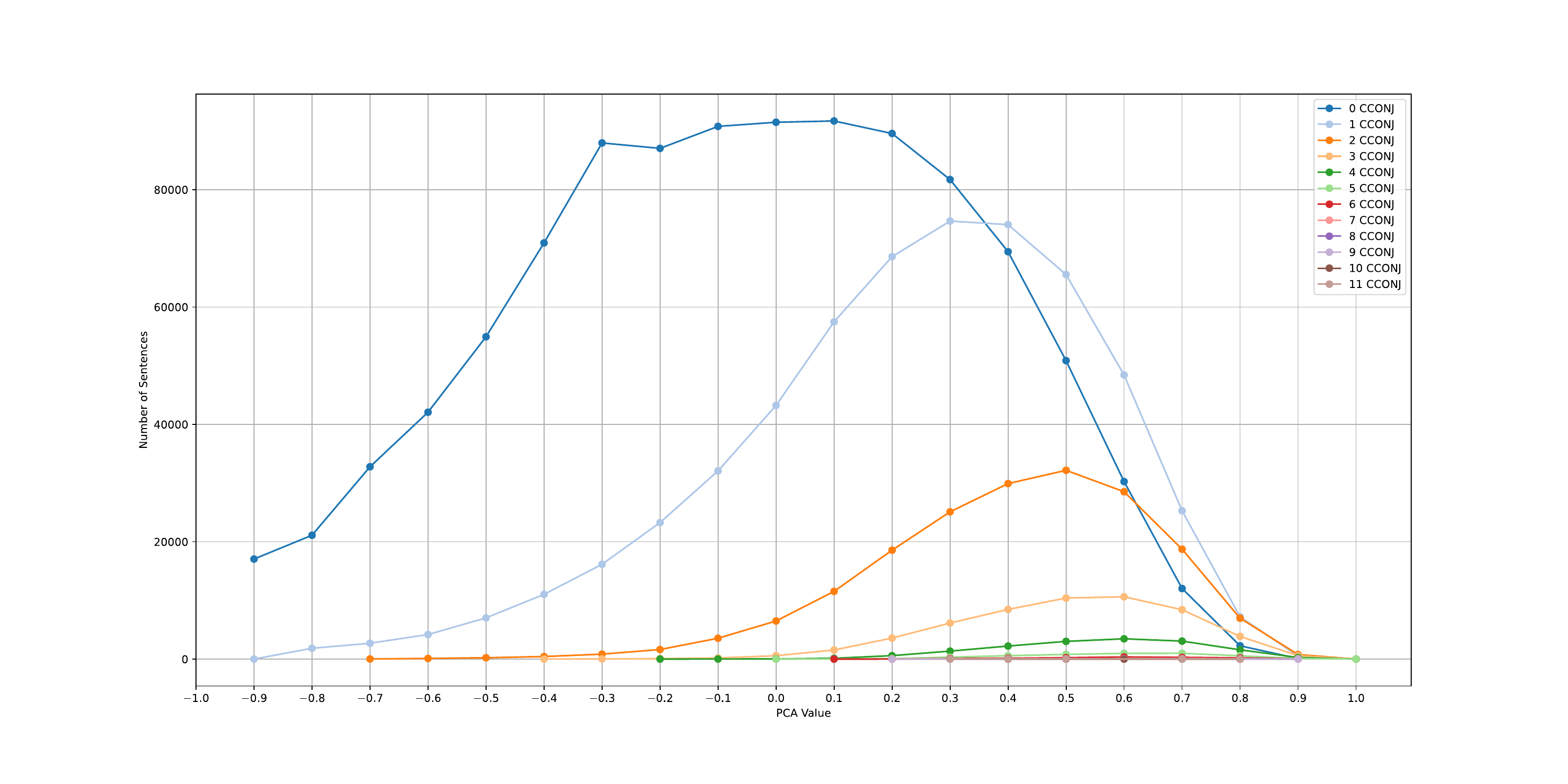}
  
\caption{Distribution of PCA1 Value v/s Number of sentence for Conjunctions.}
\label{fig:pca1conj}
\end{figure}
.



\section{Hyperparameters for Training Transformer Model}
\label{subsec:hyperparameters}
We followed the official Fairseq tutorial instructions for preprocessing, training, and translation\footnote{\url{https://fairseq.readthedocs.io/en/latest/getting_started.html}}, and customised the parameters given in Table \ref{tab:hyperparams} with respective values for all experiments.

\begin{table}[h!]
\centering
\resizebox{\columnwidth}{!}{%
\begin{tabular}{|c|c|}
\hline
\textbf{Parameter} & \textbf{Value} \\ \hline
\texttt{arch} & transformer \\ \hline
\texttt{optimizer} & adam \\ \hline
\texttt{adam-betas} & (0.9, 0.98) \\ \hline
\texttt{clip-norm} & 0.0 \\ \hline
\texttt{lr} & 5e-4 \\ \hline
\texttt{lr-scheduler} & inverse\_sqrt \\ \hline
\texttt{warmup-updates} & 4000 \\ \hline
\texttt{warmup-init-lr} & 1e-07 \\ \hline
\texttt{dropout} & 0.3 \\ \hline
\texttt{attention-dropout} & 0.1 \\ \hline
\texttt{activation-dropout} & 0.1 \\ \hline
\texttt{weight-decay} & 0.0001 \\ \hline
\texttt{criterion} & label\_smoothed\_cross\_entropy \\ \hline
\texttt{label-smoothing} & 0.1 \\ \hline
\texttt{max-tokens} & 6000 (GTX1080Ti), 1100 (RTX2080Ti), 25000 (RTX4090) \\ \hline
\texttt{max-update} & 300000 \\ \hline
\texttt{patience} & 20 \\ \hline
\texttt{update-freq} & 10 \\ \hline
\end{tabular}}
\caption{Training hyperparameters used across all experiments, with \texttt{max-tokens} for respective GPUs.}
\label{tab:hyperparams}
\end{table}
\subsection{GPU usage}
Table \ref{tab:gpu} shows the breakdown of the GPU workload for each language experiment.
\begin{table}[]
\centering
\caption{GPUs used and Total (approx) GPU hours used for each language.}
\label{tab:gpu}
\begin{tabular}{l|r|r|r}
\hline
GPU & \multicolumn{1}{l|}{GTX1080Ti} & \multicolumn{1}{l|}{RTX2080Ti} & \multicolumn{1}{l}{RTX 4090} \\ \hline
Hindi     & 18624 & 384 & 612  \\
Nepali    & 3492  & 90  & 102  \\
Odia     & 3492  & 90  & 102  \\
Norwegian & 3492  & 90  & 102  \\
German    & 0     & 0   & 332  \\ \hline
Total     & 29100 & 654 & 1250 \\ \hline
\end{tabular}%
\end{table}
\section{Performance on cluster configurations}
\label{sec:configAll}
\subsection{Four clusters}
\begin{figure}[t!]
  \centering
  \includegraphics[width=1\columnwidth]{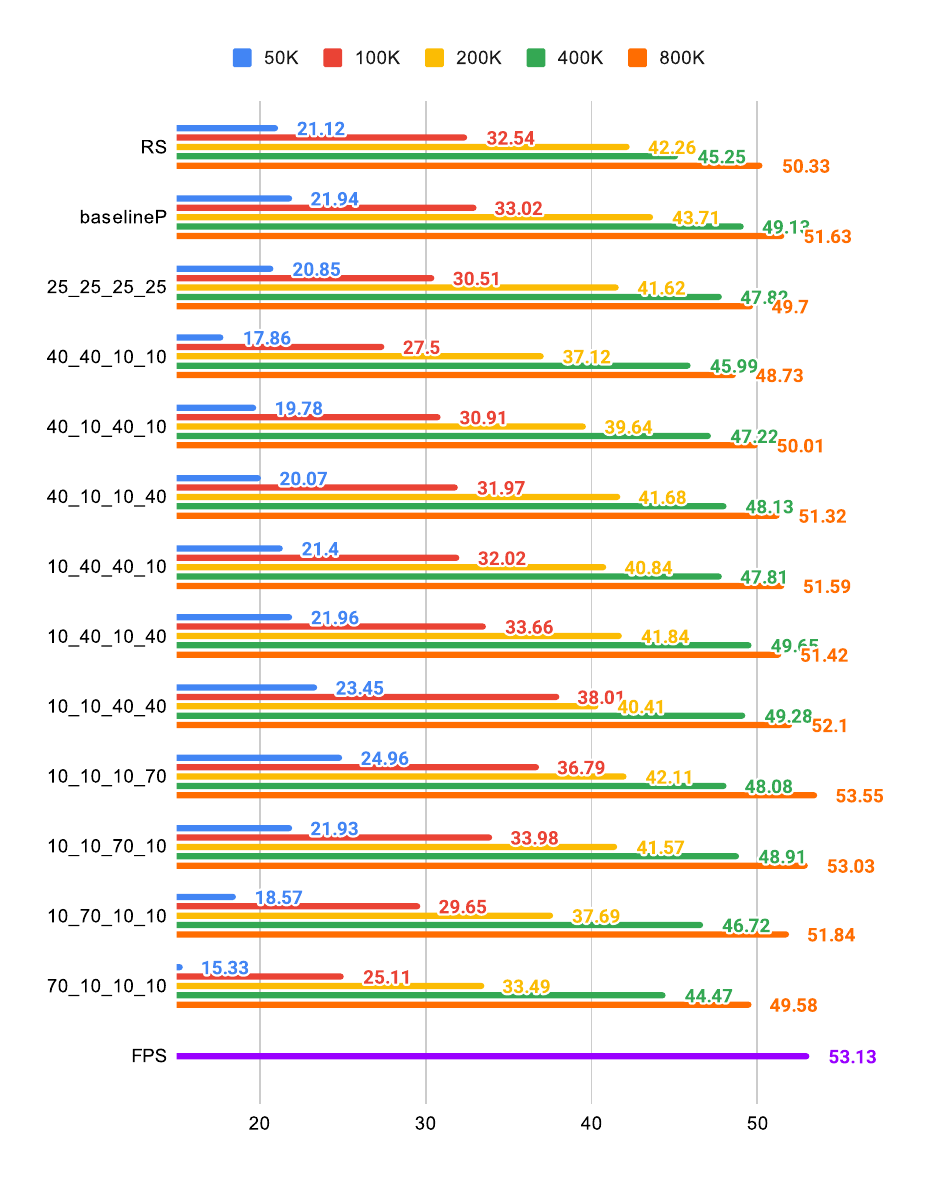}
  \caption{Performance on Systems trained on data sampled from All Clusters}
  \label{fig:allClusters}
\end{figure}

We sampled from all clusters for all dataset sizes using 
\begin{itemize}
    \item Uniform Sampling: We sampled equally from each Cluster for each dataset size.
    \item One Major $<$70,10,10,10$>$: Sampling Majorly from only one cluster. 
    \item Two Clusters High and Two Clusters Low $<$40,40,10,10$>$: Sampling the Majority of sentences from two clusters and a minor amount of text from the other two clusters.
\end{itemize}
We found that the system trained majorly from Cluster 3 (\textit{10\_10\_10\_70}) would outperform the baseline systems (Figure \ref{fig:allClusters}) followed by system trained having data mostly from Cluster 2. Whereas systems performing worse would have very less number of sentences sampled from Cluster 3.

\subsection{Three Clusters}
\begin{figure}[t!]
  \centering
  \includegraphics[width=1\columnwidth]{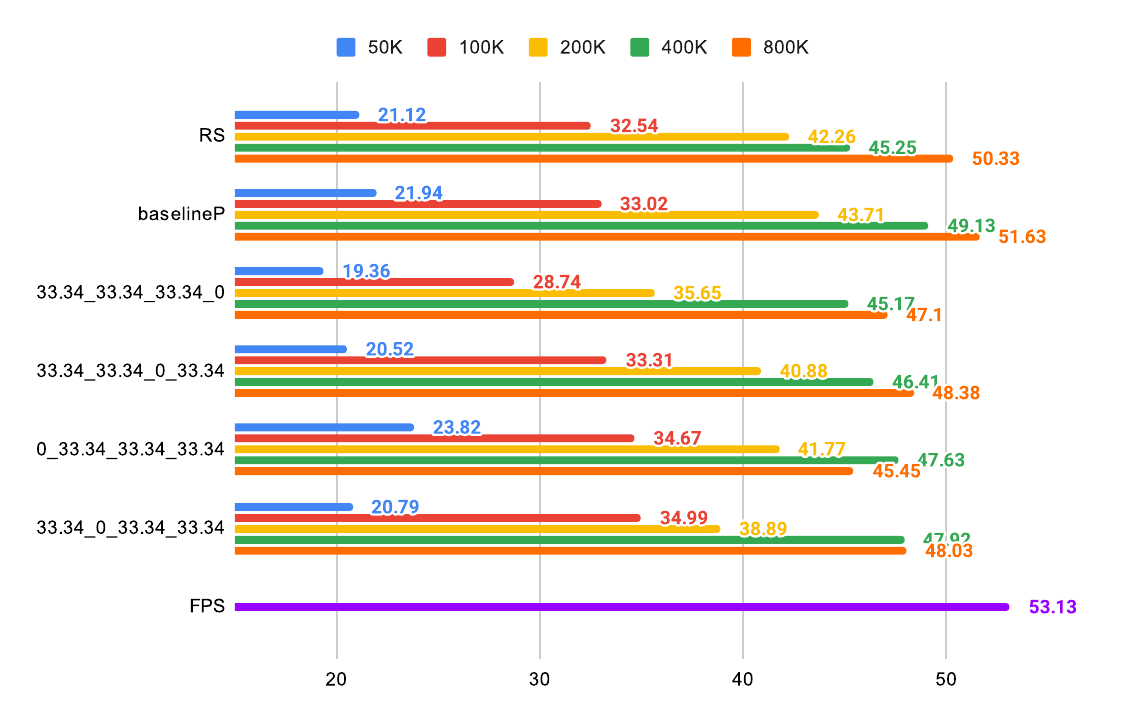}
  \caption{Performance on Systems trained on data sampled Uniformly from Three Clusters}
  \label{fig:3ClusterUniform}
\end{figure}

\begin{figure}[t!]
  \centering
  \includegraphics[width=1\columnwidth]{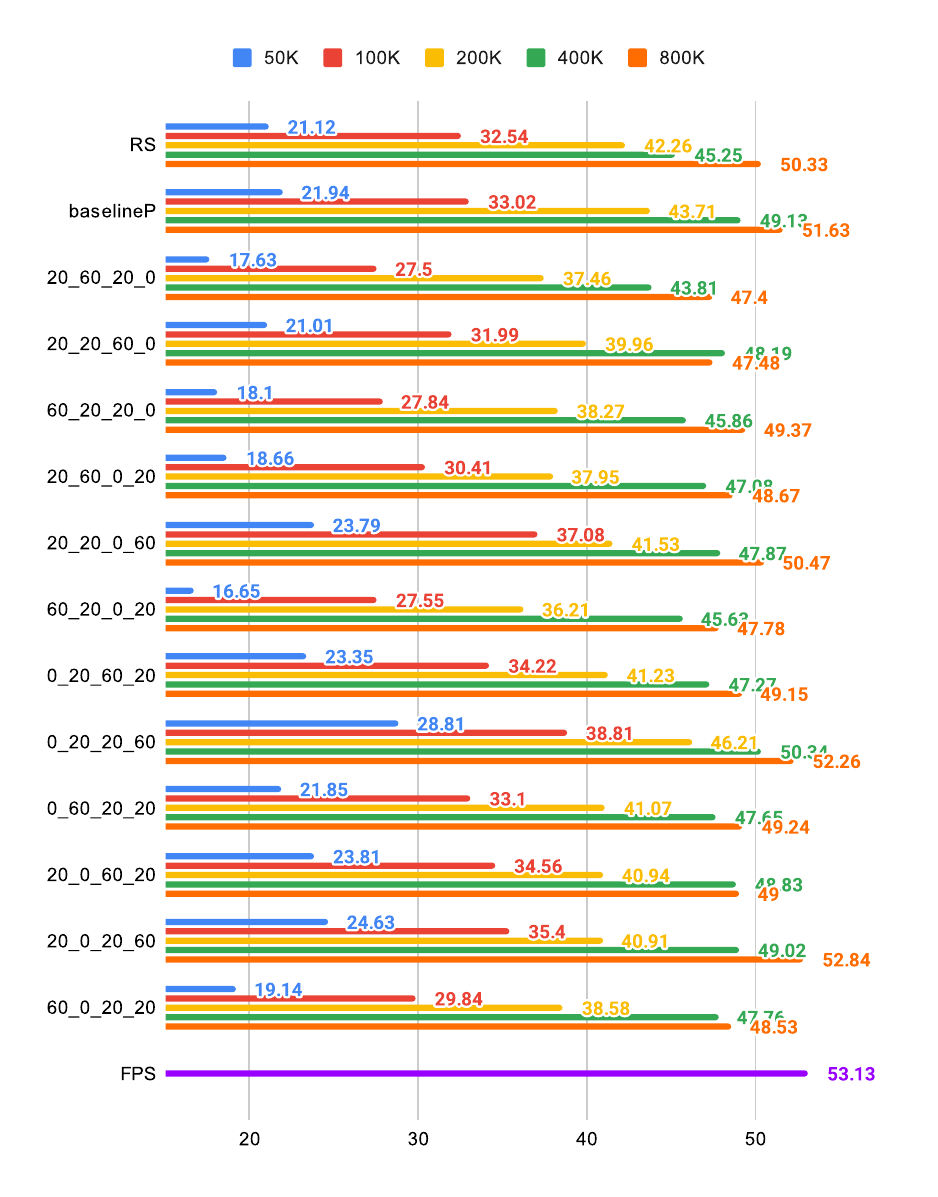}
  \caption{Performance on Systems trained on $<$60,20,20,0$>$}
  \label{fig:3Cluster1}
\end{figure}

\begin{figure}
  \centering
  \includegraphics[width=1\columnwidth]{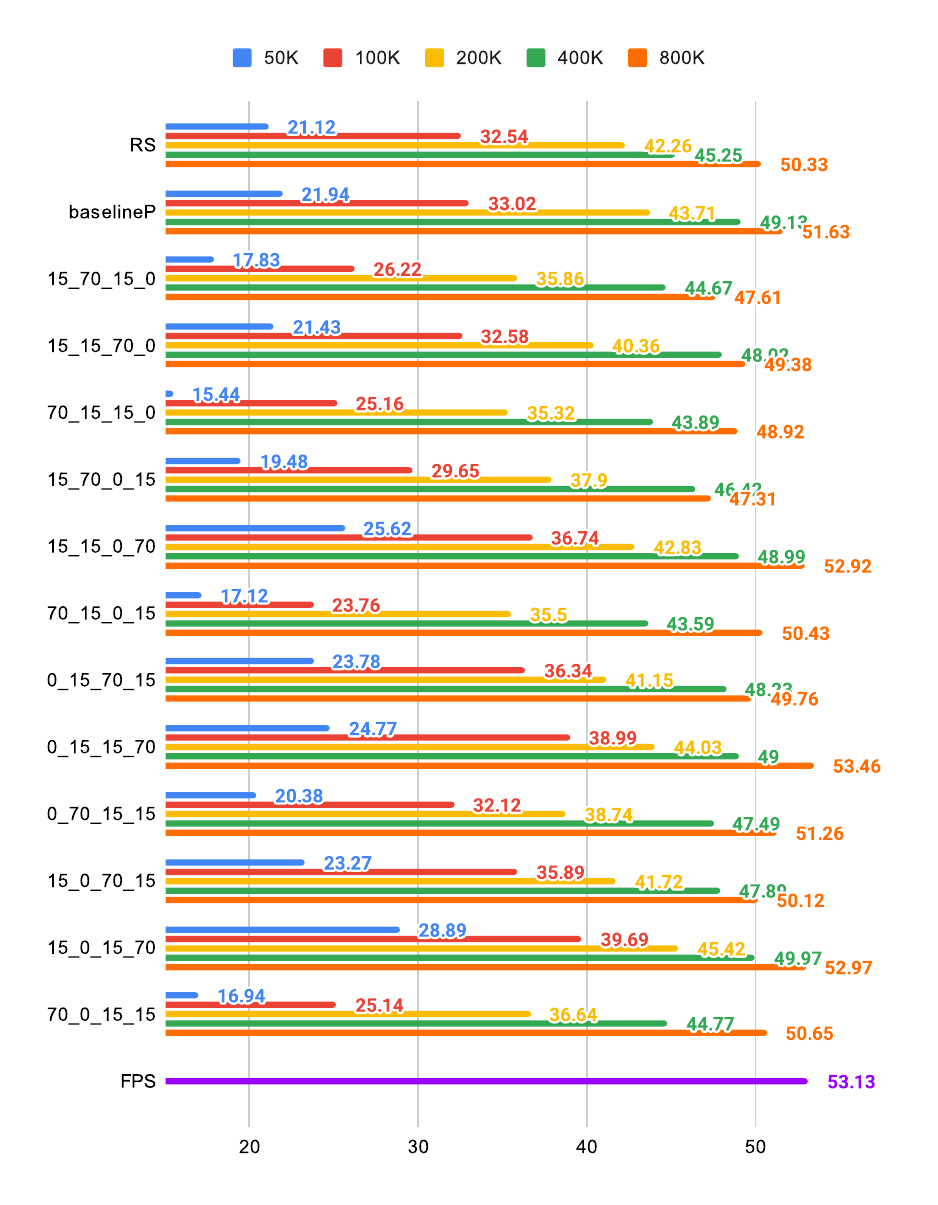}
  \caption{Performance on Systems trained on $<$70,15,15,0$>$}
  \label{fig:3Cluster2}
\end{figure}

We trained systems sampled from three clusters using every 3 cluster combination with the following sampling,
\begin{itemize}
    \item Uniform Sampling (Figure \ref{fig:3ClusterUniform}): Sampled equally from each of the three clusters.
    \item One Major $<$60,20,20,0$>$ (Figure \ref{fig:3Cluster1}): We sampled majorly (60\%) from one cluster while sampling an equal amount from the other two (20\% each). 
    \item One predominant cluster $<$70,15,15,0$>$ (Figure \ref{fig:3Cluster2}: We primarily sampled 70\% from a single cluster, with the remaining two clusters providing 15\% each.
\end{itemize}
With Uniform Sampling, we found that systems underperformed compared to baseline systems. Even the systems using majorly from Cluster 2 and 3 (33.34\% from each). This indicates that if Clusters with simple sentences (Cluster 0 or 1 ) are one-third of the data, then system would suffer. 
With $<$60,20,20,0$>$ configuration set (Figure \ref{fig:3Cluster1}), we found that system trained on data sampled majorly from Cluster 3 performed comparable to FPS, where as we saw clear decline in performance as we sampled more from Cluster 0 and 1. 
With skewed Configuration having 70\% data drawn from one cluster for training data, we found that systems with cluster 3 having higher sentence percentage (\textit{15\_15\_0\_70}, \textit{0\_15\_15\_70}, \textit{15\_0\_15\_70}) performed comparable to FPS, while outperforming other systems. On other side we saw that absence of sentences from Cluster 3 for systems underperformed consistently.
\subsection{Two Clusters}
\begin{figure}[t!]
  \centering
  \includegraphics[width=1\columnwidth]{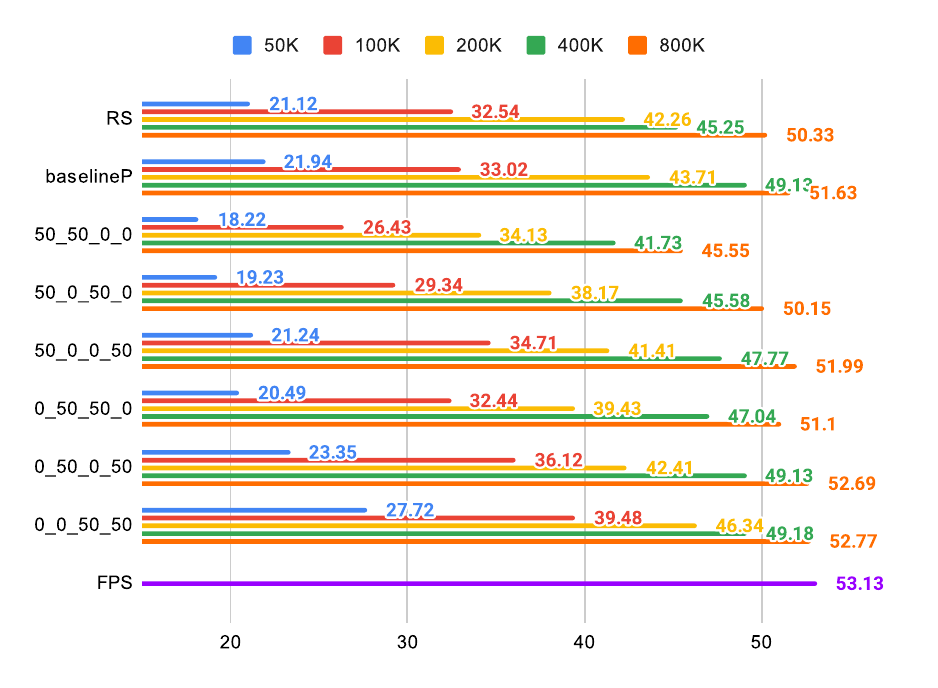}
  \caption{Performance on Systems trained on uniformly data sampled from Two Clusters}
  \label{fig:unfirom2Cluster}
\end{figure}

\begin{figure}[t!]
  \centering
  \includegraphics[width=1\columnwidth]{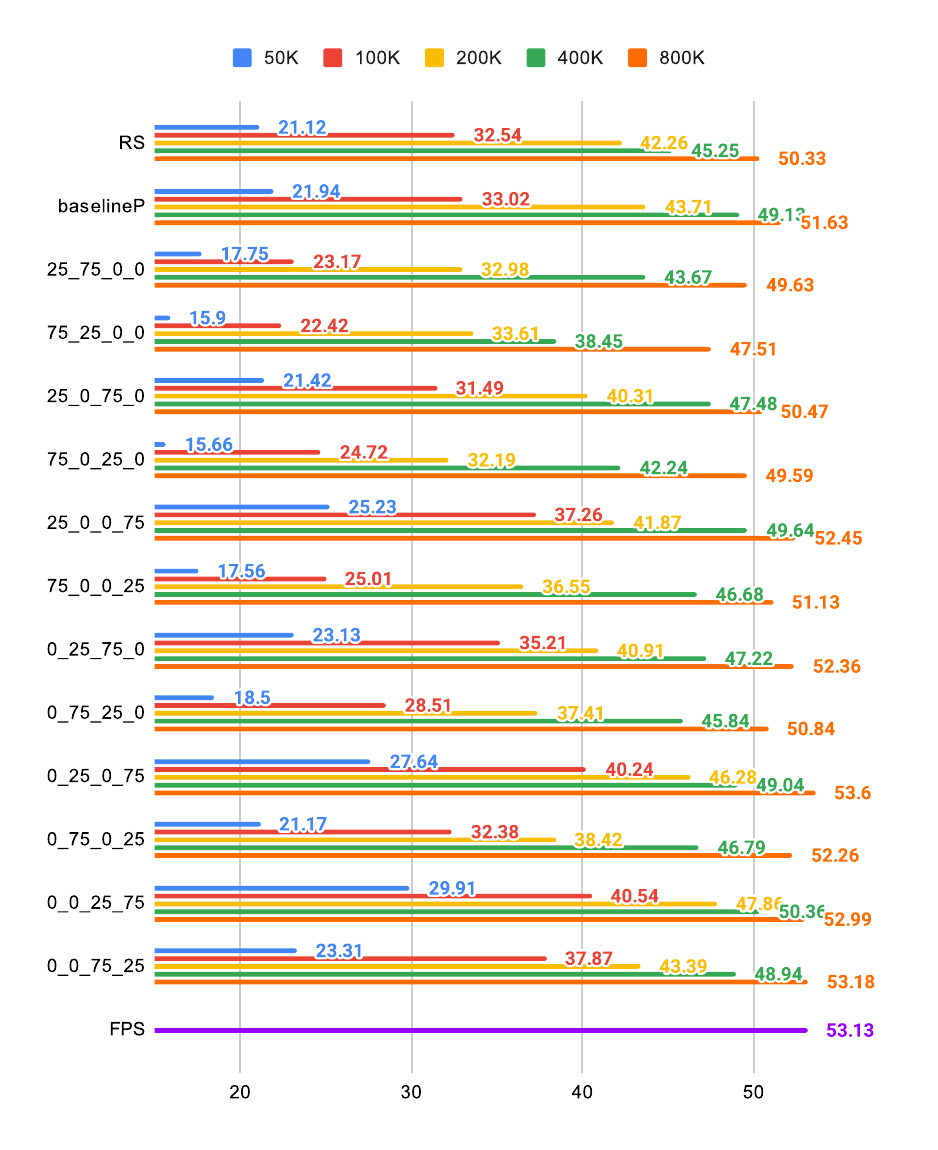}
  \caption{Performance on $<$75,25,0,0$>$ configuration set}
  \label{fig:skewed2Cluster}
\end{figure}

We sampled using two clusters with two configurations, 
\begin{itemize}
    \item Uniform sampling (Figure \ref{fig:unfirom2Cluster}): Sampling 50\% from each cluster.
    \item skewed sampling (Figure \ref{fig:skewed2Cluster}): Sampling Majorly from one cluster (75\%) while rest (25\%) from other cluster.
\end{itemize}
We again found that systems sampled with Cluster 3 ($<$*,*,*,50$>$) outperformed other configurations but underperformed compared to FPS.
We found that in skewed distribution systems using Cluster 2 or 3 performed better than other configurations, with systems having no sentences from Cluster 3 would underperform. 
\section{Total Tokens for all Experiments}
\label{appsec:totalTokens}
Table \ref{tab:totalTokens_enhi_full} lists the total tokens for each language pair for each configuration and dataset size in the English $\rightarrow$ Hindi experiments. The top-performing systems have around 30 million tokens on both the source and target sides, with 800K sentence pairs. Even in smaller dataset sizes, systems predominantly from cluster 3, such as \textit{0\_0\_0\_100}, consistently exhibit higher token counts compared to other configurations of the same dataset size. Similar pattern is observed in Tables \ref{tab:totalTokens_enord_short}, \ref{tab:totalTokens_ennpi_short}, \ref{tab:totalTokens_ennno_short} and  \ref{tab:totalTokens_ende_short} for English-Odia, Nepali, Norwegian Nynorsk and German respectively .

\begin{landscape}
\begin{longtable}{|c|cc|cc|cc|cc|cc|}
\caption{Total Tokens (in Millions) for all training data set size and configuration. The last row contains English and Hindi tokens for Full Parallel Sentences}
\label{tab:totalTokens_enhi_full}\\
\hline
\textbf{Dataset Size} &
  \multicolumn{2}{c|}{\textbf{50K}} &
  \multicolumn{2}{c|}{\textbf{100K}} &
  \multicolumn{2}{c|}{\textbf{200K}} &
  \multicolumn{2}{c|}{\textbf{400K}} &
  \multicolumn{2}{c|}{\textbf{800K}} \\ \hline
\endhead
\textbf{Language Pairs} &
  \multicolumn{1}{c|}{\textbf{English}} &
  \textbf{Hindi} &
  \multicolumn{1}{c|}{\textbf{English}} &
  \textbf{Hindi} &
  \multicolumn{1}{c|}{\textbf{English}} &
  \textbf{Hindi} &
  \multicolumn{1}{c|}{\textbf{English}} &
  \textbf{Hindi} &
  \multicolumn{1}{c|}{\textbf{English}} &
  \textbf{Hindi} \\ \hline
\textbf{RS} &
  \multicolumn{1}{c|}{1.36} &
  1.46 &
  \multicolumn{1}{c|}{2.73} &
  2.94 &
  \multicolumn{1}{c|}{5.46} &
  5.88 &
  \multicolumn{1}{c|}{10.91} &
  11.74 &
  \multicolumn{1}{c|}{21.84} &
  23.5 \\ \hline
\textbf{baselineP} &
  \multicolumn{1}{c|}{1.37} &
  1.47 &
  \multicolumn{1}{c|}{2.73} &
  2.94 &
  \multicolumn{1}{c|}{5.46} &
  5.87 &
  \multicolumn{1}{c|}{10.93} &
  11.74 &
  \multicolumn{1}{c|}{21.83} &
  23.5 \\ \hline
\textbf{100\_0\_0\_0} &
  \multicolumn{1}{c|}{0.74} &
  0.87 &
  \multicolumn{1}{c|}{1.43} &
  1.68 &
  \multicolumn{1}{c|}{2.74} &
  3.25 &
  \multicolumn{1}{c|}{5.54} &
  6.42 &
  \multicolumn{1}{c|}{10} &
  11.76 \\ \hline
\textbf{0\_100\_0\_0} &
  \multicolumn{1}{c|}{1.19} &
  1.32 &
  \multicolumn{1}{c|}{2.32} &
  2.57 &
  \multicolumn{1}{c|}{4.42} &
  4.93 &
  \multicolumn{1}{c|}{8.02} &
  9.05 &
  \multicolumn{1}{c|}{14.49} &
  16.63 \\ \hline
\textbf{0\_0\_100\_0} &
  \multicolumn{1}{c|}{1.77} &
  1.86 &
  \multicolumn{1}{c|}{3.49} &
  3.68 &
  \multicolumn{1}{c|}{6.76} &
  7.15 &
  \multicolumn{1}{c|}{12.68} &
  13.52 &
  \multicolumn{1}{c|}{23.27} &
  25.42 \\ \hline
\textbf{0\_0\_0\_100} &
  \multicolumn{1}{c|}{2.88} &
  2.9 &
  \multicolumn{1}{c|}{5.38} &
  5.45 &
  \multicolumn{1}{c|}{9.89} &
  10.11 &
  \multicolumn{1}{c|}{17.82} &
  18.38 &
  \multicolumn{1}{c|}{\textbf{35.91}} &
  \textbf{38.58} \\ \hline
\textbf{50\_50\_0\_0} &
  \multicolumn{1}{c|}{0.98} &
  1.11 &
  \multicolumn{1}{c|}{1.93} &
  2.19 &
  \multicolumn{1}{c|}{3.75} &
  4.26 &
  \multicolumn{1}{c|}{7.16} &
  8.17 &
  \multicolumn{1}{c|}{13.56} &
  15.47 \\ \hline
\textbf{25\_75\_0\_0} &
  \multicolumn{1}{c|}{0.89} &
  1.01 &
  \multicolumn{1}{c|}{1.78} &
  2.03 &
  \multicolumn{1}{c|}{3.56} &
  4.05 &
  \multicolumn{1}{c|}{7.14} &
  8.11 &
  \multicolumn{1}{c|}{13.94} &
  15.92 \\ \hline
\textbf{75\_25\_0\_0} &
  \multicolumn{1}{c|}{0.87} &
  0.99 &
  \multicolumn{1}{c|}{1.7} &
  1.96 &
  \multicolumn{1}{c|}{3.31} &
  3.8 &
  \multicolumn{1}{c|}{6.24} &
  7.27 &
  \multicolumn{1}{c|}{\textbf{12.28}} &
  \textbf{14.08} \\ \hline
\textbf{50\_0\_50\_0} &
  \multicolumn{1}{c|}{1.09} &
  1.2 &
  \multicolumn{1}{c|}{2.17} &
  2.4 &
  \multicolumn{1}{c|}{4.35} &
  4.8 &
  \multicolumn{1}{c|}{8.7} &
  9.62 &
  \multicolumn{1}{c|}{\textbf{17.41}} &
  \textbf{19.23} \\ \hline
\textbf{25\_0\_75\_0} &
  \multicolumn{1}{c|}{1.28} &
  1.4 &
  \multicolumn{1}{c|}{2.56} &
  2.8 &
  \multicolumn{1}{c|}{5.12} &
  5.6 &
  \multicolumn{1}{c|}{10.23} &
  11.2 &
  \multicolumn{1}{c|}{20.24} &
  22.24 \\ \hline
\textbf{75\_0\_25\_0} &
  \multicolumn{1}{c|}{0.9} &
  1.01 &
  \multicolumn{1}{c|}{1.8} &
  2.01 &
  \multicolumn{1}{c|}{3.59} &
  4.02 &
  \multicolumn{1}{c|}{7.18} &
  8.04 &
  \multicolumn{1}{c|}{\textbf{13.62}} &
  \textbf{15.35} \\ \hline
\textbf{50\_0\_0\_50} &
  \multicolumn{1}{c|}{1.48} &
  1.56 &
  \multicolumn{1}{c|}{2.97} &
  3.12 &
  \multicolumn{1}{c|}{5.93} &
  6.25 &
  \multicolumn{1}{c|}{11.88} &
  12.49 &
  \multicolumn{1}{c|}{23.75} &
  24.97 \\ \hline
\textbf{25\_0\_0\_75} &
  \multicolumn{1}{c|}{1.87} &
  1.94 &
  \multicolumn{1}{c|}{3.75} &
  3.88 &
  \multicolumn{1}{c|}{7.49} &
  7.75 &
  \multicolumn{1}{c|}{14.99} &
  15.5 &
  \multicolumn{1}{c|}{\textbf{29.23}} &
  \textbf{30.88} \\ \hline
\textbf{75\_0\_0\_25} &
  \multicolumn{1}{c|}{1.09} &
  1.18 &
  \multicolumn{1}{c|}{2.19} &
  2.37 &
  \multicolumn{1}{c|}{4.37} &
  4.73 &
  \multicolumn{1}{c|}{8.77} &
  9.47 &
  \multicolumn{1}{c|}{16.78} &
  18.21 \\ \hline
\textbf{0\_50\_50\_0} &
  \multicolumn{1}{c|}{1.21} &
  1.34 &
  \multicolumn{1}{c|}{2.43} &
  2.68 &
  \multicolumn{1}{c|}{4.84} &
  5.35 &
  \multicolumn{1}{c|}{9.69} &
  10.72 &
  \multicolumn{1}{c|}{19.38} &
  21.43 \\ \hline
\textbf{0\_25\_75\_0} &
  \multicolumn{1}{c|}{1.34} &
  1.47 &
  \multicolumn{1}{c|}{2.68} &
  2.94 &
  \multicolumn{1}{c|}{5.37} &
  5.88 &
  \multicolumn{1}{c|}{10.73} &
  11.74 &
  \multicolumn{1}{c|}{21.22} &
  23.34 \\ \hline
\textbf{0\_75\_25\_0} &
  \multicolumn{1}{c|}{1.08} &
  1.21 &
  \multicolumn{1}{c|}{2.16} &
  2.42 &
  \multicolumn{1}{c|}{4.32} &
  4.84 &
  \multicolumn{1}{c|}{8.65} &
  9.69 &
  \multicolumn{1}{c|}{17} &
  19.09 \\ \hline
\textbf{0\_50\_0\_50} &
  \multicolumn{1}{c|}{1.61} &
  1.7 &
  \multicolumn{1}{c|}{3.21} &
  3.39 &
  \multicolumn{1}{c|}{6.42} &
  6.78 &
  \multicolumn{1}{c|}{12.85} &
  13.58 &
  \multicolumn{1}{c|}{25.71} &
  27.17 \\ \hline
\textbf{0\_25\_0\_75} &
  \multicolumn{1}{c|}{2.52} &
  2.56 &
  \multicolumn{1}{c|}{4.76} &
  4.86 &
  \multicolumn{1}{c|}{8.88} &
  9.15 &
  \multicolumn{1}{c|}{16.32} &
  16.96 &
  \multicolumn{1}{c|}{31.78} &
  33.88 \\ \hline
\textbf{0\_75\_0\_25} &
  \multicolumn{1}{c|}{1.28} &
  1.39 &
  \multicolumn{1}{c|}{2.56} &
  2.78 &
  \multicolumn{1}{c|}{5.12} &
  5.56 &
  \multicolumn{1}{c|}{10.24} &
  11.12 &
  \multicolumn{1}{c|}{20.16} &
  21.95 \\ \hline
\textbf{0\_0\_50\_50} &
  \multicolumn{1}{c|}{1.87} &
  1.95 &
  \multicolumn{1}{c|}{3.74} &
  3.91 &
  \multicolumn{1}{c|}{7.47} &
  7.84 &
  \multicolumn{1}{c|}{14.94} &
  15.65 &
  \multicolumn{1}{c|}{29.85} &
  31.28 \\ \hline
\textbf{0\_0\_25\_75} &
  \multicolumn{1}{c|}{2.67} &
  2.69 &
  \multicolumn{1}{c|}{5.05} &
  5.14 &
  \multicolumn{1}{c|}{9.46} &
  9.7 &
  \multicolumn{1}{c|}{17.48} &
  18.06 &
  \multicolumn{1}{c|}{34.12} &
  36.11 \\ \hline
\textbf{0\_0\_75\_25} &
  \multicolumn{1}{c|}{1.67} &
  1.78 &
  \multicolumn{1}{c|}{3.33} &
  3.55 &
  \multicolumn{1}{c|}{6.67} &
  7.1 &
  \multicolumn{1}{c|}{13.34} &
  14.21 &
  \multicolumn{1}{c|}{26.45} &
  28.27 \\ \hline
\textbf{33.34\_33.34\_33.34\_0} &
  \multicolumn{1}{c|}{1.04} &
  1.16 &
  \multicolumn{1}{c|}{2.08} &
  2.32 &
  \multicolumn{1}{c|}{4.18} &
  4.65 &
  \multicolumn{1}{c|}{8.35} &
  9.3 &
  \multicolumn{1}{c|}{16.69} &
  18.6 \\ \hline
\textbf{20\_60\_20\_0} &
  \multicolumn{1}{c|}{1.01} &
  1.13 &
  \multicolumn{1}{c|}{2.01} &
  2.26 &
  \multicolumn{1}{c|}{4.03} &
  4.52 &
  \multicolumn{1}{c|}{8.06} &
  9.05 &
  \multicolumn{1}{c|}{16.08} &
  18.04 \\ \hline
\textbf{20\_20\_60\_0} &
  \multicolumn{1}{c|}{1.21} &
  1.34 &
  \multicolumn{1}{c|}{2.43} &
  2.67 &
  \multicolumn{1}{c|}{4.86} &
  5.35 &
  \multicolumn{1}{c|}{9.7} &
  10.67 &
  \multicolumn{1}{c|}{19.42} &
  21.38 \\ \hline
\textbf{60\_20\_20\_0} &
  \multicolumn{1}{c|}{1.05} &
  1.17 &
  \multicolumn{1}{c|}{2.08} &
  2.33 &
  \multicolumn{1}{c|}{4.08} &
  4.56 &
  \multicolumn{1}{c|}{7.9} &
  8.87 &
  \multicolumn{1}{c|}{15.64} &
  17.36 \\ \hline
\textbf{15\_70\_15\_0} &
  \multicolumn{1}{c|}{0.99} &
  1.12 &
  \multicolumn{1}{c|}{1.99} &
  2.24 &
  \multicolumn{1}{c|}{3.98} &
  4.47 &
  \multicolumn{1}{c|}{7.94} &
  8.94 &
  \multicolumn{1}{c|}{15.67} &
  17.69 \\ \hline
\textbf{15\_15\_70\_0} &
  \multicolumn{1}{c|}{1.28} &
  1.4 &
  \multicolumn{1}{c|}{2.56} &
  2.8 &
  \multicolumn{1}{c|}{5.1} &
  5.6 &
  \multicolumn{1}{c|}{10.22} &
  11.21 &
  \multicolumn{1}{c|}{20.35} &
  22.36 \\ \hline
\textbf{70\_15\_15\_0} &
  \multicolumn{1}{c|}{0.98} &
  1.1 &
  \multicolumn{1}{c|}{1.93} &
  2.18 &
  \multicolumn{1}{c|}{3.78} &
  4.25 &
  \multicolumn{1}{c|}{7.23} &
  8.22 &
  \multicolumn{1}{c|}{14.35} &
  16.09 \\ \hline
\textbf{33.34\_33.34\_0\_33.34} &
  \multicolumn{1}{c|}{1.31} &
  1.4 &
  \multicolumn{1}{c|}{2.61} &
  2.8 &
  \multicolumn{1}{c|}{5.23} &
  5.6 &
  \multicolumn{1}{c|}{10.46} &
  11.22 &
  \multicolumn{1}{c|}{20.92} &
  22.42 \\ \hline
\textbf{20\_60\_0\_20} &
  \multicolumn{1}{c|}{1.17} &
  1.28 &
  \multicolumn{1}{c|}{2.33} &
  2.55 &
  \multicolumn{1}{c|}{4.66} &
  5.1 &
  \multicolumn{1}{c|}{9.33} &
  10.2 &
  \multicolumn{1}{c|}{18.61} &
  20.35 \\ \hline
\textbf{20\_20\_0\_60} &
  \multicolumn{1}{c|}{1.69} &
  1.76 &
  \multicolumn{1}{c|}{3.38} &
  3.53 &
  \multicolumn{1}{c|}{6.75} &
  7.06 &
  \multicolumn{1}{c|}{13.51} &
  14.12 &
  \multicolumn{1}{c|}{26.86} &
  28.23 \\ \hline
\textbf{60\_20\_0\_20} &
  \multicolumn{1}{c|}{1.07} &
  1.17 &
  \multicolumn{1}{c|}{2.14} &
  2.33 &
  \multicolumn{1}{c|}{4.27} &
  4.66 &
  \multicolumn{1}{c|}{8.53} &
  9.31 &
  \multicolumn{1}{c|}{16.78} &
  18.32 \\ \hline
\textbf{15\_70\_0\_15} &
  \multicolumn{1}{c|}{1.11} &
  1.23 &
  \multicolumn{1}{c|}{2.22} &
  2.45 &
  \multicolumn{1}{c|}{4.45} &
  4.91 &
  \multicolumn{1}{c|}{8.89} &
  9.81 &
  \multicolumn{1}{c|}{17.57} &
  19.42 \\ \hline
\textbf{15\_15\_0\_70} &
  \multicolumn{1}{c|}{1.83} &
  1.9 &
  \multicolumn{1}{c|}{3.66} &
  3.81 &
  \multicolumn{1}{c|}{7.32} &
  7.6 &
  \multicolumn{1}{c|}{14.66} &
  15.23 &
  \multicolumn{1}{c|}{28.77} &
  30.37 \\ \hline
\textbf{70\_15\_0\_15} &
  \multicolumn{1}{c|}{0.97} &
  1.07 &
  \multicolumn{1}{c|}{1.95} &
  2.14 &
  \multicolumn{1}{c|}{3.91} &
  4.29 &
  \multicolumn{1}{c|}{7.81} &
  8.6 &
  \multicolumn{1}{c|}{15.03} &
  16.61 \\ \hline
\textbf{0\_33.34\_33.34\_33.34} &
  \multicolumn{1}{c|}{1.56} &
  1.66 &
  \multicolumn{1}{c|}{3.12} &
  3.33 &
  \multicolumn{1}{c|}{6.24} &
  6.66 &
  \multicolumn{1}{c|}{12.49} &
  13.3 &
  \multicolumn{1}{c|}{24.98} &
  26.63 \\ \hline
\textbf{0\_20\_60\_20} &
  \multicolumn{1}{c|}{1.53} &
  1.64 &
  \multicolumn{1}{c|}{3.05} &
  3.28 &
  \multicolumn{1}{c|}{6.1} &
  6.54 &
  \multicolumn{1}{c|}{12.2} &
  13.1 &
  \multicolumn{1}{c|}{24.39} &
  26.19 \\ \hline
\textbf{0\_20\_20\_60} &
  \multicolumn{1}{c|}{2.41} &
  2.45 &
  \multicolumn{1}{c|}{4.6} &
  4.71 &
  \multicolumn{1}{c|}{8.7} &
  8.98 &
  \multicolumn{1}{c|}{16.25} &
  16.89 &
  \multicolumn{1}{c|}{30.55} &
  32.14 \\ \hline
\textbf{0\_60\_20\_20} &
  \multicolumn{1}{c|}{1.32} &
  1.44 &
  \multicolumn{1}{c|}{2.63} &
  2.86 &
  \multicolumn{1}{c|}{5.27} &
  5.73 &
  \multicolumn{1}{c|}{10.54} &
  11.46 &
  \multicolumn{1}{c|}{21.04} &
  22.87 \\ \hline
\textbf{0\_15\_70\_15} &
  \multicolumn{1}{c|}{1.51} &
  1.62 &
  \multicolumn{1}{c|}{3.02} &
  3.26 &
  \multicolumn{1}{c|}{6.04} &
  6.51 &
  \multicolumn{1}{c|}{12.09} &
  13.03 &
  \multicolumn{1}{c|}{24.09} &
  25.98 \\ \hline
\textbf{0\_15\_15\_70} &
  \multicolumn{1}{c|}{1.95} &
  2.02 &
  \multicolumn{1}{c|}{3.89} &
  4.04 &
  \multicolumn{1}{c|}{7.79} &
  8.1 &
  \multicolumn{1}{c|}{15.57} &
  16.16 &
  \multicolumn{1}{c|}{30.6} &
  32.25 \\ \hline
\textbf{0\_70\_15\_15} &
  \multicolumn{1}{c|}{1.23} &
  1.34 &
  \multicolumn{1}{c|}{2.45} &
  2.68 &
  \multicolumn{1}{c|}{4.9} &
  5.37 &
  \multicolumn{1}{c|}{9.82} &
  10.75 &
  \multicolumn{1}{c|}{19.4} &
  21.3 \\ \hline
\textbf{33.34\_0\_33.34\_33.34} &
  \multicolumn{1}{c|}{1.48} &
  1.58 &
  \multicolumn{1}{c|}{2.96} &
  3.15 &
  \multicolumn{1}{c|}{5.93} &
  6.3 &
  \multicolumn{1}{c|}{11.84} &
  12.59 &
  \multicolumn{1}{c|}{23.67} &
  25.16 \\ \hline
\textbf{20\_0\_60\_20} &
  \multicolumn{1}{c|}{1.47} &
  1.58 &
  \multicolumn{1}{c|}{2.95} &
  3.16 &
  \multicolumn{1}{c|}{5.9} &
  6.33 &
  \multicolumn{1}{c|}{11.8} &
  12.65 &
  \multicolumn{1}{c|}{23.61} &
  25.32 \\ \hline
\textbf{20\_0\_20\_60} &
  \multicolumn{1}{c|}{1.79} &
  1.87 &
  \multicolumn{1}{c|}{3.59} &
  3.75 &
  \multicolumn{1}{c|}{7.17} &
  7.48 &
  \multicolumn{1}{c|}{14.34} &
  14.96 &
  \multicolumn{1}{c|}{28.52} &
  29.88 \\ \hline
\textbf{60\_0\_20\_20} &
  \multicolumn{1}{c|}{1.17} &
  1.26 &
  \multicolumn{1}{c|}{2.34} &
  2.53 &
  \multicolumn{1}{c|}{4.68} &
  5.07 &
  \multicolumn{1}{c|}{9.36} &
  10.12 &
  \multicolumn{1}{c|}{18.43} &
  19.97 \\ \hline
\textbf{15\_0\_70\_15} &
  \multicolumn{1}{c|}{1.47} &
  1.58 &
  \multicolumn{1}{c|}{2.96} &
  3.18 &
  \multicolumn{1}{c|}{5.89} &
  6.33 &
  \multicolumn{1}{c|}{11.79} &
  12.69 &
  \multicolumn{1}{c|}{23.49} &
  25.31 \\ \hline
\textbf{15\_0\_15\_70} &
  \multicolumn{1}{c|}{2.47} &
  2.5 &
  \multicolumn{1}{c|}{4.67} &
  4.77 &
  \multicolumn{1}{c|}{8.76} &
  9.02 &
  \multicolumn{1}{c|}{16.2} &
  16.82 &
  \multicolumn{1}{c|}{31.36} &
  33.25 \\ \hline
\textbf{70\_0\_15\_15} &
  \multicolumn{1}{c|}{1.06} &
  1.15 &
  \multicolumn{1}{c|}{2.11} &
  2.31 &
  \multicolumn{1}{c|}{4.22} &
  4.61 &
  \multicolumn{1}{c|}{8.43} &
  9.22 &
  \multicolumn{1}{c|}{16.27} &
  17.84 \\ \hline
\textbf{25\_25\_25\_25} &
  \multicolumn{1}{c|}{1.35} &
  1.45 &
  \multicolumn{1}{c|}{2.7} &
  2.9 &
  \multicolumn{1}{c|}{5.39} &
  5.8 &
  \multicolumn{1}{c|}{10.78} &
  11.58 &
  \multicolumn{1}{c|}{21.57} &
  23.19 \\ \hline
\textbf{40\_40\_10\_10} &
  \multicolumn{1}{c|}{1.04} &
  1.15 &
  \multicolumn{1}{c|}{2.08} &
  2.3 &
  \multicolumn{1}{c|}{4.15} &
  4.59 &
  \multicolumn{1}{c|}{8.3} &
  9.18 &
  \multicolumn{1}{c|}{16.58} &
  18.35 \\ \hline
\textbf{40\_10\_40\_10} &
  \multicolumn{1}{c|}{1.19} &
  1.3 &
  \multicolumn{1}{c|}{2.38} &
  2.6 &
  \multicolumn{1}{c|}{4.77} &
  5.21 &
  \multicolumn{1}{c|}{9.53} &
  10.4 &
  \multicolumn{1}{c|}{19.07} &
  20.82 \\ \hline
\textbf{40\_10\_10\_40} &
  \multicolumn{1}{c|}{1.43} &
  1.52 &
  \multicolumn{1}{c|}{2.86} &
  3.03 &
  \multicolumn{1}{c|}{5.72} &
  6.06 &
  \multicolumn{1}{c|}{11.43} &
  12.12 &
  \multicolumn{1}{c|}{22.87} &
  24.26 \\ \hline
\textbf{10\_40\_40\_10} &
  \multicolumn{1}{c|}{1.26} &
  1.38 &
  \multicolumn{1}{c|}{2.53} &
  2.76 &
  \multicolumn{1}{c|}{5.07} &
  5.54 &
  \multicolumn{1}{c|}{10.13} &
  11.08 &
  \multicolumn{1}{c|}{20.25} &
  22.12 \\ \hline
\textbf{10\_40\_10\_40} &
  \multicolumn{1}{c|}{1.5} &
  1.6 &
  \multicolumn{1}{c|}{3.01} &
  3.2 &
  \multicolumn{1}{c|}{6.01} &
  6.4 &
  \multicolumn{1}{c|}{12.02} &
  12.78 &
  \multicolumn{1}{c|}{24.05} &
  25.58 \\ \hline
\textbf{10\_10\_40\_40} &
  \multicolumn{1}{c|}{1.66} &
  1.75 &
  \multicolumn{1}{c|}{3.32} &
  3.51 &
  \multicolumn{1}{c|}{6.64} &
  7.02 &
  \multicolumn{1}{c|}{13.28} &
  14.03 &
  \multicolumn{1}{c|}{26.54} &
  28.05 \\ \hline
\textbf{10\_10\_10\_70} &
  \multicolumn{1}{c|}{1.89} &
  1.97 &
  \multicolumn{1}{c|}{3.79} &
  3.94 &
  \multicolumn{1}{c|}{7.59} &
  7.88 &
  \multicolumn{1}{c|}{15.17} &
  15.76 &
  \multicolumn{1}{c|}{29.8} &
  31.4 \\ \hline
\textbf{10\_10\_70\_10} &
  \multicolumn{1}{c|}{1.41} &
  1.53 &
  \multicolumn{1}{c|}{2.84} &
  3.08 &
  \multicolumn{1}{c|}{5.68} &
  6.15 &
  \multicolumn{1}{c|}{11.36} &
  12.3 &
  \multicolumn{1}{c|}{22.65} &
  24.56 \\ \hline
\textbf{10\_70\_10\_10} &
  \multicolumn{1}{c|}{1.11} &
  1.23 &
  \multicolumn{1}{c|}{2.22} &
  2.46 &
  \multicolumn{1}{c|}{4.44} &
  4.91 &
  \multicolumn{1}{c|}{8.88} &
  9.84 &
  \multicolumn{1}{c|}{17.55} &
  19.47 \\ \hline
\textbf{70\_10\_10\_10} &
  \multicolumn{1}{c|}{0.96} &
  1.07 &
  \multicolumn{1}{c|}{1.92} &
  2.13 &
  \multicolumn{1}{c|}{3.85} &
  4.26 &
  \multicolumn{1}{c|}{7.71} &
  8.52 &
  \multicolumn{1}{c|}{14.82} &
  16.45 \\ \hline
\textbf{FPS} &
  \multicolumn{1}{c|}{50.47} &
  54.31 &
  \multicolumn{1}{c|}{} &
   &
  \multicolumn{1}{c|}{} &
   &
  \multicolumn{1}{c|}{} &
   &
  \multicolumn{1}{c|}{} &
   \\ \hline

\end{longtable}
\end{landscape}
\begin{table}[h!]
\caption{Total Tokens (in Millions) for each training data set size and configuration. The last row contains English (Eng) and Nepali (Npi) tokens for Full Parallel Sentences. Best Systems are marked in Bold}\label{tab:totalTokens_ennpi_short}
\begin{tabularx}{\textwidth}{c*{5}{|YY}}
\hline
\textbf{Data Set Size} & \multicolumn{2}{>{\centering\arraybackslash}X|}{\textbf{50K}} & \multicolumn{2}{>{\centering\arraybackslash}X|}{\textbf{100K}} & \multicolumn{2}{>{\centering\arraybackslash}X|}{\textbf{200K}} & \multicolumn{2}{>{\centering\arraybackslash}X|}{\textbf{400K}} & \multicolumn{2}{>{\centering\arraybackslash}X}{\textbf{800K}} \\ \hline
\textbf{Language Pair} & \textbf{Eng} & \textbf{Npi} & \textbf{Eng} & \textbf{Npi} & \textbf{Eng} & \textbf{Npi} & \textbf{Eng} & \textbf{Npi} & \textbf{Eng} & \textbf{Npi} \\ \hline
\textbf{RS} & 0.87 & 0.91 & 1.74 & 1.82 & 3.49 & 3.65 & 8.91 & 10.23 & 17.85 & 20.49 \\
\textbf{baselineP} & 1.19 & 1.2 & 2.32 & 2.36 & 4.47 & 4.57 & 8.49 & 8.72 & 15.86 & 16.43 \\
\textbf{100\_0\_0\_0} & 0.6 & 0.67 & 1.17 & 1.29 & 2.25 & 2.48 & 4.14 & 4.57 & 7.2 & 8.32 \\
\textbf{0\_100\_0\_0} & 0.96 & 1.02 & 1.79 & 1.92 & 3.45 & 3.72 & 6.27 & 6.81 & 14.12 & 15.65 \\
\textbf{0\_0\_100\_0} & 1.6 & 1.62 & 3.09 & 3.13 & 5.75 & 5.87 & 11.27 & 11.93 & 24.7 & 27.44 \\
\textbf{0\_0\_0\_100} & 2.54 & 2.41 & 4.59 & 4.41 & 9.5 & 9.77 & 24.87 & 28.65 & \textbf{52.34} & \textbf{62.2} \\
\textbf{50\_50\_0\_0} & 0.8 & 0.86 & 1.56 & 1.69 & 2.96 & 3.21 & 5.7 & 6.2 & 10.41 & 11.38 \\
\textbf{75\_25\_0\_0} & 0.71 & 0.77 & 1.38 & 1.5 & 2.67 & 2.91 & 5.01 & 5.47 & 9.16 & 10.05 \\
\textbf{0\_25\_0\_75} & 2.22 & 2.12 & 4.09 & 3.96 & 7.36 & 7.24 & 19.2 & 21.43 & \textbf{42.45} & \textbf{49.66} \\
\textbf{0\_0\_25\_75} & 2.39 & 2.28 & 4.42 & 4.26 & 8.01 & 7.83 & 20.5 & 22.64 & \textbf{44.75} & \textbf{51.8} \\
\textbf{60\_20\_20\_0} & 0.9 & 0.95 & 1.77 & 1.87 & 3.46 & 3.66 & 6.62 & 7.02 & 12.35 & 13.16 \\
\textbf{0\_20\_20\_60} & 2.15 & 2.07 & 4.02 & 3.91 & 7.4 & 7.28 & 16.7 & 17.84 & \textbf{38.15} & \textbf{43.52} \\
\textbf{70\_15\_15\_0} & 0.83 & 0.88 & 1.63 & 1.73 & 3.17 & 3.38 & 6.07 & 6.47 & 11.21 & 12 \\
\textbf{15\_0\_15\_70} & 2.2 & 2.1 & 4.07 & 3.94 & 7.4 & 7.26 & 18.49 & 20.31 & \textbf{41.26} & \textbf{47.78} \\ \midrule
\textbf{FPS} & 28.45 & 29.77 & & & & & & & & \\ \hline
\end{tabularx}
\end{table}

\begin{table}[h!]
\caption{Total Tokens (in Millions) for each training data set size and configuration. The last row contains English (Eng) and Odia (Ord) tokens for Full Parallel Sentences. Best Systems are marked in Bold}\label{tab:totalTokens_enord_short}
\begin{tabularx}{\textwidth}{c*{5}{|YY}}
\hline
\textbf{Data Set Size} & \multicolumn{2}{>{\centering\arraybackslash}X|}{\textbf{50K}} & \multicolumn{2}{>{\centering\arraybackslash}X|}{\textbf{100K}} & \multicolumn{2}{>{\centering\arraybackslash}X|}{\textbf{200K}} & \multicolumn{2}{>{\centering\arraybackslash}X|}{\textbf{400K}} & \multicolumn{2}{>{\centering\arraybackslash}X}{\textbf{800K}} \\ \hline
\textbf{Language Pair} & \textbf{Eng} & \textbf{Ord} & \textbf{Eng} & \textbf{Ord} & \textbf{Eng} & \textbf{Ord} & \textbf{Eng} & \textbf{Ord} & \textbf{Eng} & \textbf{Ord} \\ \hline
\textbf{RS} & 0.6 & 0.59 & 1.2 & 1.17 & 2.4 & 2.34 & 6.88 & 8.9 & 13.77 & 17.88 \\
\textbf{baselineP} & 0.81 & 0.76 & 1.59 & 1.51 & 3.13 & 2.98 & 6.11 & 5.81 & 11.74 & 11.2 \\
\textbf{100\_0\_0\_0} & 0.52 & 0.5 & 1.03 & 1.0 & 2.05 & 1.99 & 4.04 & 3.93 & 7.92 & 7.73 \\
\textbf{0\_100\_0\_0} & 0.88 & 0.84 & 1.71 & 1.63 & 3.23 & 3.08 & 6.06 & 5.8 & 11.36 & 10.87 \\
\textbf{0\_0\_100\_0} & 1.62 & 1.52 & 3.13 & 2.95 & 5.85 & 5.55 & 10.21 & 9.74 & 22.6 & 30.37 \\
\textbf{0\_0\_0\_100} & 2.42 & 2.16 & 4.37 & 3.96 & 9.76 & 10.73 & 21.45 & 28.08 & \textbf{40.19} & \textbf{57.66} \\
\textbf{50\_50\_0\_0} & 0.7 & 0.67 & 1.4 & 1.34 & 2.74 & 2.63 & 5.28 & 5.07 & 10.1 & 9.73 \\
\textbf{75\_25\_0\_0} & 0.61 & 0.59 & 1.22 & 1.17 & 2.42 & 2.34 & 4.76 & 4.6 & 9.23 & 8.93 \\
\textbf{0\_25\_0\_75} & 2.11 & 1.89 & 3.87 & 3.51 & 7.12 & 6.61 & 17.61 & 21.34 & \textbf{34.62} & \textbf{46.62} \\
\textbf{0\_0\_25\_75} & 2.3 & 2.07 & 4.25 & 3.86 & 7.86 & 7.29 & 19.03 & 22.66 & \textbf{37.24} & \textbf{49.09} \\
\textbf{60\_20\_20\_0} & 0.82 & 0.78 & 1.64 & 1.56 & 3.25 & 3.1 & 6.38 & 6.1 & 12.28 & 11.79 \\
\textbf{0\_20\_20\_60} & 2.06 & 1.86 & 3.85 & 3.5 & 7.11 & 6.53 & 16.24 & 18.13 & \textbf{33.04} & \textbf{41.53} \\
\textbf{70\_15\_15\_0} & 0.75 & 0.71 & 1.49 & 1.42 & 2.96 & 2.83 & 5.83 & 5.59 & 11.34 & 10.9 \\
\textbf{15\_0\_15\_70} & 2.11 & 1.89 & 3.89 & 3.52 & 7.08 & 6.5 & 17.28 & 20.39 & \textbf{34.44} & \textbf{45.25} \\
\midrule
\textbf{FPS} & 69.42 & 67.84 & & & & & & & & \\ \hline
\end{tabularx}
\end{table}
\begin{table}[h!]
\caption{Total Tokens (in Millions) for each training data set size and configuration. The last row contains English (Eng) and Norwegian Nynorsk (Nno) tokens for Full Parallel Sentences. Best Systems are marked in Bold}\label{tab:totalTokens_ennno_short}
\begin{tabularx}{\textwidth}{c*{5}{|YY}}
\hline
\textbf{Data Set Size} & \multicolumn{2}{>{\centering\arraybackslash}X|}{\textbf{50K}} & \multicolumn{2}{>{\centering\arraybackslash}X|}{\textbf{100K}} & \multicolumn{2}{>{\centering\arraybackslash}X|}{\textbf{200K}} & \multicolumn{2}{>{\centering\arraybackslash}X|}{\textbf{400K}} & \multicolumn{2}{>{\centering\arraybackslash}X}{\textbf{800K}} \\ \hline
\textbf{Language Pair} & \textbf{Eng} & \textbf{Nno} & \textbf{Eng} & \textbf{Nno} & \textbf{Eng} & \textbf{Nno} & \textbf{Eng} & \textbf{Nno} & \textbf{Eng} & \textbf{Nno} \\ \hline
\textbf{RS} & 0.95 & 0.96 & 1.9 & 1.93 & 3.81 & 3.86 & 8.95 & 9.28 & 17.92 & 18.56 \\
\textbf{baselineP} & 1.32 & 1.3 & 2.6 & 2.57 & 4.97 & 4.93 & 9.44 & 9.42 & 16.86 & 16.95 \\
\textbf{100\_0\_0\_0} & 0.57 & 0.59 & 1.1 & 1.14 & 2.17 & 2.28 & 3.79 & 4.04 & 8.55 & 9.1 \\
\textbf{0\_100\_0\_0} & 0.94 & 0.96 & 1.84 & 1.89 & 3.45 & 3.57 & 6.94 & 7.17 & 14.68 & 15.27 \\
\textbf{0\_0\_100\_0} & 1.71 & 1.68 & 3.26 & 3.22 & 5.45 & 5.45 & 10.47 & 10.64 & 22.21 & 22.89 \\
\textbf{0\_0\_0\_100} & 2.7 & 2.58 & 4.84 & 4.65 & 9.22 & 8.98 & 17.75 & 17.81 & \textbf{30.05} & \textbf{30.63} \\
\textbf{50\_50\_0\_0} & 0.77 & 0.8 & 1.51 & 1.55 & 2.93 & 3.04 & 5.62 & 5.85 & 10.73 & 11.21 \\
\textbf{75\_25\_0\_0} & 0.67 & 0.69 & 1.32 & 1.37 & 2.57 & 2.68 & 4.93 & 5.16 & 9.38 & 9.92 \\
\textbf{0\_25\_0\_75} & 2.35 & 2.25 & 4.29 & 4.14 & 7.68 & 7.47 & 15.52 & 15.48 & \textbf{28.58} & \textbf{29.07} \\
\textbf{0\_0\_25\_75} & 2.55 & 2.44 & 4.68 & 4.5 & 8.46 & 8.19 & 16.94 & 16.81 & \textbf{30.58} & \textbf{30.95} \\
\textbf{60\_20\_20\_0} & 0.89 & 0.9 & 1.76 & 1.79 & 3.42 & 3.48 & 6.7 & 6.84 & 12.01 & 12.42 \\
\textbf{0\_20\_20\_60} & 2.28 & 2.19 & 4.24 & 4.1 & 7.76 & 7.55 & 15.2 & 15.03 & \textbf{28.59} & \textbf{28.84} \\
\textbf{70\_15\_15\_0} & 0.81 & 0.82 & 1.6 & 1.64 & 3.14 & 3.21 & 6.07 & 6.24 & 11.38 & 11.82 \\
\textbf{15\_0\_15\_70} & 2.34 & 2.24 & 4.29 & 4.13 & 7.77 & 7.54 & 15.53 & 15.41 & \textbf{28.85} & \textbf{29.18} \\
\midrule
\textbf{FPS} & 23.68 & 23.98 & & & & & & & & \\ \hline
\end{tabularx}
\end{table}
 
\begin{table}[h!]
\centering
\caption{Total Tokens (In Millions) for baseline systems and data curated using Cluster 3 for English (Eng) and German (De), FP consisted of 20 Million sentences.}
\label{tab:totalTokens_ende_short}
\begin{tabular}{l|rr}
\hline
\textbf{Dataset Size} & \multicolumn{2}{l}{\textbf{8 Million}} \\ \hline
\textbf{Langauge Pair} & \multicolumn{1}{l}{\textbf{Eng}} & \multicolumn{1}{l}{\textbf{De}} \\ \hline
\textbf{RS}           & 185.22            & 205.87             \\
\textbf{baselineP}    & 185.25            & 205.99             \\
\textbf{0\_0\_0\_100} & \textbf{366.9}    & \textbf{366.84}    \\ \hline
\textbf{FP}           & 465.08            & 517.06            
\end{tabular}%
\end{table}
\end{appendices}
\newpage

\bibliography{sn-article}

\end{document}